\documentclass[11pt]{article}

\usepackage{epsfig,amsmath,latexsym,amssymb}
\usepackage{graphicx}
\usepackage{lscape}
\usepackage{picture, eso-pic, tikz} %% for gray boxes
\usepackage{dsfont}
\usepackage{listings}
\usepackage{changes}
\usepackage{hyperref}

\renewcommand{\baselinestretch}{1.1}
\def\R{{\mathbb R}}  %%
\def\E{{\mathbb E}}  %

\newcommand{\Remm}[1]{}
\newtheorem{theo}{Theorem}[section]

\newtheorem{model ass}[theo]{Model Assumptions}

\newtheorem{example}[theo]{Example}

\newtheorem{rem}[theo]{Remark}
\newtheorem{rems}[theo]{Remark}

\def\EndExample{\hfill {\scriptsize $\blacksquare$}}
\numberwithin{equation}{section}

\definecolor{MyGray}{rgb}{0.92,0.92,0.92}


\definecolor{British racing}{rgb}{0.0, 0.5, 0.0}
\def\bx{\boldsymbol{x}}

\def\bz{\boldsymbol{z}}

\def\bX{\boldsymbol{X}}
\def\b0{\boldsymbol{0}}
\def\ba{\boldsymbol{a}}
\def\bbeta{\boldsymbol{\beta}}

\begin{document}
\author{Michael Merz\footnote{University of Hamburg,
Faculty of Business Administration, michael.merz@uni-hamburg.de} 
\and Ronald Richman\footnote{QED Actuaries \& Consultants, ronald.richman@qedact.com} \footnote{University of the Witwatersrand, Johannesburg}
\and Andreas Tsanakas\footnote{The Business School (formerly Cass), City, University of London, A.Tsanakas.1@city.ac.uk}
\and Mario V.~W\"uthrich\footnote{RiskLab, Department of Mathematics, ETH Zurich,
mario.wuethrich@math.ethz.ch}}

\date{Version of \today}
\title{Interpreting Deep Learning Models with\\ Marginal Attribution by Conditioning on Quantiles}
\maketitle

\begin{abstract}
\noindent  
A vastly growing literature on explaining deep learning models has emerged. This paper contributes to that literature by introducing a global gradient-based model-agnostic method, which we call Marginal Attribution by Conditioning on Quantiles (MACQ). Our approach is based on analyzing the marginal attribution of predictions (outputs) to individual features (inputs). Specifically, we consider variable importance by fixing (global) output levels and, thus, explain how features marginally contribute across different regions of the prediction space. Hence, MACQ can be seen as a marginal attribution counterpart to approaches such as accumulated local effects (ALE), which study the sensitivities of outputs by perturbing inputs. Furthermore, MACQ allows us to separate marginal attribution of individual features from interaction effect, and visually illustrate the 3-way relationship between marginal attribution, output level, and feature value.

~


\end{abstract}

\section{Introduction}
Deep learning models are typically trained to provide an optimal predictive performance.
Interpreting and explaining the results of deep learning models has, until recently, only played a subordinate role. With growing complexity
of deep learning models, the need and requirement of being able to explain deep learning solutions
has become increasingly important. This applies to almost all fields of their applications: deep learning findings
in medical fields and health care need to make sense to patients, loan and mortgage evaluations and credit approvals need
to be understandable to customers, insurance pricing must be explained to insurance policyholders, 
business processes and decisions need to be transparent to regulators, autonomous robotic tools need to
comply with safety standards according to admission offices and governments, etc. These needs are even reinforced
by the requirements of being able to prove that deep learning solutions do not discriminate w.r.t.~protected
features and are in line with data protection regulation. Thus, there is substantial social
and political pressure to be able to explain, illustrate and verify deep learning solutions, 
in order to provide reassurance that these work properly.

Recent research focuses on different methods of explaining deep learning decision making; an
overview is given in \cite{Samek}. 
Some of these methods provide a post-hoc analysis which aims at understanding global model behavior,
explaining individual outcomes and learned representations. Often this is done by explaining representative
examples. We are going to discuss some of these post-hoc analysis
methods
in the literature overview presented in the next section. Other
methods aim at a wider interdisciplinary approach by more broadly examining how decision
making is done in a  social context, see e.g.~\cite{Miller}. All these approaches
have in common that they try to ``open up the black-box'' to make decision making
explainable to stakeholders.

Our paper contributes to this literature. 
Our main contribution is that we provide a novel gradient-based model-agnostic tool
that is motivated by analyzing marginal contributions to deep learning decisions in the spirit of
salience methods, as described in \cite{Ancona}. Salience methods are local
model-agnostic tools that attribute marginal effects on outputs
to different inputs.
Motivated
by sensitivity analysis tools in risk measurement, we aggregate local marginal attributions to obtain
a global picture at a given quantile level of the output variable. 
We call this method Marginal Attribution by Conditioning on Quantiles (MACQ).
It describes a global variable
importance measure that varies with the output level. The aggregation of local marginal effects is justified
by the fact that this aggregation can be seen as a directional derivative of a distortion risk measure, 
see \cite{Hong1} and Proposition 1 in \cite{Tsanakas1}.
As second contribution, we extend this view by including higher order derivatives beyond 
linear marginal contributions. 
This additional step can 
be seen in the context of deep Taylor decompositions (DTD), similar to \cite{Montavon}.
A difficulty in Taylor decompositions is that they depend on a reference point. By rearranging the terms
and by taking advantage of our quantile view, we determine an optimal global reference point
that allows us to quantify both variable importance and interaction strength in our MACQ approach.
The third contribution is that we provide graphic tools that provide a 3-way relationship between 
(i) marginal attribution, (ii) response/output level and (iii) feature value.

\medskip

{\bf Organization.} In the next section we give a literature overview that embeds our MACQ method
into the present toolbox of model explainability. This literature overview is also used to introduce the relevant notation.
In Section \ref{Sensitivities for regression functions} we present our main idea of 
aggregating local marginal attributions to a quantile sensitivity analysis. Section \ref{Interaction strengths}
presents a higher order expansion which grounds a study of interaction strength.
Section \ref{Choice of reference point and local model-diagnostics} discusses the choice of the
reference point. An extended example is presented in Section \ref{Example}. Finally,
in Section \ref{Conclusions} we state brief conclusions.

\section{Literature overview}
\label{Literature overview}
We give a brief summary of recent developments in post-hoc interpretability and explainability tools for deep learning
models. This summary also serves to introduce the relevant notation for this paper.
Assume the following regression function is smooth (in fact, we are only going to use twice
differentiable in our setting)
\begin{equation}\label{plain vanilla regression model}
\mu: \R^q \to \R, \qquad \bx \mapsto \mu(\bx),
\end{equation}
with feature $\bx=(x_1,\ldots, x_q)^\top \in \R^q$. This regression function is assumed
to describe the systematic effects of features on the random variable $Y$ via the (conditional) expectation
\begin{equation*}
\E[Y|\bx]= \mu(\bx).
\end{equation*}
We assume smoothness of regression function \eqref{plain vanilla regression model} because our 
model-agnostic proposal will be gradient-based. In our example in Section \ref{Example}, we will use
a deep feed-forward neural network on tabular
input data, having the hyperbolic tangent as activation function. This gives us
a smooth regression function and formal derivation can be done in standard software such as 
TensorFlow/Keras and PyTorch.

\subsection{Model-agnostic tools}
Recent literature aims understanding such regression functions 
\eqref{plain vanilla regression model} coming from deep learning models.
One approach is to analyze marginal plots.
We select one component $x_j$ of $\bx$ and study the function
\begin{equation*}
x_j \in \R ~\mapsto ~
\mu(x_j , \bx_{\setminus j}),
\end{equation*}
where $\bx_{\setminus j}$ denotes the remaining components
of $\bx$ which are kept fixed. This is the method of individual conditional expectation (ICE)
of \cite{Goldstein}. 
If we have thousands or millions of instances $(Y,\bx)$, it might be advantageous to study 
ICE profiles on an aggregated level. 
This is the proposal of \cite{friedman2001greedy} and \cite{zhao2019causal}
called partial dependence plots (PDPs). We introduce the feature distribution $P$ which describes
the family of all (potential) features $\bX \sim P$. The PDP profile of component $1\le j \le q$
is defined by
\begin{equation*}
x_j \mapsto \E_P\left[\mu(x_j, \bX_{\setminus j}) \right] = 
\int \mu(x_j, \bx_{\setminus j}) dP(\bx_{\setminus j}).
\end{equation*}
The critical point in this approach is that it does not reflect the (true) dependence structure between 
feature components $X_j$ and $\bX_{\setminus j}$, i.e., as described by feature
distribution $P$. The method of accumulated local effects (ALEs) introduced by 
 \cite{Apley} aims at correctly incorporating the dependence structure in $\bX$.
The local effect of
component $x_j$ in individual feature $\bx$ is given by the partial derivative
\begin{equation}\label{local effects}
\mu_{j}(\bx) = \frac{\partial \mu(\bx)}{\partial x_j}.
\end{equation}
The average local effect of component $1\le j \le q$ is obtained by
\begin{equation}\label{average local effect}
x_j \mapsto \Delta_j(x_j) =\E_P\left[\left.\mu_{j}(\bX)\right| X_j=x_j\right] 
=\int \mu_{j}(x_j, \bx_{\setminus j})  dP(\bx_{\setminus j}|x_j),
\end{equation}
where $P(\bx_{\setminus j}|x_j)$ denotes the conditional distribution of 
$\bX_{\setminus j}$, given $X_j=x_j$.
ALEs integrate the average local effects $\Delta_j(\cdot)$ over their domain,
thus, the ALE profile is defined by
\begin{equation}\label{ALE}
x_j \mapsto \int_{x_{j_0}}^{x_j} \Delta_j(z_j)  dz_j
=\int_{x_{j_0}}^{x_j} \int \mu_{j}(z_j, \bx_{\setminus j})  dP(\bx_{\setminus j}|z_j)  dz_j,
\end{equation}
where ${x_{j_0}}$ is a given initialization point.
The main difference between PDPs and ALEs is that the latter correctly considers
the dependence structure between $X_j$ and $\bX_{\setminus j}$.
\begin{rems}\normalfont
\begin{itemize}
\item The main difference between PDPs and ALEs is that the latter correctly considers
the dependence structure between $X_j$ and $\bX_{\setminus j}$. The two profiles
coincide if $X_j$ and $\bX_{\setminus j}$ are independent under $P$.
\item \cite{Apley} provide a discretized version of  the ALE profile that
can also be applied to non-differentiable regression functions $\mu(\cdot)$.
Basically, this can be received either by finite differences or by a local analysis in an
environment of a selected feature value $x_j$.
\item More generally the local effect \eqref{local effects} allows us to consider a 1st order Taylor expansion.
Denote by $\nabla_{\bx} \mu(\bx)$ the gradient of $\mu(\cdot)$ w.r.t.~$\bx$. We 
have
\begin{equation}\label{first order Taylor expansion}
\mu(\bx+ \boldsymbol{\epsilon})= \mu(\bx) + (\nabla_{\bx} \mu(\bx))^\top \boldsymbol{\epsilon} + o(\|\boldsymbol{\epsilon}\|), 
\end{equation}
for $\boldsymbol{\epsilon} \in \R^q$ going to zero.
This gives us a 1st order local approximation to $\mu(\cdot)$ in $\bx$, which reflects the local (linear)
behavior similar to the locally interpretable model-agnostic explanation (LIME) introduced
by \cite{Ribeiro}. That is, \eqref{first order Taylor expansion} fits a local
linear regression model around $\mu(\bx)$ with regression parameters described
by the components of the gradient $\nabla_{\bx} \mu(\bx)$. LIME then uses regularization,
e.g.~LASSO, to select the most relevant feature components in the neighborhood
of $\mu(\bx)$. 
\item More generally, \eqref{first order Taylor expansion} defines a local surrogate model that
can be used for a local sensitivity analysis by perturbing $\bx$ within a small environment.
White-box surrogate models are popular tools to explain complex regression functions, for instance,
decision trees can be fit to network regression models for extracting the most relevant feature information.
\end{itemize}
\end{rems}

\subsection{Gradient based model-agnostic tools}
Gradient-based model-agnostic tools can be used to attribute outputs to (feature) inputs. Attribution denotes the
process of assigning a relevance index to input components, in order to explain a certain output, see \cite{Efron}. \cite{Ancona}
provide a nice overview of gradient-based attribution methods. In formula
\eqref{local effects} of the previous subsection we have met a first attribution method which gives 
the sensitivity of the output $\mu(\bx)$ as a function of the input $\bx$. The
marginal attribution we are going to present considers the contribution to a given output in the
spirit of salience methods.

Marginal attribution is obtained by considering the directional derivative w.r.t.~the 
features
\begin{equation}
\label{marginal attribution}
x_j~\mapsto~x_j\mu_{j}(\bx) ~=~ x_j \frac{\partial \mu(\bx)}{\partial x_j}.
\end{equation}
This has first been discussed in the machine learning community by \cite{Shrikumar} 
who observed that this can make attribution more concise; these directional derivatives
have been coined Gradient*Input in the machine learning literature, see Ancona \cite{Ancona}. Mathematically speaking, these marginal attributions
can be understood as individual contributions to a certain value in a Taylor series sense (and relative to a reference point). 
Having a linear regression model $\bx\mapsto \beta_0 + \sum_{j=1}^q \beta_j x_j$, the marginal attributions
give an additive decomposition of the regression function, and $\beta_j$ can be considered as the
relevance index of component $j$. In non-linear regression models, such a linear decomposition
only holds true very locally, see \eqref{first order Taylor expansion}, and other methods such as the
Shapley value \cite{Shapley} are used to quantify non-linear effects and interaction effects, see 
\cite{LundbergLee}. We also mention \cite{Sundararajan}, who consider
integrated gradients 
\begin{equation}\label{integrated gradients}
x_j~\mapsto ~x_j \int_0^1  \mu_j\left(\bx_0+ z (\bx-\bx_0)\right) dz,
\end{equation}
for a given reference point $\bx_0$. This mitigates the problem of only being accurate locally.
In practice, however, this is computationally demanding, similarly to the Shapley value.

There are other methods that are specific to deep networks.
We mention layer-wise propagation (LRP) by \cite{Binder} and
DeepLIFT (Deep Learning Important FeaTures) by \cite{DeepLIFT}.
These methods use a backward pass from the output to the input. In this backward
pass a relevance index (budget) is locally redistributed (recursively from layer to layer), 
resulting in a relevance index on the inputs (for the given output). 
\cite{Ancona} show in Propositions 1 and 2 that these two methods can
be understood as an average over marginal attributions. We remark that
these methods are mainly used for convolutional neural networks (CNNs), e.g., in 
image recognition, whereas our MACQ proposal is more suitable for tabular data because
we require differentiability w.r.t.~the inputs $\bx$. CNNs architectures are often non-differentiable because
of the use of max-pooling layers.

Our contribution builds on marginal attributions \eqref{marginal attribution}. Marginal attributions
are, by definition, local explanations, and we are going to show how to integrate these local
considerations into a global variable importance analysis. \cite{Samek}
call such an aggregation of indivdiual explanations a {\it global meta-explanation}. As a consequence,
our MACQ approach is the marginal attribution counterpart to ALEs by fixing (global) output levels and
describing how features marginally contribute to these levels, whereas ALEs rather study the
sensitivities of the outputs by perturbing the inputs.

\section{Marginal attribution by conditioning on quantiles}
\label{Sensitivities for regression functions}

We consider regression model \eqref{plain vanilla regression model} from a marginal
attribution point of view. Motivated by the risk sensitivity tools of \cite{Hong1} and 
\cite{Tsanakas1}, we do not consider average local effects \eqref{average local effect} 
conditioned on event $\{X_j=x_j\}$, but we would rather like to understand how feature components
contribute to a certain response level $\mu(\bx)$. The former studies sensitivities of outputs $\mu(\bx)$
in inputs $\bx$, whereas the latter considers marginal attribution of outputs $\mu(\bx)$ to inputs $\bx$. This allows us to study how the 
response levels are composed in different regions of the decision space, as this is of intrinsic interest e.g.~in financial 
applications.

Select a quantile level $\alpha \in (0,1)$, the $\alpha$-quantile
of $\mu(\bX)$ is given by
\begin{equation*}
F_{\mu(\bX)}^{-1}(\alpha) = \inf \left\{ y \in \R; ~F_{\mu(\bX)}(y) \ge \alpha \right\},
\end{equation*}
where $F_{\mu(\bX)}(y)=P[\mu(\bX) \le y ]$ describes the distribution function of $\mu(\bX)$.

\medskip

The {\it 1st order attributions} to components $1\le j \le q$ on quantile level $\alpha$ are defined by
\begin{equation}\label{VaR sensitivity}
S_j( \mu; \alpha) 
~=~\E_P \left[X_j \mu_j(\bX) \left| \mu(\bX)=F_{\mu(\bX)}^{-1}(\alpha)\right.\right].
\end{equation}
These are the marginal attributions by conditioning on quantiles (MACQ).

\cite{Tsanakas1} show that \eqref{VaR sensitivity} naturally arises
as sensitivities of distortion risk measures, and choosing the $\alpha$-Dirac distortion we exactly receive
\eqref{VaR sensitivity}, which corresponds to the sensitivities of the value-at-risk  (VaR) risk measure
on the given quantile level. Thus, the sensitivities of the VaR risk measure can be described
by the average of the marginal attributions $X_j \mu_j(\bX)$, conditioned on being on the corresponding
quantile level. The interested reader is referred to Appendix \ref{Sensitivities in distortion risk measures}
for a more detailed description of distortion risk measures.

Alternatively, we can describe the 1st order attributions \eqref{VaR sensitivity} by a 
1st order Taylor expansion \eqref{first order Taylor expansion} in feature perturbation
$\boldsymbol{\epsilon}=-\bx$
\begin{equation}\label{LIME for 0}
\mu(\b0) ~\approx~ \mu\left(\bx\right) - \left(\nabla_{\bx}\mu(\bx)\right)^\top \bx.
\end{equation}
This explains that the 1st order attributions \eqref{VaR sensitivity} describe a 1st order Taylor
approximation at the common {\it reference point} $\b0$, and rearranging  the terms we get the {\it 1st order contributions}
to a given response level
\begin{equation}\label{first order contribution}
F_{\mu(\bX)}^{-1}(\alpha)=
\E_P \left[\mu\left(\bX\right)\left| \mu(\bX)=F_{\mu(\bX)}^{-1}(\alpha)\right.\right] ~\approx~  \mu\left(\b0\right) + 
\sum_{j=1}^q S_j( \mu; \alpha).
\end{equation}

\begin{rems}\normalfont
\begin{itemize}
\item 
A 1st order Taylor expansion \eqref{first order Taylor expansion} gives a local model-agnostic description
in the spirit of LIME. Explicit choice $\boldsymbol{\epsilon}=-\bx$ provides \eqref{LIME for 0},
which can be viewed as a local description of $\mu(\b0)$ relative to $\bx$. The 1st order contributions
\eqref{first order contribution} combine all these local descriptions \eqref{VaR sensitivity} w.r.t.~a given
quantile level to get the integrated MACQ view of $\mu(\b0)$, i.e.
\begin{equation*}
\mu(\b0) ~\approx~ \E\left[ \mu\left(\bX\right) - \left(\nabla_{\bx}\mu(\bX)\right)^\top \bX
\left| \mu(\bX)=F_{\mu(\bX)}^{-1}(\alpha)\right.\right]=
F_{\mu(\bX)}^{-1}(\alpha)-\sum_{j=1}^q S_j( \mu; \alpha).
\end{equation*}
This exactly corresponds to 1st order approximation \eqref{first order contribution}. 
In the sequel it is less important that we can approximate $\mu(\b0)$ by this integrated view, but
$\mu(\b0)$ plays the role of the {\it reference level} that calibrates our global meta-explanation. Thus,
all explanations made are understood relative to this reference level $\mu(\b0)$.
\item In \eqref{LIME for 0}-\eqref{first order contribution} we implicitly assumed that $\b0$
is a suitable reference point for calibrating our global meta-explanation. We further explore and improve
this calibration in Section \ref{Choice of reference point and local model-diagnostics}, below. 
\item Integrated gradients \eqref{integrated gradients} integrate along a single path from a reference point
$\bx_0$ to $\bx$ to make the 1st order Taylor approximation precise. We exchange the roles of the points, here,
and we approximate the reference point by aggregating over all local descriptions in features $\bX$.
\item 1st order contributions \eqref{first order contribution} provide a 3-way description of the regression
function, namely, they combine (i) marginal attribution $S_j( \mu; \alpha)$ as a function of $1\le j \le q$, (ii)
response level $F_{\mu(\bX)}^{-1}(\alpha)$ as a function of $\alpha$, and (iii) feature values $x_j$. In our application in Section \ref{Example} we will illustrate the data from these different angles, each
having its importance in explaining the response.
\item 1st order attribution \eqref{VaR sensitivity} combines marginal attributions $X_j \mu_j(\bX)$ 
by focusing on a common
quantile level. A similar approach could also be done for other model-agnostic tools, such as the Shapley
value.
\end{itemize}
\end{rems}

\begin{example}[linear regression]\label{linear regression model}
\normalfont
A linear regression model considers regression function
\begin{equation}\label{linear regression definition}
\bx ~\mapsto ~ \mu(\bx) = \beta_0+  \bbeta^\top \bx,
\end{equation}
with bias/intercept $\beta_0 \in \R$ and regression parameter $\bbeta \in \R^q$. The 1st order contributions
\eqref{first order contribution} are for $\alpha \in (0,1)$ given by
\begin{equation}\label{contribution linear regression model}
F_{\mu(\bX)}^{-1}(\alpha)
=\mu\left(\b0\right) + 
\sum_{j=1}^q S_j( \mu; \alpha)
= \beta_0 + 
\sum_{j=1}^q \beta_j \E_P\left[X_j\left| \mu(\bX)=F_{\mu(\bX)}^{-1}(\alpha)\right.\right].
\end{equation}
Thus, we weight regression parameters $\beta_j$ with the feature components $X_j$ according to their
contributions to quantile $F_{\mu(\bX)}^{-1}(\alpha)$; and the reference point $\b0$ is given
naturally providing initialization $\mu(\b0)=\beta_0$. 

This MACQ explanation \eqref{contribution linear regression model} is rather different from the ALE profile\eqref{ALE}.
If we
initialize ${x_{j_0}}=0$ we receive ALE profile for the linear regression model
\begin{equation*}
x_j \mapsto \int_{0}^{x_j} \Delta_j(z_j)  dz_j
= \beta_j x_j.
\end{equation*}
This is exactly the marginal attribution \eqref{marginal attribution} of component $j$ 
in the linear regression model and it
 explains the change of the linear regression function if we change feature component $x_j$, whereas \eqref{contribution linear regression model} describes the contribution of each feature component to an expected response  level $\mu(\bx)$.
\EndExample
\end{example}

In general, Taylor expansion \eqref{first order contribution} is accurate if the distance between
$\boldsymbol{0}$ and $\bX$ is small enough for all relevant $\bX$, and if the regression function can be well described around
$\mu(\bX)$ by a linear function. 
The former requires that the reference point is chosen
somewhere ``in the middle'' of the feature distribution $P$. The accuracy of the 1st order
approximation is quantified by
\begin{equation}\label{linear accuracy}
\left|
F_{\mu(\bX)}^{-1}(\alpha)
-\mu\left(\b0\right) -
\sum_{j=1}^q S_j( \mu; \alpha) \right|.
\end{equation}
Thus, we want \eqref{linear accuracy} to be small uniformly in quantile level $\alpha$, for the given reference
point $\b0$, as then the 1st order attributions give a good description on all quantile levels $\alpha$. In the
linear regression case this description is exact, see \eqref{contribution linear regression model}.
In contrast to the Taylor decomposition in \cite{Montavon}, the quantiles 
$F_{\mu(\bX)}^{-1}(\alpha)$ give
us a natural anchor point for determining a suitable reference point, which is also computationally feasible.
This will be done in Section \ref{Choice of reference point and local model-diagnostics}.

\section{Interaction strength}
\label{Interaction strengths}
\cite{FriedmanPopescu} and \cite{Apley} have shown how higher order derivatives of $\mu(\cdot)$ allow us to study interaction 
strength in systematic effects. This requires the study of higher order Taylor expansions.
The 2nd order
Taylor expansion is given by
\begin{equation}\label{second order local expansion}
\mu(\bx+ \boldsymbol{\epsilon})= \mu(\bx) + (\nabla_{\bx} \mu(\bx))^\top \boldsymbol{\epsilon} 
+\frac{1}{2}\boldsymbol{\epsilon}^\top (\nabla_{\bx}^2\mu(\bx))\boldsymbol{\epsilon}
+ o(\|\boldsymbol{\epsilon}\|^2), 
\end{equation}
where $\nabla_{\bx}^2\mu$ denotes the Hessian of $\mu$ w.r.t.~$\bx$.
Setting $\boldsymbol{\epsilon}=-\bx$ allows us, in complete analogy to \eqref{first order contribution}, to
study {\it 2nd order contributions}
\begin{equation}\label{second order contributions}
F_{\mu(\bX)}^{-1}(\alpha)
 ~\approx~  \mu\left(\b0\right) + 
\sum_{j=1}^q S_j( \mu; \alpha)
-\frac{1}{2} \sum_{j,k=1}^q T_{j,k}(\mu; \alpha),
\end{equation}
with 2nd order attributions, for $1\le j,k \le q$,
\begin{equation}\label{second order sensitivities}
T_{j,k}(\mu; \alpha)=\E_P \left[X_j X_k \mu_{j,k}(\bX)\left| \mu(\bX)=F_{\mu(\bX)}^{-1}(\alpha)\right.\right].
\end{equation}
Slightly rearranging  the terms in \eqref{second order local expansion} allows us to study individual feature contributions
and interaction terms separately, that is,
\begin{equation}\label{second order contributions 2}
F_{\mu(\bX)}^{-1}(\alpha)
 ~\approx~  \mu\left(\b0\right) + 
\sum_{j=1}^q \left( S_j( \mu; \alpha)-\frac{1}{2} T_{j,j}(\mu; \alpha) \right)
- \sum_{1\le j < k\le q} T_{j,k}(\mu; \alpha).
\end{equation}
The latter term quantifies all 2nd order contributions coming from interactions between $X_j$ and $X_k$, $j\neq k$.
We will show how interaction effects can be included in individual features' marginal
attributions towards the end of Section \ref{Interaction terms section}.

\begin{rem} \normalfont
The motivation for studying 1st order attributions \eqref{VaR sensitivity}
has been given in terms of the risk sensitivity tools of  \cite{Hong1} and 
\cite{Tsanakas1}. These are obtained by calculating directional
derivatives of distortion risk measures (using a Dirac distortion, see
Appendix \ref{Sensitivities in distortion risk measures}). This argumentation does not carry forward to the 
2nd order terms \eqref{second order sensitivities}, as 2nd order directional derivatives
of distortion risk measures turn out to be much more complicated, even in the linear case, see Property 1 in \cite{Gourieroux}.
\end{rem}

\section{Choice of reference point}
\label{Choice of reference point and local model-diagnostics}
To obtain sufficient accuracy in 1st and 2nd order approximations, respectively,  the reference point should lie somewhere
``in the middle'' of the feature distribution $P$. We elaborate on this
 in this section. Typically, we want to get the following expression small, uniformly
in $\alpha \in (0,1)$,
\begin{equation*}
\left|
F_{\mu(\bX)}^{-1}(\alpha)
-\mu\left(\b0\right) -
\sum_{j=1}^q S_j( \mu; \alpha) +\frac{1}{2} \sum_{j,k=1}^q T_{j,k}(\mu; \alpha) \right|.
\end{equation*}
This expression is for reference point $\b0$. However, we can select any other reference point $\ba \in \R^q$,
by exploring the 2nd order Taylor expansion \eqref{second order local expansion} for $\boldsymbol{\epsilon}=\ba-\bx$.
This latter reference point then provides us with a 2nd order approximation
\begin{eqnarray}\label{second order contributions a}
F_{\mu(\bX)}^{-1}(\alpha)
 &\approx&  \mu\left(\ba\right) -
 \E_P \left[(\ba-\bX)^\top \nabla_{\bx}\mu(\bX)\left| \mu(\bX)=F_{\mu(\bX)}^{-1}(\alpha)\right.\right]
\\\nonumber&&\hspace{2cm}-~  
\frac{1}{2}\E_P \left[(\ba-\bX)^\top (\nabla_{\bx}^2\mu(\bX))(\ba-\bX)\left| \mu(\bX)
=F_{\mu(\bX)}^{-1}(\alpha)\right.\right].
\end{eqnarray}
The same can be received by translating the distribution $P$ of the features
by setting $\bX^{\ba}= \bX-\ba$ and letting $\mu^{\ba}(\cdot)=\mu(\ba+\cdot)$. 
Approximation \eqref{second order contributions a} motivates us to look for a reference
point $\ba \in \R^q$ which makes the 2nd order approximation as accurate as possible for ``all'' quantile levels.
Being a bit less ambitious, we select a  discrete
quantile grid $0 < \alpha_1 < \ldots < \alpha_L <1$ on which we would like to have a good approximation
capacity. Define the events ${\cal A}_l=\{\mu(\bX)=F_{\mu(\bX)}^{-1}(\alpha_l)\}$ for $1\le l \le L$. 
Consider the objective function
\begin{eqnarray}\label{objective function}
G(\ba;\mu)&=&
\sum_{l=1}^L
 \Bigg(
F_{\mu(\bX)}^{-1}(\alpha_l)-\mu\left(\ba\right) +
 \E_P\left.\left[(\ba-\bX)^\top \nabla_{\bx} \mu(\bX)\right| {\cal A}_l\right]
\\&&\hspace{4cm}+~\frac{1}{2}~
\E_P \left.\left[(\ba-\bX)^\top (\nabla^2_{\bx} \mu(\bX))(\ba-\bX)^\top\right| {\cal A}_l\right]
\Bigg)^2.\nonumber
\end{eqnarray}
Minimizing this objective function in $\ba$ gives an optimal reference point 
w.r.t.~the quantile levels $(\alpha_l)_{1\le l \le L}$. Unfortunately, $\ba \mapsto
G(\ba;\mu)$ is not a convex function, and therefore numerical methods may only find
local minima. These can be found by a plain vanilla gradient descent method.
We calculate the gradient of $G$ w.r.t.~$\ba$
\begin{eqnarray*}
\nabla_{\ba}G(\ba;\mu)&=&2
\sum_{l=1}^L
 \Bigg(
F_{\mu(\bX)}^{-1}(\alpha_l)-\mu\left(\ba\right) +
 \E_P\left.\left[(\ba-\bX)^\top \nabla_{\bx} \mu(\bX)\right| {\cal A}_l\right]
\\&&\hspace{4cm}+~\frac{1}{2}~
\E_P \left.\left[(\ba-\bX)^\top (\nabla^2_{\bx} \mu(\bX))(\ba-\bX)^\top\right| {\cal A}_l\right]
\Bigg)
\\&& \hspace{1cm}
\times \Bigg(-\nabla_{\ba}\mu\left(\ba\right)
+\E_P \left.\left[ \nabla_{\bx}\mu(\bX)\right| {\cal A}_l\right]
\\&& \hspace{3cm}
-~\E_P\left.\left[ \bX^\top\nabla^2_{\bx}\mu(\bX)\right| {\cal A}_l\right]
+\frac{1}{2}\ba^\top\E_P\left.\left[ \nabla^2_{\bx}\mu(\bX)\right|{\cal A}_l\right]
\Bigg).
\end{eqnarray*}
The gradient descent algorithm then provides for a tempered learning rate $\varepsilon_{t+1}>0$ updates
at algorithmic time $t$
\begin{equation}\label{gradient descent update}
\ba^{(t)} ~\mapsto ~\ba^{(t+1)}=\ba^{(t)} -\varepsilon_{t+1} \nabla_{\ba}G(\ba^{(t)};\mu).
\end{equation}
This step-wise locally decreases the objective function $G$.

\begin{rem}\normalfont
The above algorithm provides a global optimal reference point, thus, a calibration for a global
2nd order  meta-explanation. In some cases this global calibration
may not be satisfactory, in particular, if the reference point is far from the feature values $\bX=\bx$
that mainly describe a given quantile level $F_{\mu(\bX)}^{-1}(\alpha)$, i.e.~through the corresponding
conditional probability $P[\,\cdot\,|\mu(\bX)=F_{\mu(\bX)}^{-1}(\alpha)]$. In that case, one may be interested in
different local reference points that are optimal
for certain quantile levels, say, between 95\% and 99\%. In some sense, this
will provide a more ``honest'' description \eqref{second order contributions} because we do 
not try to simultaneously describe all quantile levels. The downside of multiple reference points
is that we lose comparability of marginal effects across the whole decision space.
\end{rem}

\section{Example}
\label{Example}

\subsection{Model choice and model fitting}
\label{Model choice and model fitting}
We consider the bike rental example of  
\cite{Fanaee} which has also been studied in \cite{Apley}.
The data describes the bike sharing process over the years 2011 and 2012 of the Capital
Bikesharing system in Washington DC. On an hourly time grid we have information about the proportion of 
casual bike rentals relative to all  bike rentals of casual and registered users. This data is supported
by explanatory variables such as weather conditions and seasonal variables. We provide
a descriptive analysis of this data in Appendix \ref{Descriptive analysis of bike rental example}.
On average 17\% of all bike rentals are made by casual users and 83\%  are done
by registered users. However, these proportions heavily fluctuate w.r.t.~daytime,
holidays, weather conditions, etc. This variability is illustrated in Figure \ref{descriptive counts 2}
in Appendix \ref{Descriptive analysis of bike rental example}. We design a neural network regression function to
forecast the proportion of casual rentals. We denote this response variable (proportion) by $Y$, and
we denote the features (explanatory variables) by $\bx \in \R^q$.

For our example we choose a fully-connected feed-forward neural 
network $\theta: \R^q \to \R$ of depth $d=3$ having $(q_1,q_2,q_3)=(20,15,10)$ neurons in the
three hidden layers. This provides us with network regression function
\begin{equation}\label{logistic network definition}
\bx\in \R^q ~~\mapsto ~~ \mu(\bx) = \sigma (\theta(\bx))~\in (0,1),
\end{equation}
where $\sigma$ is the sigmoid output activation, and $\bx \mapsto \theta(\bx)$ models
the canonical parameter of a logistic regression model.
In order to have a smooth network regression function
we choose the hyperbolic tangent as activation function in the three hidden layers.
We have implemented this network in \cite{TensorFlow} and \cite{Keras}, these allow us to directly
formally calculate gradients and Hessians.

In all what follows we do not consider the
attributions of the regression function $\bx \mapsto \mu(\bx)$, but we directly focus on the corresponding
attributions on the canonical scale $\bx \mapsto \theta(\bx)$. This has the advantage that
the results do not get distorted by the sigmoid output activation. Thus, we replace $\mu$
by $\theta$ in \eqref{second order contributions 2}, resulting in the study of
2nd order contributions
\begin{equation}\label{second order contributions 3}
F_{\theta(\bX)}^{-1}(\alpha)
 ~\approx~  \theta\left(\b0\right) + 
\sum_{j=1}^q \left( S_j( \theta; \alpha)-\frac{1}{2} T_{j,j}(\theta; \alpha) \right)
- \sum_{1\le j < k\le q} T_{j,k}(\theta; \alpha).
\end{equation}
The network architecture is fitted to the available data using early stopping
to prevent from over-fitting. Importantly, we do not say here anything about the quality of the 
predictive model, but we aim at understanding the fitted regression function 
$\bx \mapsto \theta(\bx)$.
This can be done regardless whether the chosen model is suitable for the predictive task at hand.

\begin{figure}[htb!]
\begin{center}
\begin{minipage}[t]{0.45\textwidth}
\begin{center}
\includegraphics[width=\textwidth]{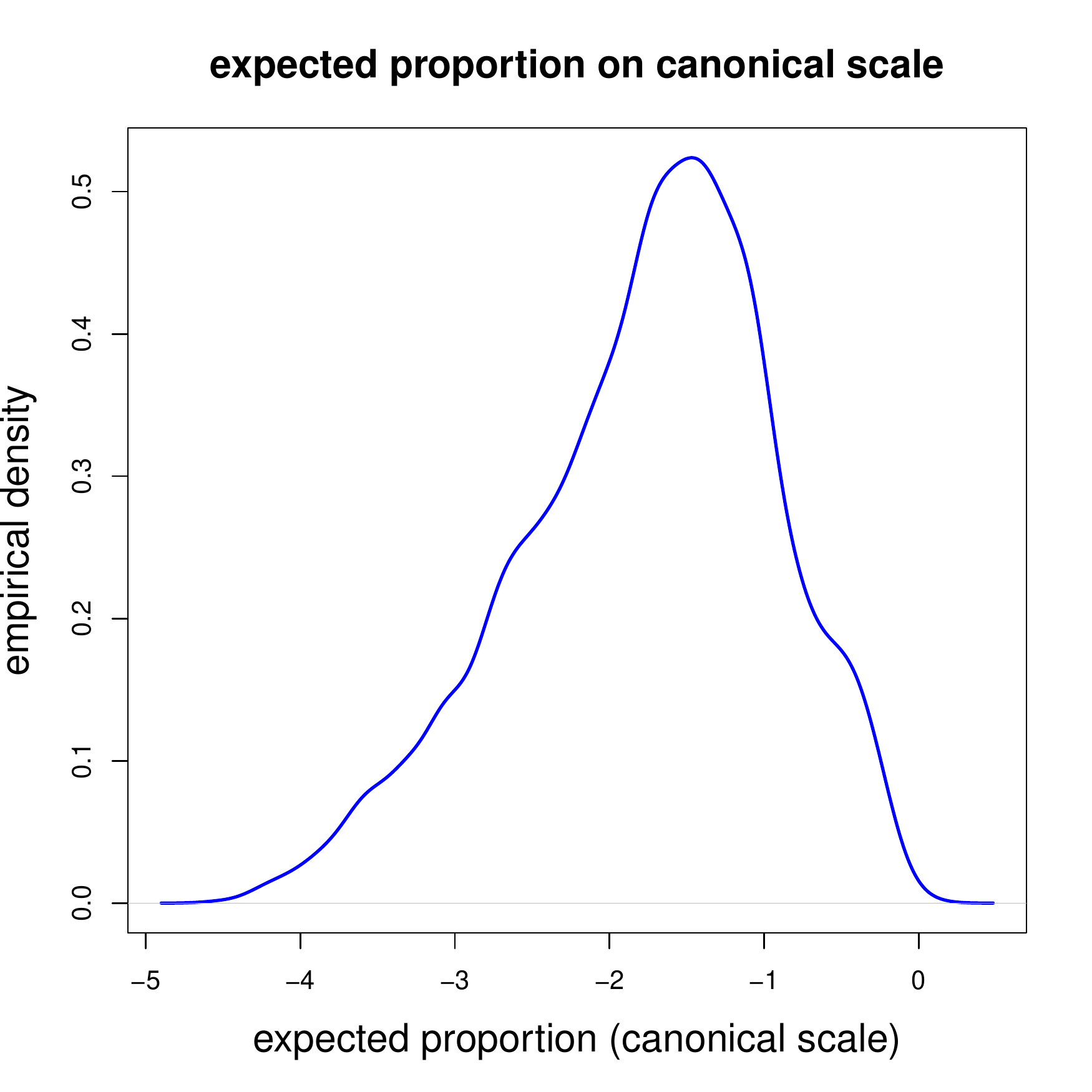}
\end{center}
\end{minipage}
\begin{minipage}[t]{0.45\textwidth}
\begin{center}
\includegraphics[width=\textwidth]{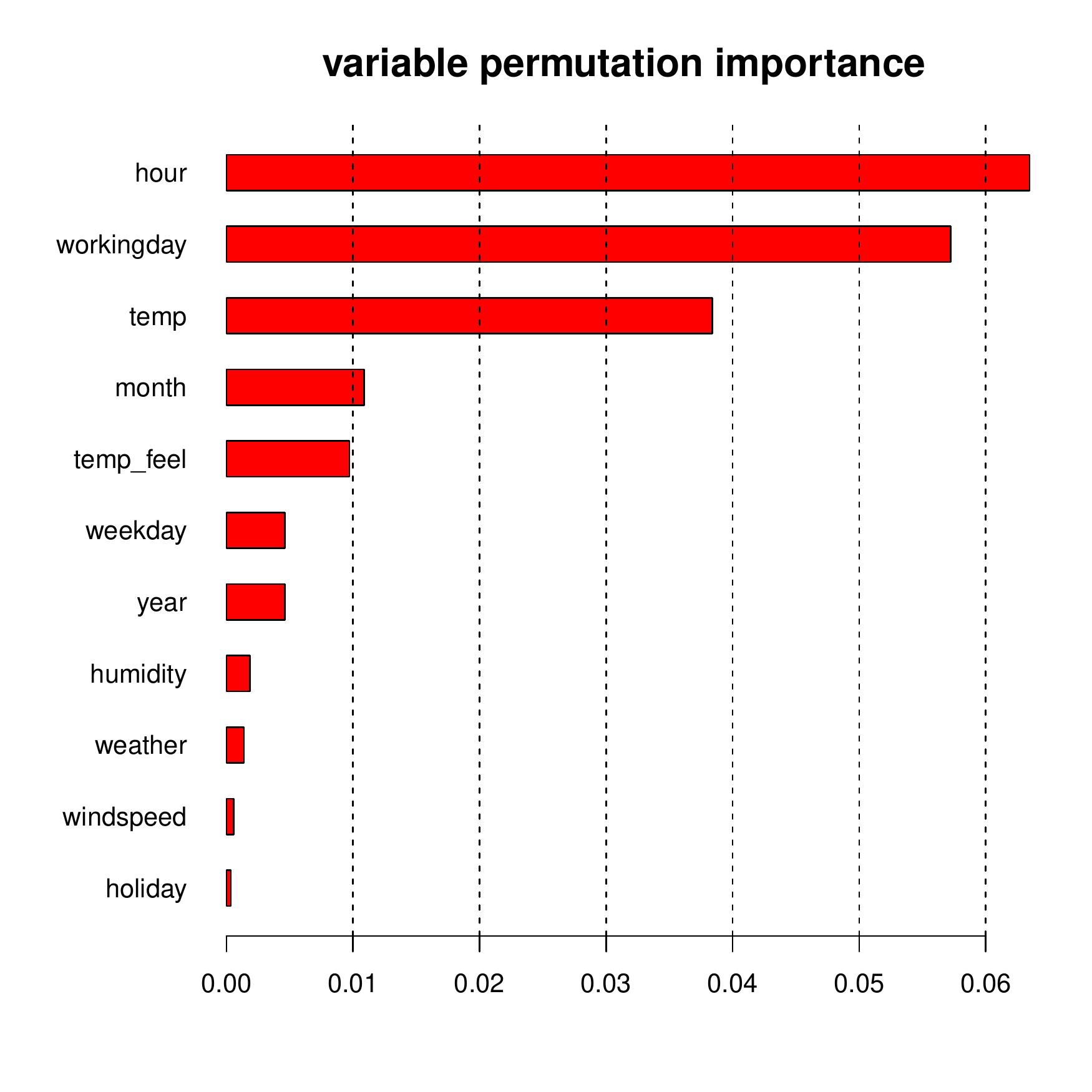}
\end{center}
\end{minipage}
\end{center}
\caption{(lhs) Empirical density of canonical parameters $(\theta(\bx_i))_{1\le i \le n}$,
(rhs) variable permutation importance.}
\label{empirical density of frequencies}
\end{figure}

Figure \ref{empirical density of frequencies} (lhs) shows the empirical density of the
canonical parameters $\bx_i \mapsto \theta(\bx_i)$ of the
fitted model over our data $1\le i \le n$. We have negative skewness in this empirical
density. A simple way of analyzing importance of feature components is to randomly permute
one component $x_j$ at a time across all records $1\le i \le n$ and study the increase 
in objective function; this is the method of variable permutation importance introduced
by  \cite{BreimanRF}. We use as objective function the Bernoulli deviance loss
which is proportional to the binary cross-entropy (also called log loss).
Figure \ref{empirical density of frequencies} (rhs) shows the
variable permutation importances. There are three variables (hour, working day and temperature)
that highly dominate the others.
Note that variable permutation importance does not properly consider the dependence
structure in $\bX$, similarly to ICEs and PDPs, because permutation of $x_j$ is done without
impacting $\bx_{\setminus j}$.

\subsection{1st and 2nd order contributions}
\begin{figure}[htb!]
\begin{center}
\begin{minipage}[t]{0.45\textwidth}
\begin{center}
\includegraphics[width=\textwidth]{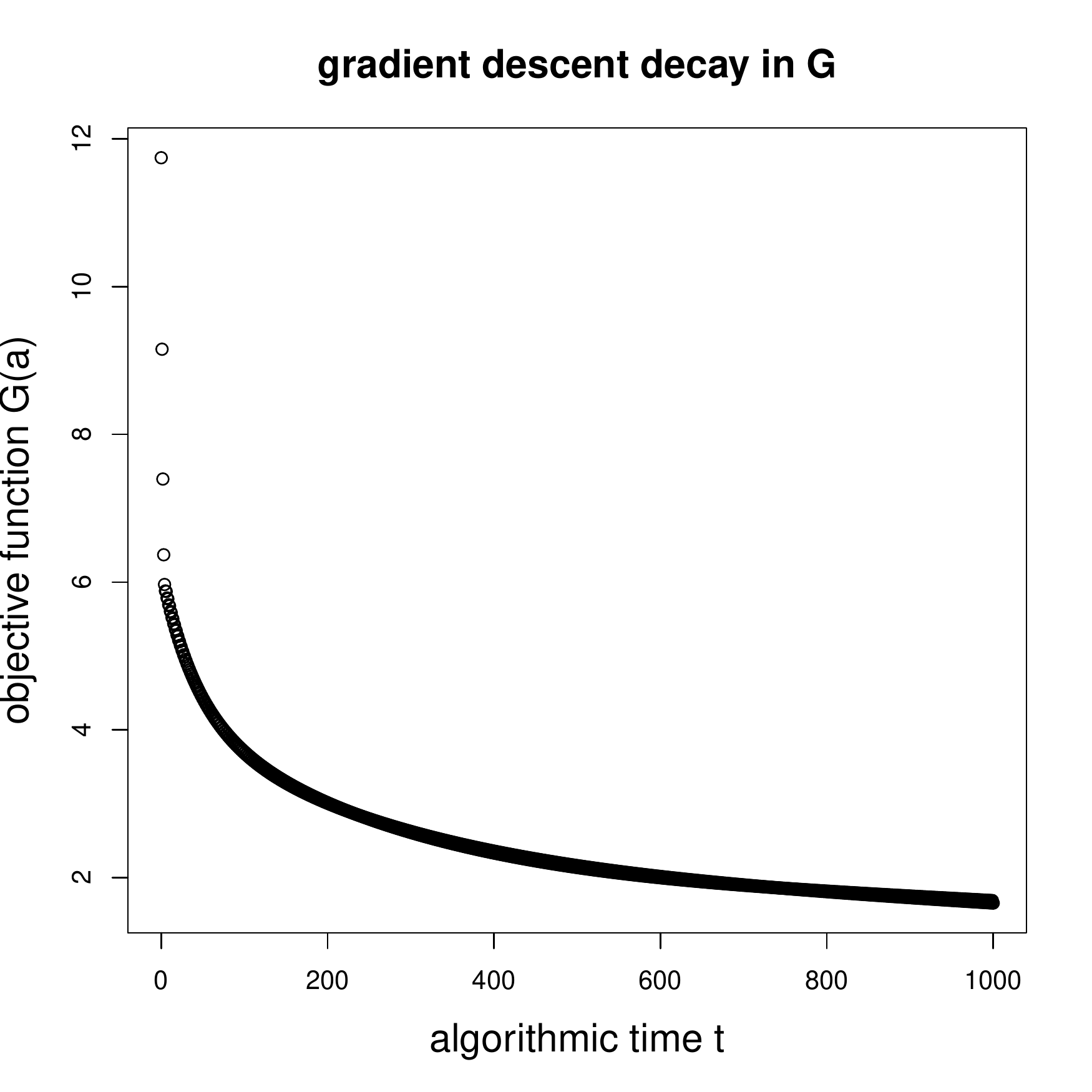}
\end{center}
\end{minipage}
\begin{minipage}[t]{0.45\textwidth}
\begin{center}
\includegraphics[width=\textwidth]{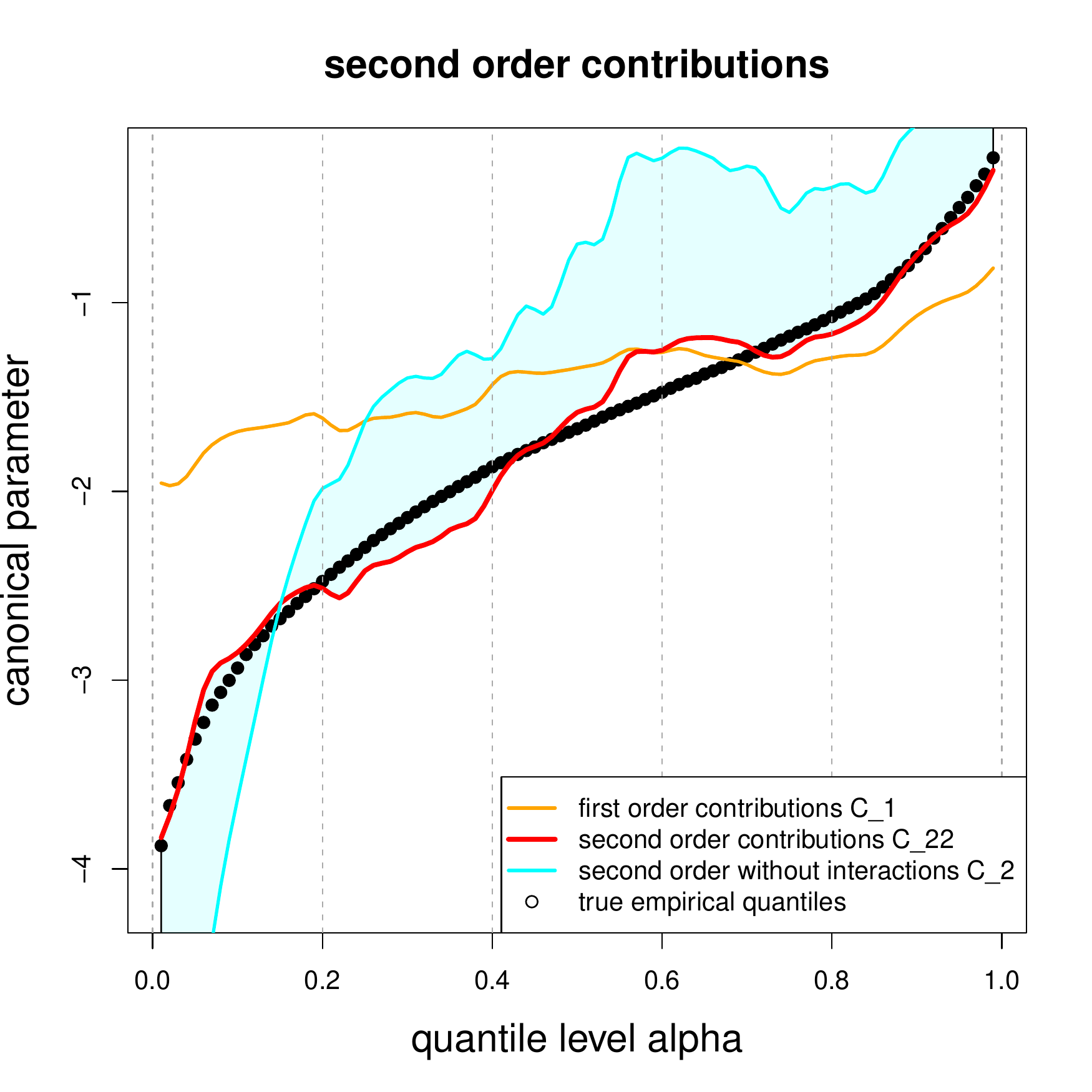}
\end{center}
\end{minipage}
\end{center}
\caption{(lhs) Gradient descent for reference point $\ba$, (rhs) 2nd order contributions \eqref{second order contributions 3}.}
\label{gradient descent for reference}
\end{figure}

The accuracy of the 2nd order contributions \eqref{second order contributions 3}
will depend on the choice of the reference point $\ba \in \R^q$. For network gradient descent
fitting we have normalized the feature components to be centered and having unit variance, i.e.~$\E_P[\bX]=\b0$
and ${\rm Var}_P(X_j)=1$ for all $1\le j \le q$.
This pre-processing is needed to efficiently apply stochastic gradient descent network
fitting. We now translate these feature components by choosing a reference
point $\ba$ such that the objective function $G(\ba;\theta)$ is minimized, see
\eqref{objective function}. We use a plain vanilla gradient descent update
\eqref{gradient descent update} using a learning rate of 
$\varepsilon_{t+1}=10^{-2}/\|\nabla_{\ba}G(\ba^{(t)};\theta)\|$.
For the quantile grid we choose $\alpha_l= l/100$ for $1\le l \le L=99$, thus, $\alpha \in \{ 1\%, \ldots, 99\%\}$.
The resulting decrease in objective function $G(\cdot;\theta)$ is plotted in Figure \ref{gradient descent for reference}.

Working with observed data, we need to discretize the MACQ analysis
for quantile levels $\{\theta(\bX)=F_{\theta(\bX)}^{-1}(\alpha)\}$, $\alpha \in (0,1)$.
We do this on a discrete grid by using a local
smoother of degree 2, in particular, we use the {\sf R} function {\tt locfit} with
parameters {\tt deg=2} and {\tt alpha=0.1} (being the chosen bandwidth)
for observations $x^{\ba}_{i,j} \theta_j(\bx_i)$ and $x^{\ba}_{i,j} x^{\ba}_{i,k} \theta_{j,k}(\bx_i)$, $1\le i \le n$,
where we set $\bx_i^{\ba}=\bx_i-\ba$.
We then fit the local smoother to these observations being ordered according to the ranks
of $\theta(\bx_i)$, to work with the corresponding empirical output quantiles. Thus, for instance, the $\ba$-adjusted
1st order attributions $S_j(\theta;\alpha_l)$, $1\le l \le L$, are estimated empirically by using the
pseudo code
\begin{equation*}
{\tt predict(locfit(}x^{\ba}_{i,j} \theta_j(\bx_i) \sim {\rm rank}(\theta(\bx_i))/n, {\tt alpha=0.1, deg=2), newdata=c(1:99)/100)}.
\end{equation*}
Figure \ref{gradient descent for reference} (rhs) gives the results after optimizing for the
reference point $\ba$. The orange color shows the 1st order contributions
$C_1=\theta(\ba) + \sum_{j=1}^q S_j( \theta; \alpha)$,
the cyan line shows the 2nd order contributions without interaction terms
$C_2=\theta(\ba) + 
\sum_{j=1}^q (S_j( \theta; \alpha)-\frac{1}{2} T_{j,j}(\theta; \alpha))$
and the red line shows the full 2nd order contributions 
$C_{2,2}=\theta(\ba) + 
\sum_{j=1}^q (S_j( \theta; \alpha)-\frac{1}{2} T_{j,j}(\theta; \alpha))
- \sum_{1\le j < k\le q} T_{j,k}(\theta; \alpha)$.

We observe from Figure \ref{gradient descent for reference} (rhs) 
that the full 2nd order contributions $C_{2,2}$ match the empirical
quantiles (black dots) quite well which explains that there is a reference point $\ba$ that
allows for suitable 2nd order approximations over the entire quantile set. 
The shaded cyan area between $C_2$ (cyan line) and $C_{2,2}$ (red line) shows the influence
of the interaction terms $T_{j,k}(\theta; \alpha)$, $j\neq k$, which illustrates that this model undergoes
substantial interactions, and a simple generalized additive model (GAM) will not be able
to model this data accurately.

\begin{figure}[htb!]
\begin{center}
\begin{minipage}[t]{0.45\textwidth}
\begin{center}
\includegraphics[width=\textwidth]{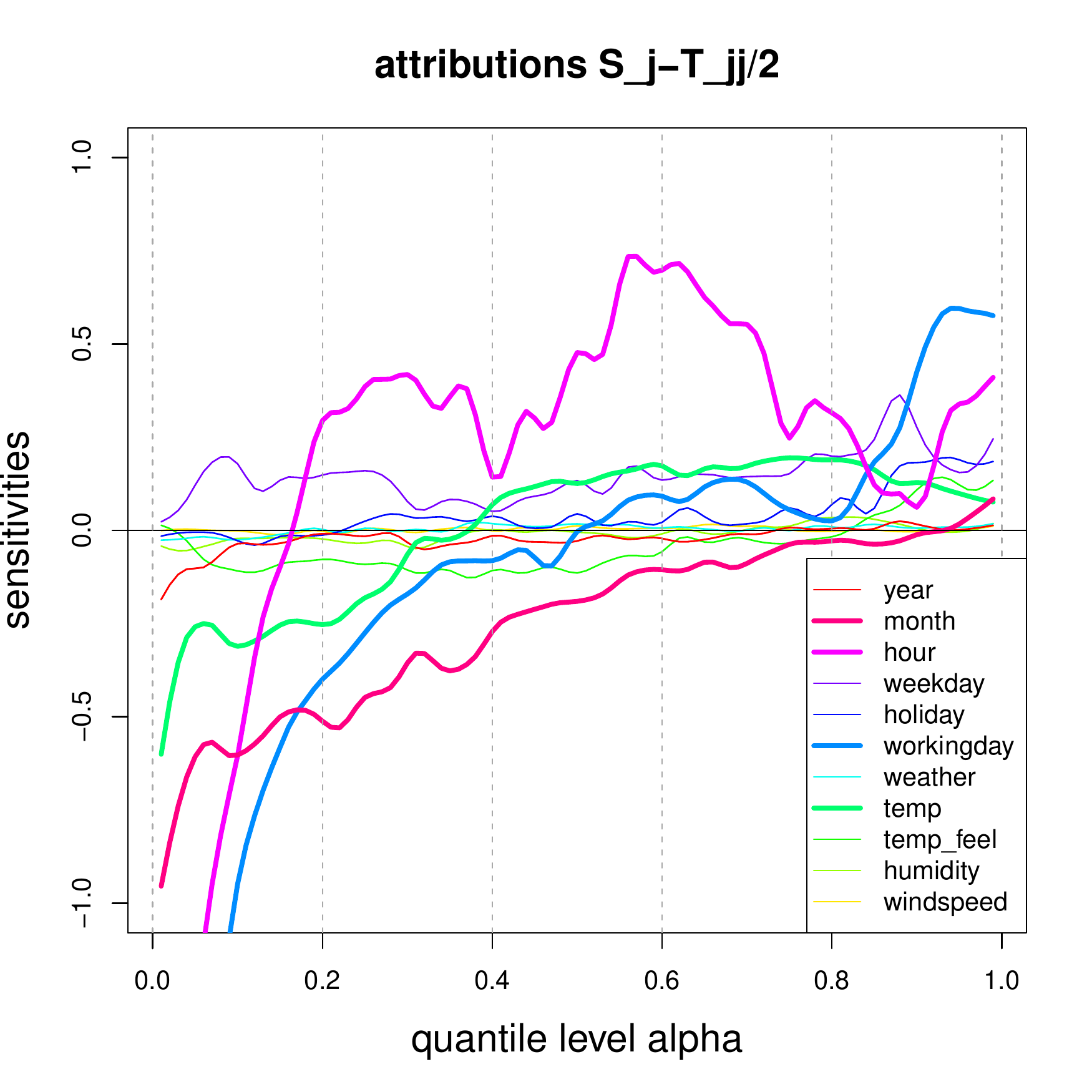}
\end{center}
\end{minipage}
\begin{minipage}[t]{0.45\textwidth}
\begin{center}
\includegraphics[width=\textwidth]{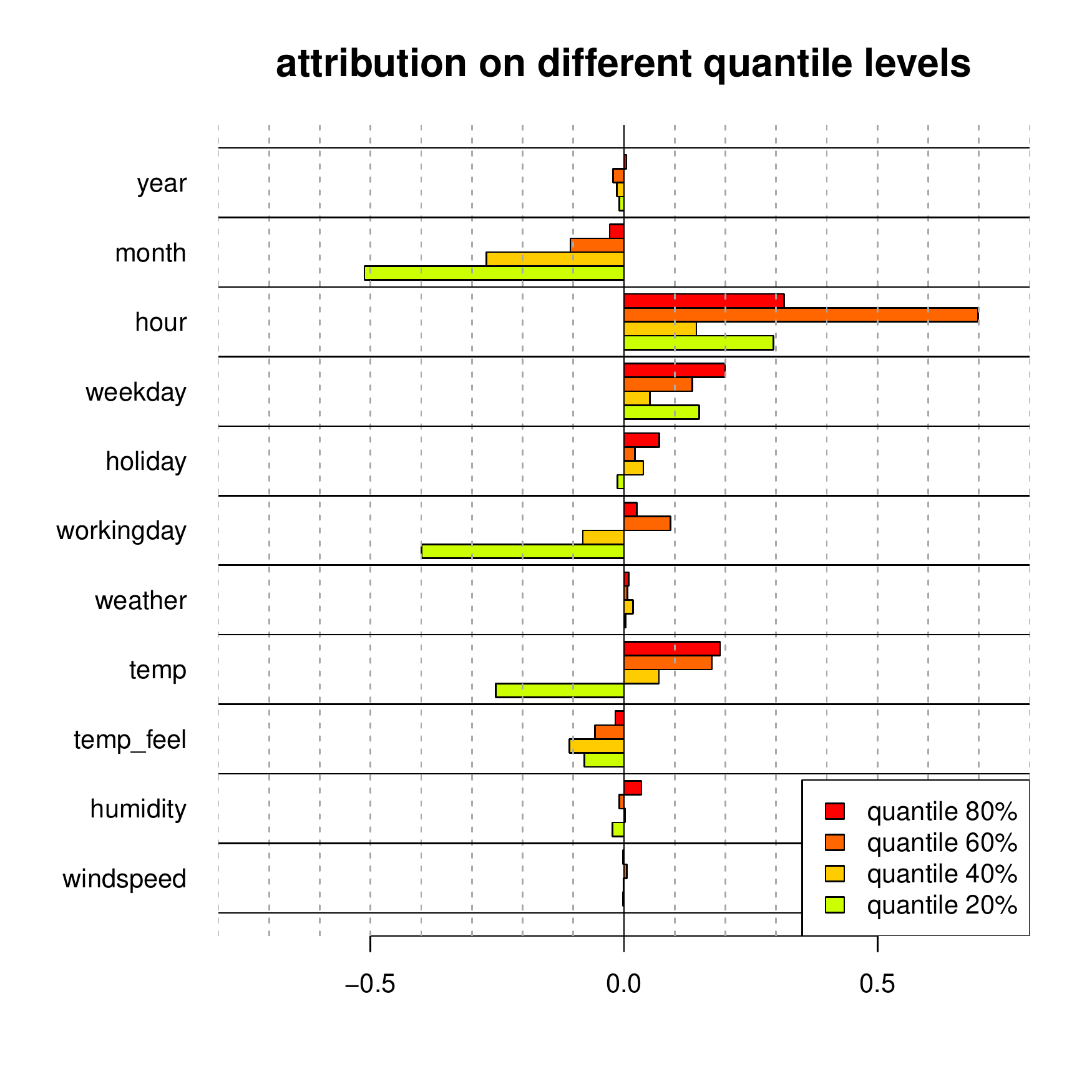}
\end{center}
\end{minipage}
\end{center}
\caption{(lhs) attributions $S_j( \theta; \alpha)-\frac{1}{2} T_{j,j}(\theta; \alpha)$
excluding interaction terms, see \eqref{second order contributions 3}, 
(rhs) attributions $S_j( \theta; \alpha)-\frac{1}{2} T_{j,j}(\theta; \alpha)$ for
selected quantile levels
$\alpha \in \{20\%, 40\%,60\%, 80\%\}$.}
\label{2nd order contributions w/o interaction}
\end{figure}

In Figure \ref{2nd order contributions w/o interaction} (lhs) we show the 
attributions $S_j( \theta; \alpha)-\frac{1}{2} T_{j,j}(\theta; \alpha)$,
excluding interaction terms $T_{j,k}(\theta; \alpha)$, $j\neq k$, 
relative to the optimal reference point $\ba$.
These attributions show the differences relative to 
canonical parameter in the reference point $\theta(\ba)$; when aggregating over $1\le j \le q$
this results in the
cyan line of Figure \ref{gradient descent for reference} (rhs).
Figure \ref{2nd order contributions w/o interaction} (lhs) shows substantial sensitivities in the
variables month, hour, working day and temperature. From this we conclude that these
are the important variables in our regression model for differentiating the responses $Y$ w.r.t.~available feature
information $\bx$. In contrast to the variable permutation
importance plot of Figure \ref{empirical density of frequencies} (rhs), this assessment correctly considers
the dependence structure within the features $\bX$.
Moreover, this plot now allows us to analyze variable importance on different
quantile levels by considering vertical slices in Figure \ref{2nd order contributions w/o interaction} (lhs).
We consider such vertical slices in Figure \ref{2nd order contributions w/o interaction} (rhs) for four 
selected quantile levels 
$\alpha \in \{20\%, 40\%, 60\%, 80\%\}$. 
We observe that the variables {\tt month}, {\tt hour}, {\tt workingday} and {\tt temp} undergo the biggest
changes when moving from small quantiles to big ones. The quantile level at 20\% can be explained by
the three features {\tt temp}, {\tt month} and {\tt workingday}, whereas for the quantile level 
at 60\% has {\tt hour} (daytime) as an important variable, see Figure \ref{2nd order contributions w/o interaction} (rhs). Note that this is not the full
picture, yet, as we do not consider interactions in these vertical slices; the importance of interactions
is indicated by the cyan shaded area in Figure \ref{gradient descent for reference} (rhs) for 
different quantile levels.

\begin{figure}[htb!]
\begin{center}
\begin{minipage}[t]{0.24\textwidth}
\begin{center}
\includegraphics[width=\textwidth]{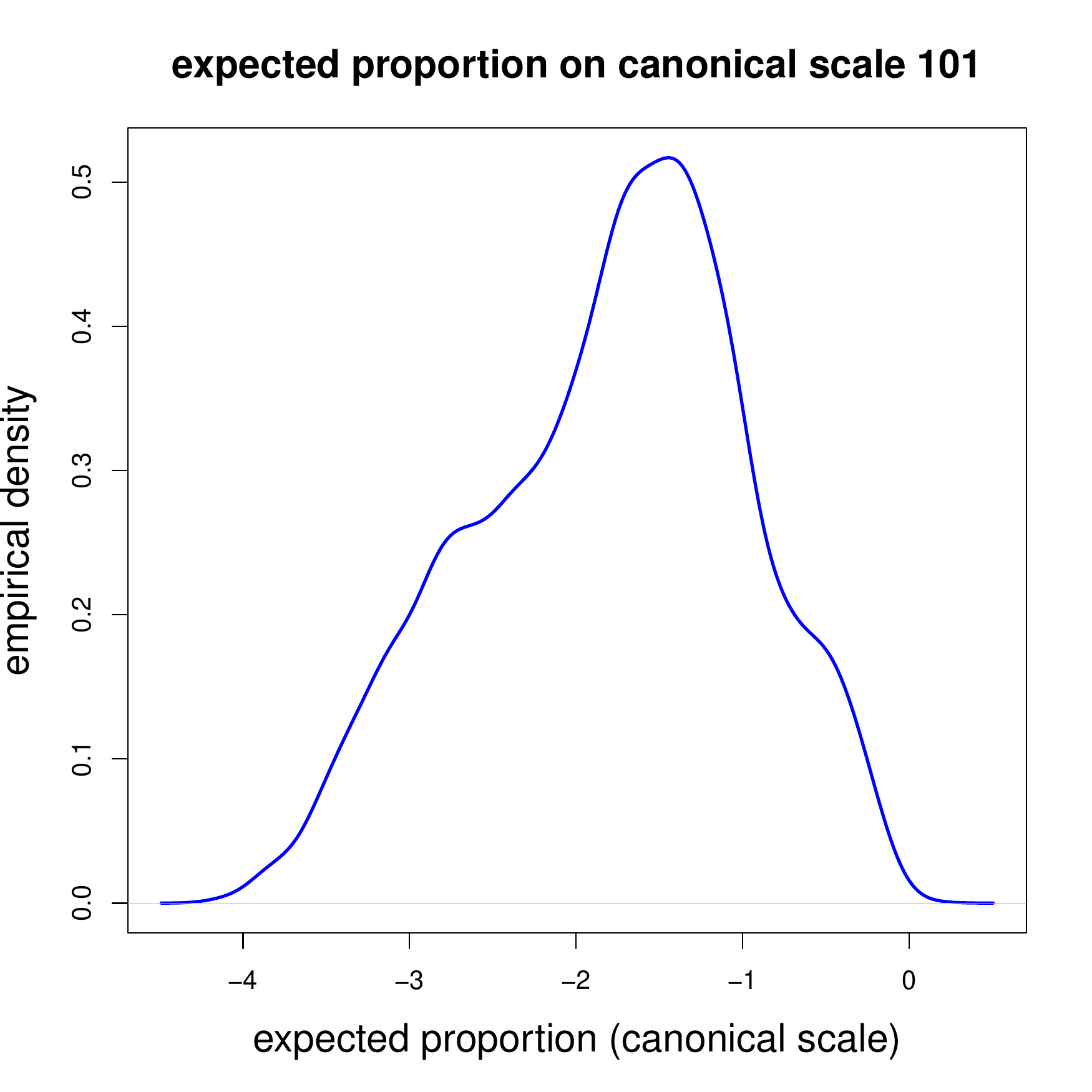}
\end{center}
\end{minipage}
\begin{minipage}[t]{0.24\textwidth}
\begin{center}
\includegraphics[width=\textwidth]{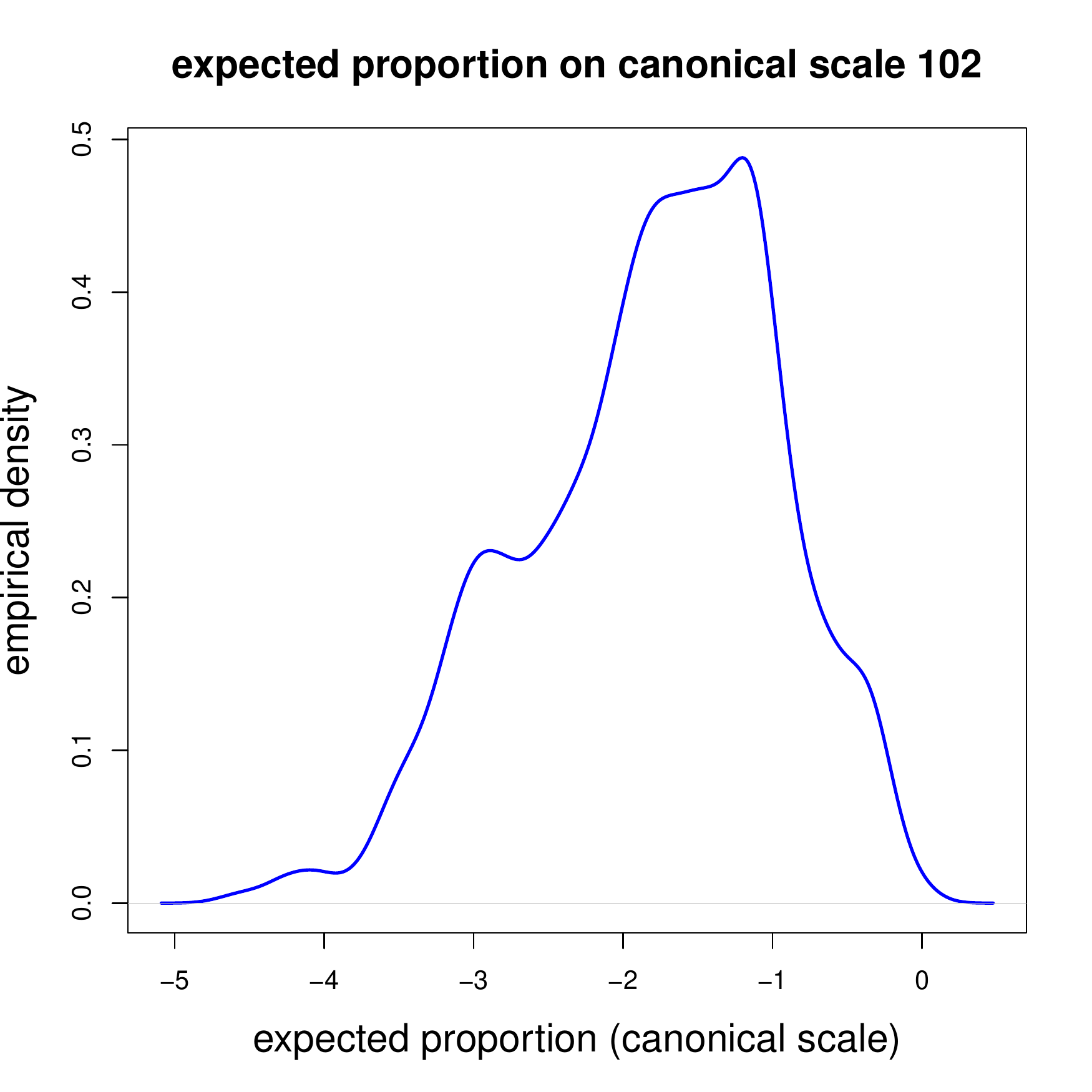}
\end{center}
\end{minipage}
\begin{minipage}[t]{0.24\textwidth}
\begin{center}
\includegraphics[width=\textwidth]{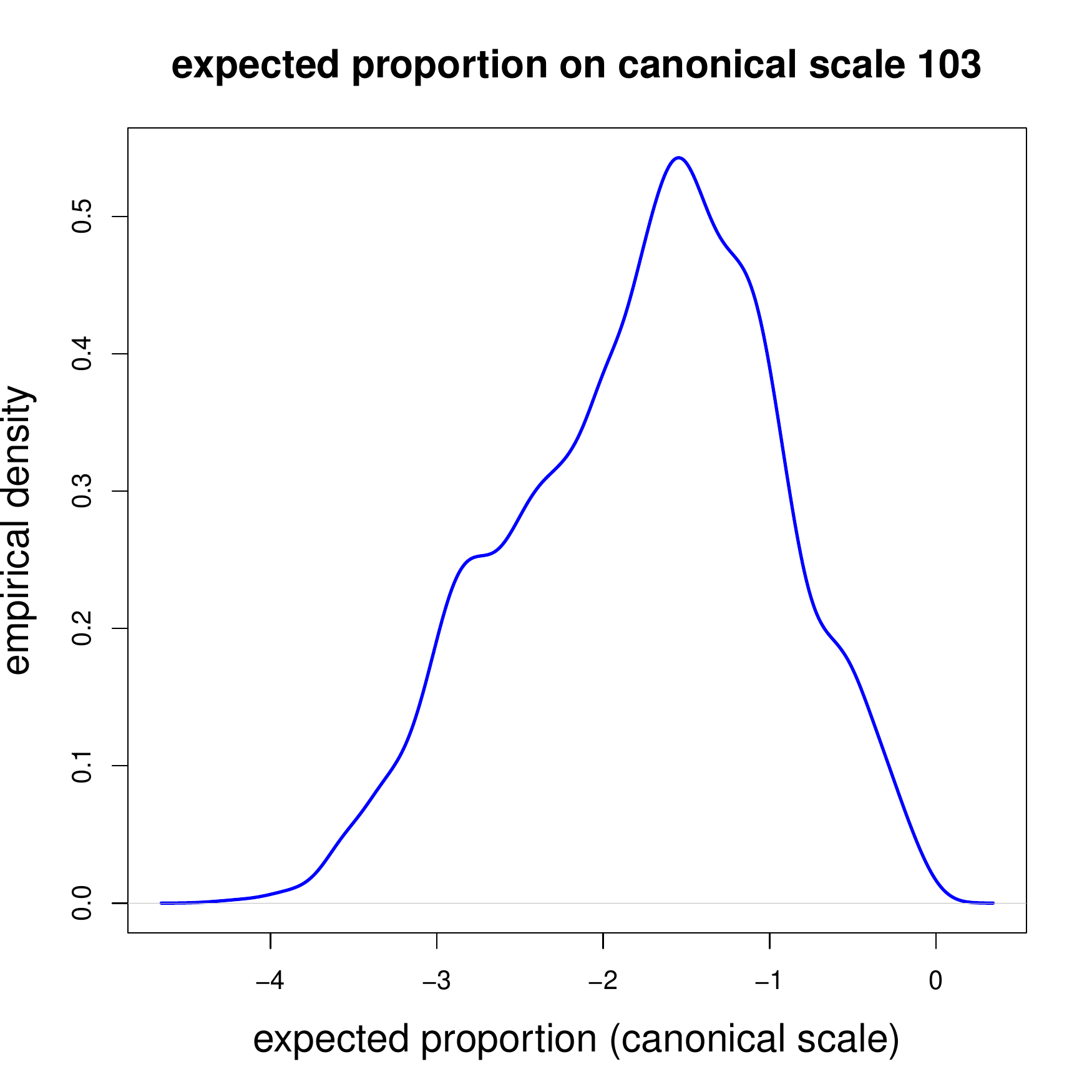}
\end{center}
\end{minipage}
\begin{minipage}[t]{0.24\textwidth}
\begin{center}
\includegraphics[width=\textwidth]{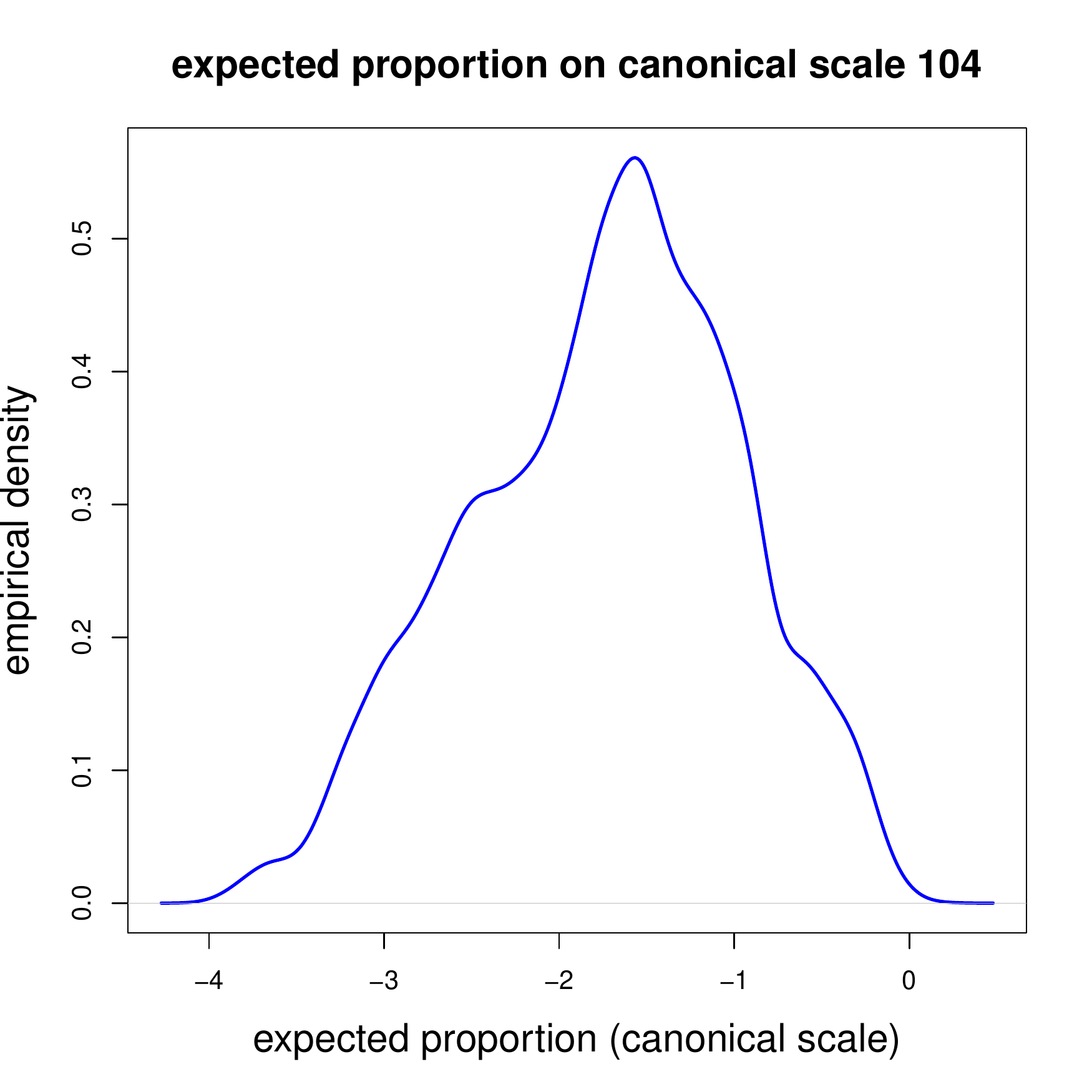}
\end{center}
\end{minipage}

\begin{minipage}[t]{0.24\textwidth}
\begin{center}
\includegraphics[width=\textwidth]{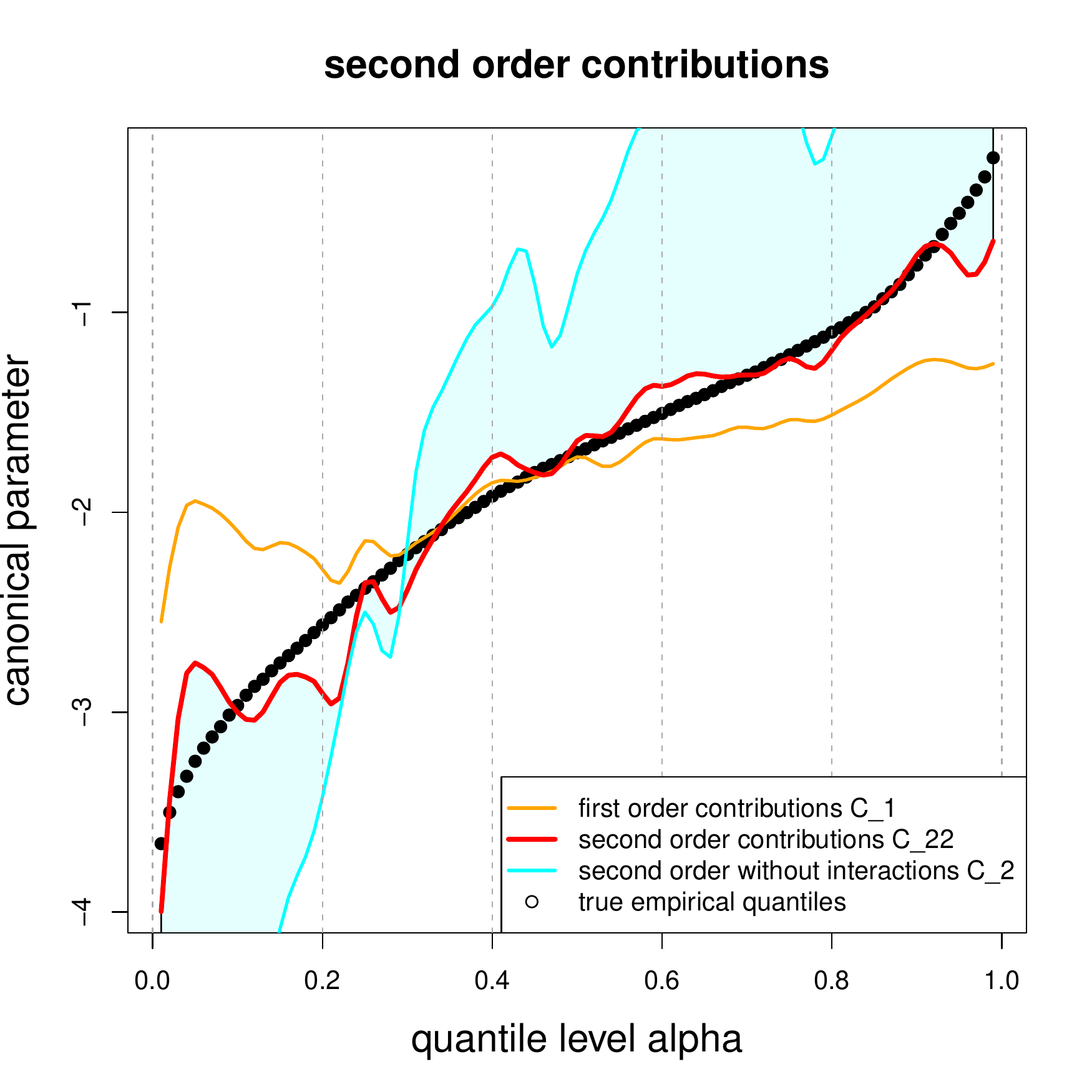}
\end{center}
\end{minipage}
\begin{minipage}[t]{0.24\textwidth}
\begin{center}
\includegraphics[width=\textwidth]{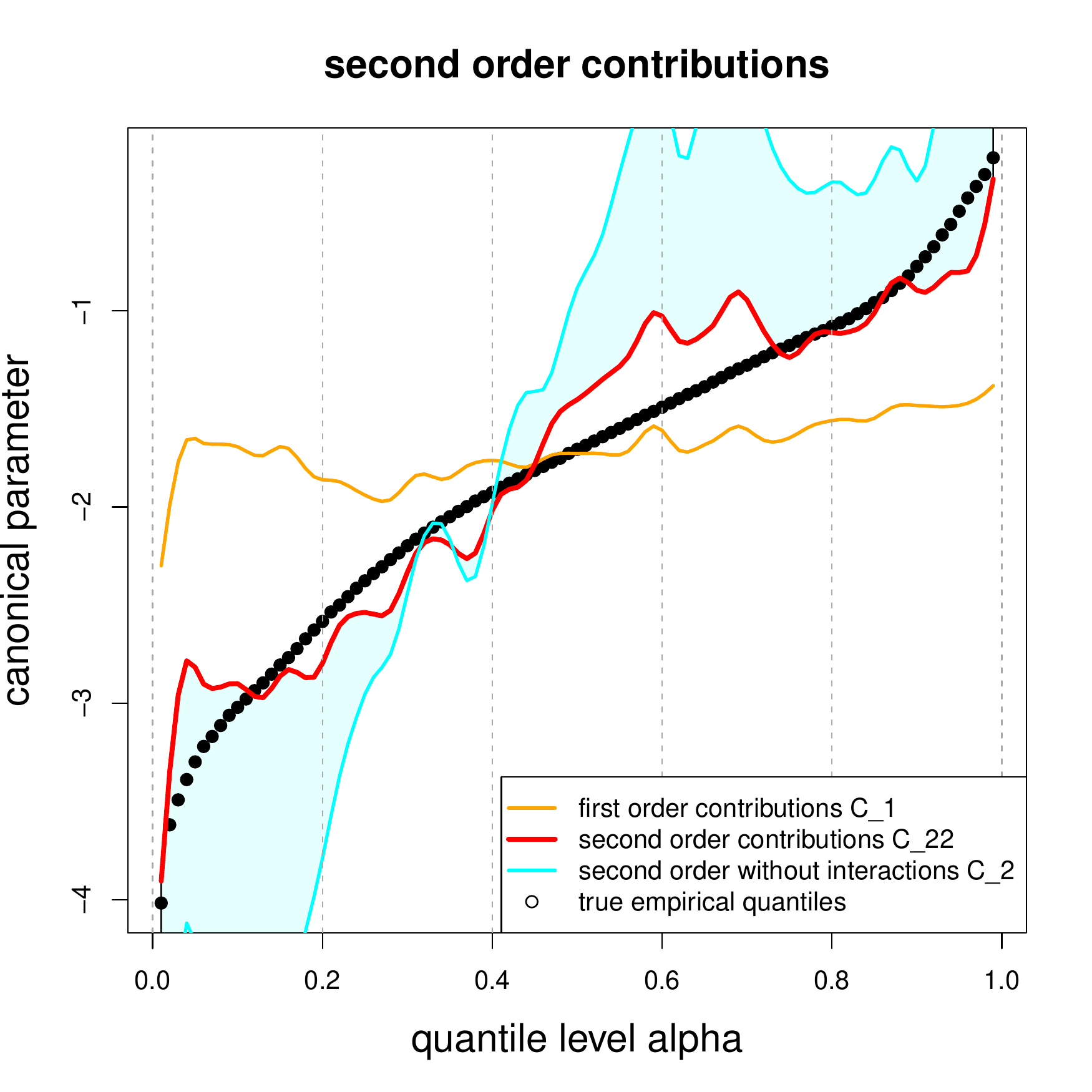}
\end{center}
\end{minipage}
\begin{minipage}[t]{0.24\textwidth}
\begin{center}
\includegraphics[width=\textwidth]{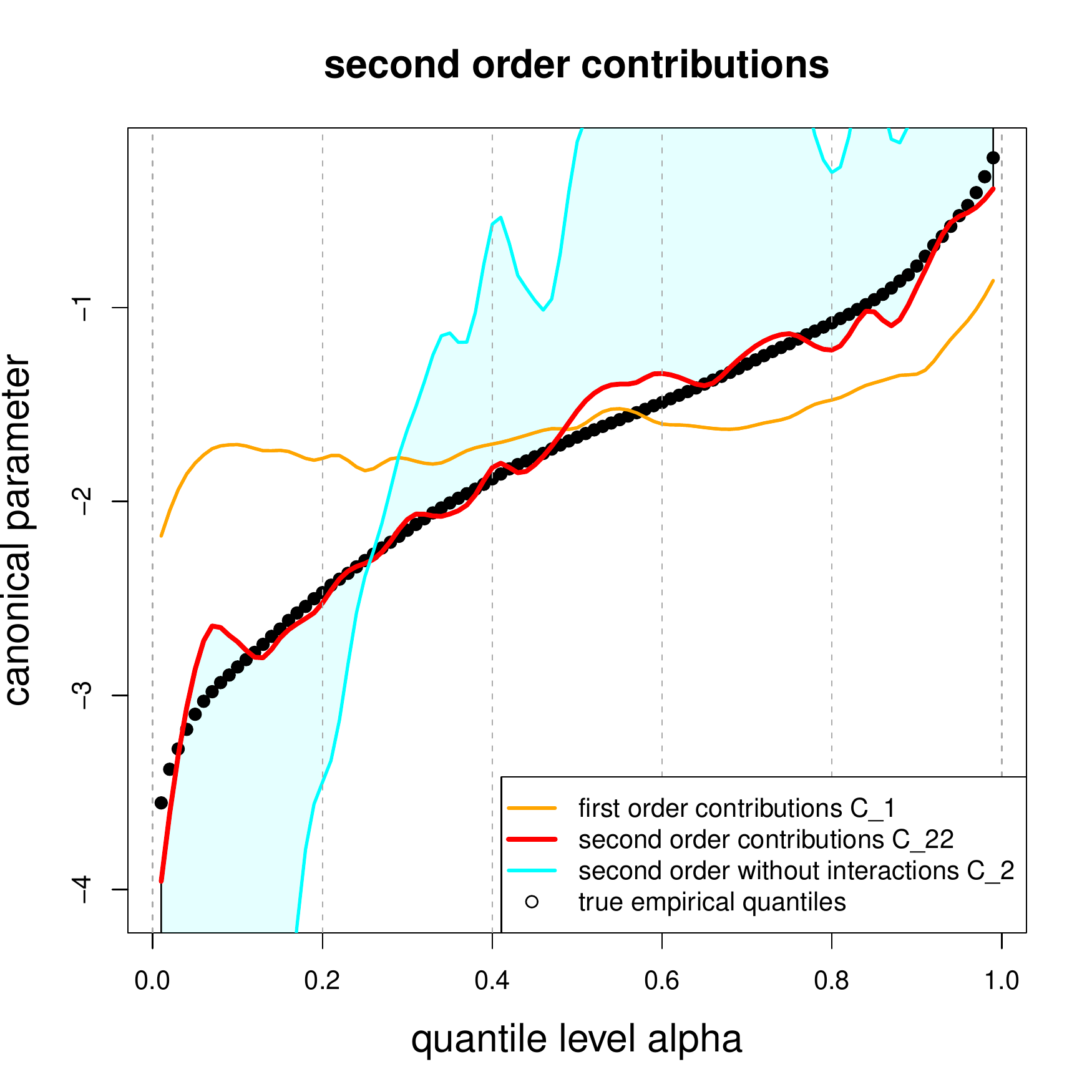}
\end{center}
\end{minipage}
\begin{minipage}[t]{0.24\textwidth}
\begin{center}
\includegraphics[width=\textwidth]{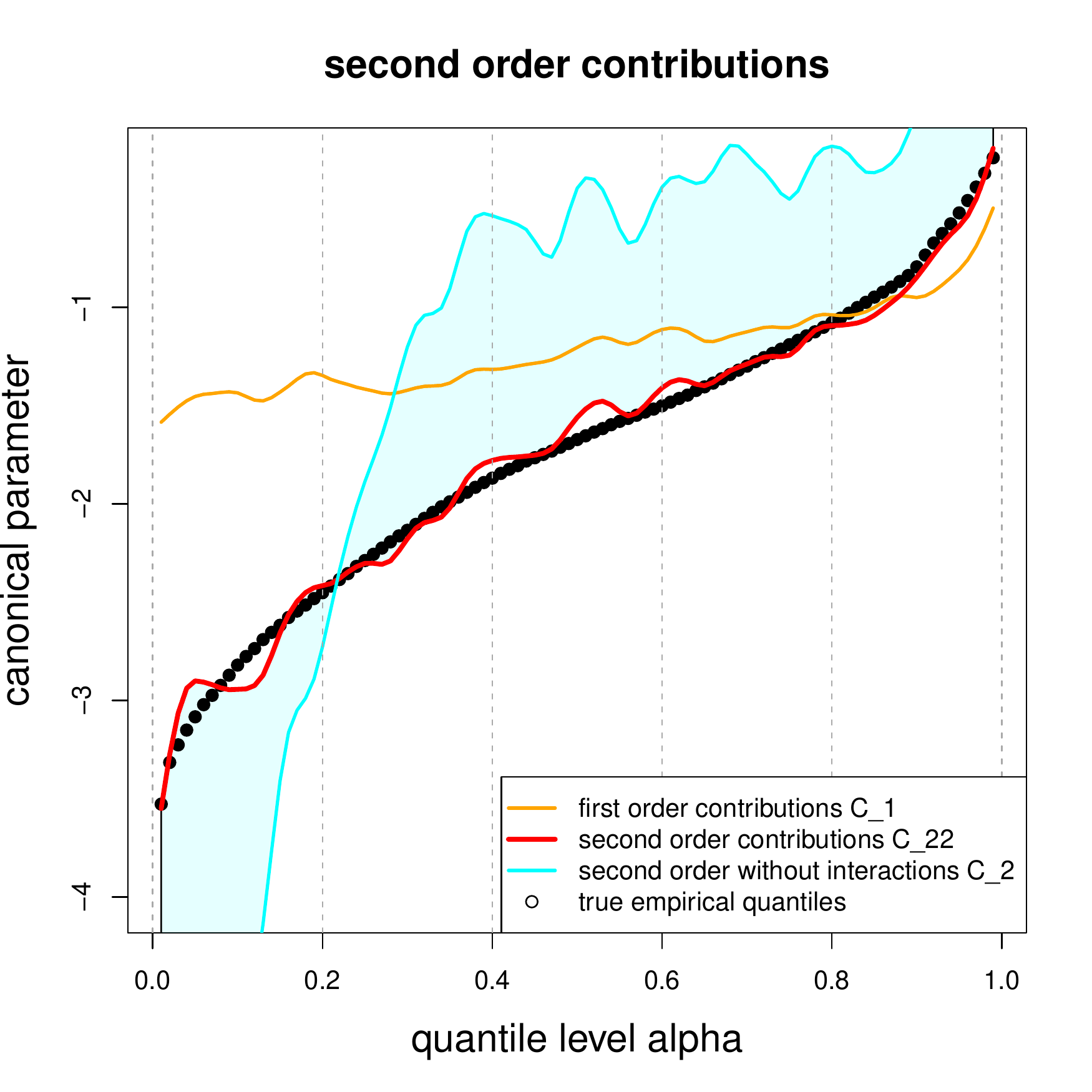}
\end{center}
\end{minipage}

\end{center}
\caption{Robustness of 2nd order contributions across 4 different networks: (top row) empirical
densities of canonical parameters $(\theta(\bx_i))_{1\le i \le n}$, (bottom row) 
2nd order contributions \eqref{second order contributions 3}.}
\label{robustness of results}
\end{figure}

In Figure \ref{robustness of results} we analyze the robustness of the attribution
results. We do this by considering different networks  $\bx \mapsto \theta(\bx)$ for predicting
the response variable $Y$. Network regression models lack a certain degree of robustness as gradient
descent network fitting explores different (local) minima of the objective function; note that,
in general, neural network fitting is not a convex minimization problem. This issue of non-uniqueness
of good predictive models has been widely discussed in the literature, and ensembling may be one
solution to mitigate this problem, we refer to \cite{Dietterich1, Dietterich2}, \cite{Zhou2},
\cite{Zhou1} and \cite{RichmanW2}. The top row shows the empirical distributions of 
the canonical parameters $(\theta(\bx_i))_{1\le i \le n}$ for 4 different networks; we observe
that there are some differences in these empirical densities. The bottom row shows the corresponding 2nd order 
contributions \eqref{second order contributions 3}, split by 1st order contributions $C_1$,
2nd order contributions without interactions $C_2$ and the full 2nd order contributions $C_{2,2}$.
At this level, we judge the attributions made to be rather robust over the different models, the general
shapes of these graphs being similar, and also the interaction terms $C_{2,2}-C_{2}$ showing a similar
structure and magnitude across the 4 different network models.

\begin{figure}[htb!]
\begin{center}
\begin{minipage}[t]{0.45\textwidth}
\begin{center}
\includegraphics[width=\textwidth]{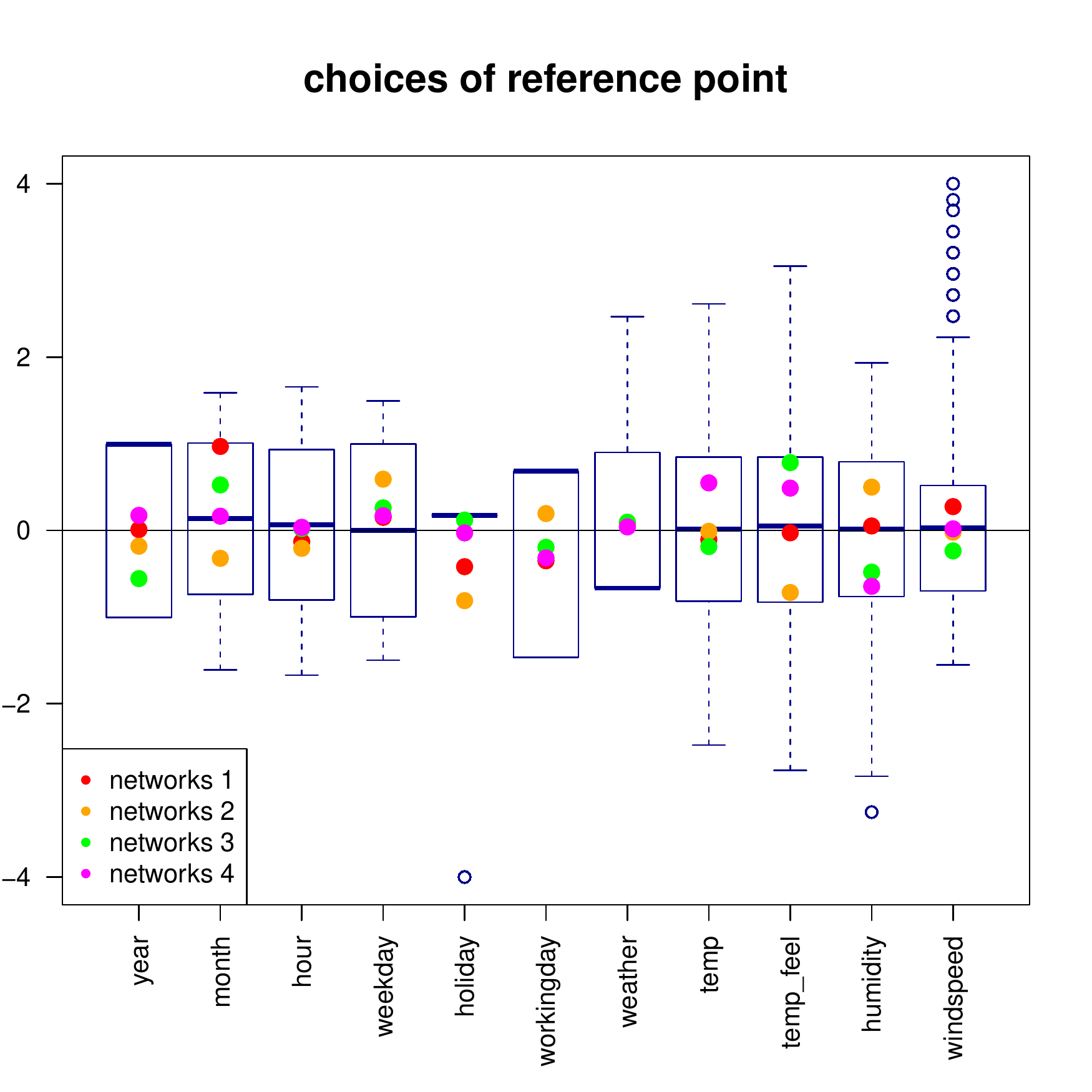}
\end{center}
\end{minipage}
\end{center}
\caption{Choice of reference point $\ba$ across 4 different networks illustrated
for all feature components $1\le j \le q$.}
\label{choices of reference points}
\end{figure}

From Figure \ref{robustness of results} we also observe that the 1st order
contributions $C_1$ intersect the quantiles  $F_{\theta(\bX)}^{-1}(\alpha)$
at different levels for the 4 different calibrations. This indicates that the optimal reference
point $\ba$ is chosen differently in the different networks. Figure \ref{choices of reference points}
shows the chosen reference points $\ba$ in relation to the features $(\bx_i)_{1\le i \le n}$; as explained above,
we have centered and normalized the feature components for gradient descent network fitting. The boxplots
in Figure \ref{choices of reference points} show these centered and normalized features in comparison
to the reference points of the 4 different networks. Some feature components have a very skewed
distribution as can be seen from the thicker horizontal boxplot lines showing the median
of each feature component $(x_{i,j})_{1\le i \le n}$, $1\le j \le q$. The reference point mostly
lies within the interquartile range (IQR). 
\begin{rem}\normalfont
The feature components of $\bx$ need pre-processing in order to be suitable for
gradient descent fitting. Continuous and binary variables have been centered and normalized
so that their gradients live in a similar range. This makes gradient descent fitting more
efficient because all partial derivatives of the gradient are directly comparable. 
Our example does not have categorical feature components. Categorical feature components
can be treated in different ways. For our MACQ proposal we envisage two different treatments. Firstly, dummy coding
could be used. This requires the choice of a reference level, and considers all other levels relative to this
reference level. The resulting marginal attributions should then be interpreted as differences to the
reference level. Secondly, one can use embedding layers for categorical variables, see
\cite{Bengio2003} and \cite{Guo}. In that case the attribution analysis can directly be done
on these learned embeddings of categorical levels, in complete analogy to the continuous variables.
\end{rem}

\subsection{Attribution to individual instances}
Next, we focus on individual instances $\bx^{\ba}_i=\bx_i-\ba$ and study individual marginal contributions
$\omega_{i,j}=(x_{i,j}-a_j) \theta_j(\bx_i)-(x_{i,j}-a_j)^2 \theta_{j,j}(\bx_i)/2$ 
to attribution $S_j(\theta; \alpha)-T_{j,j}(\theta; \alpha)/2$.

\begin{figure}[htb!]
\begin{center}
\begin{minipage}[t]{0.45\textwidth}
\begin{center}
\includegraphics[width=\textwidth]{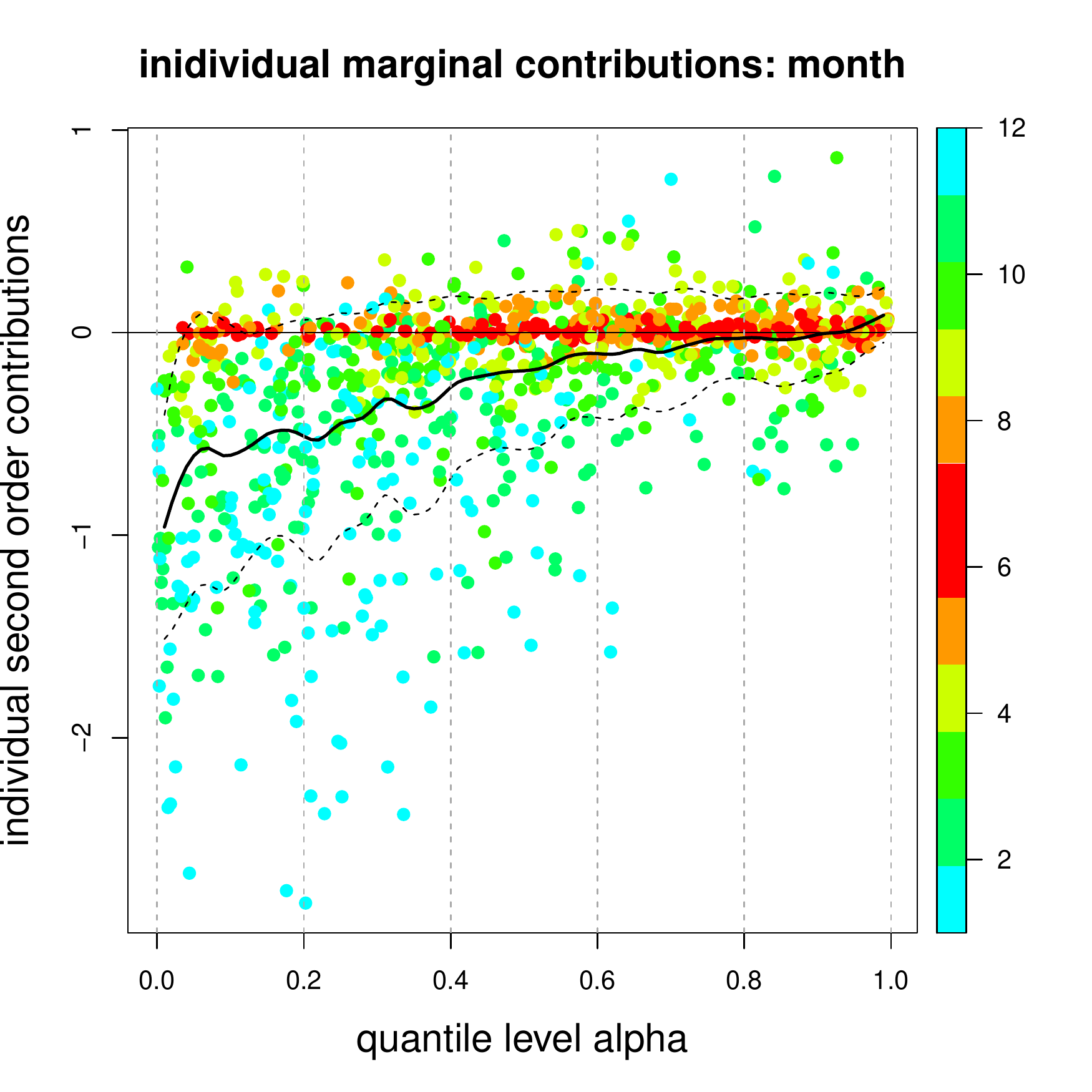}
\end{center}
\end{minipage}
\begin{minipage}[t]{0.45\textwidth}
\begin{center}
\includegraphics[width=\textwidth]{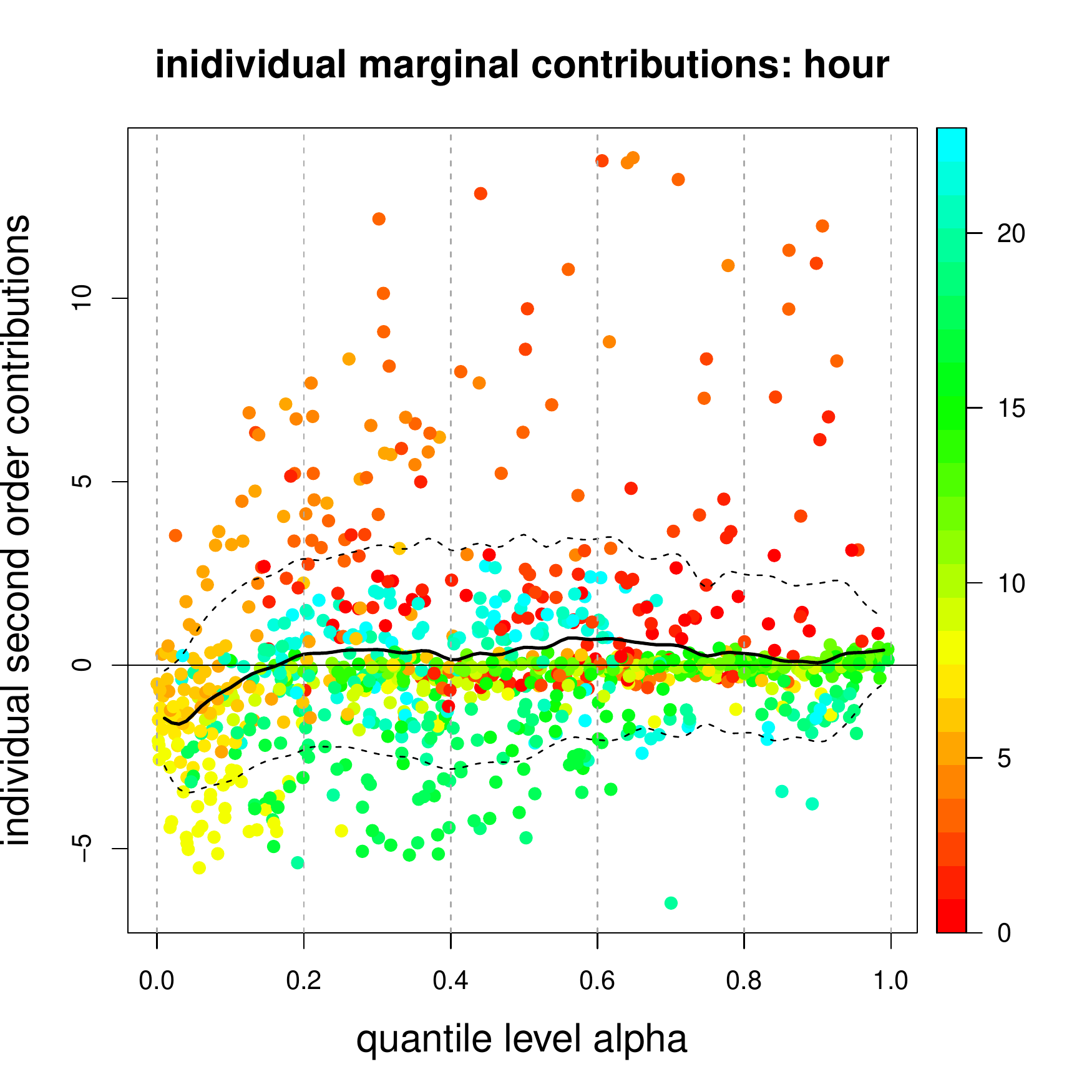}
\end{center}
\end{minipage}
\begin{minipage}[t]{0.45\textwidth}
\begin{center}
\includegraphics[width=\textwidth]{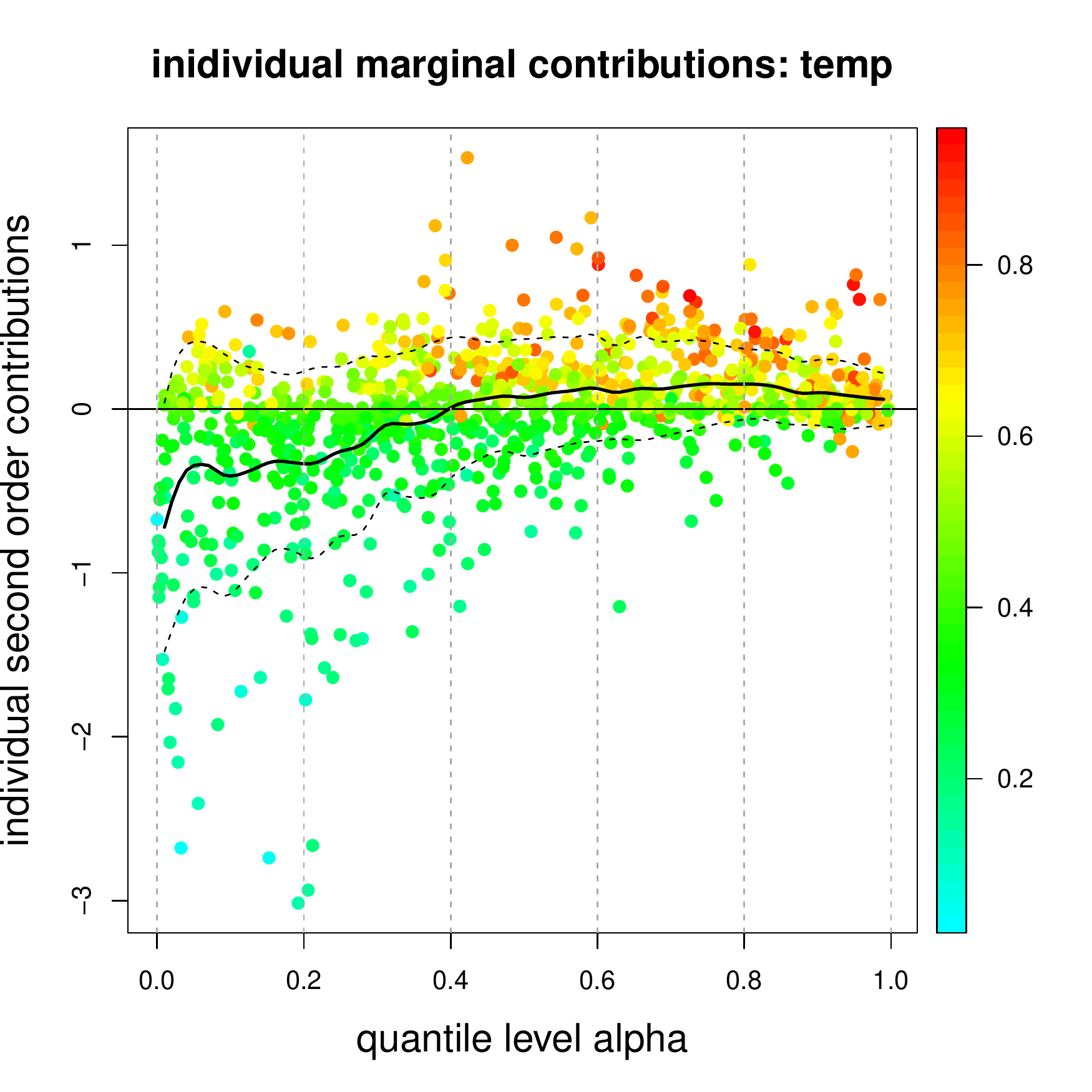}
\end{center}
\end{minipage}
\begin{minipage}[t]{0.45\textwidth}
\begin{center}
\includegraphics[width=\textwidth]{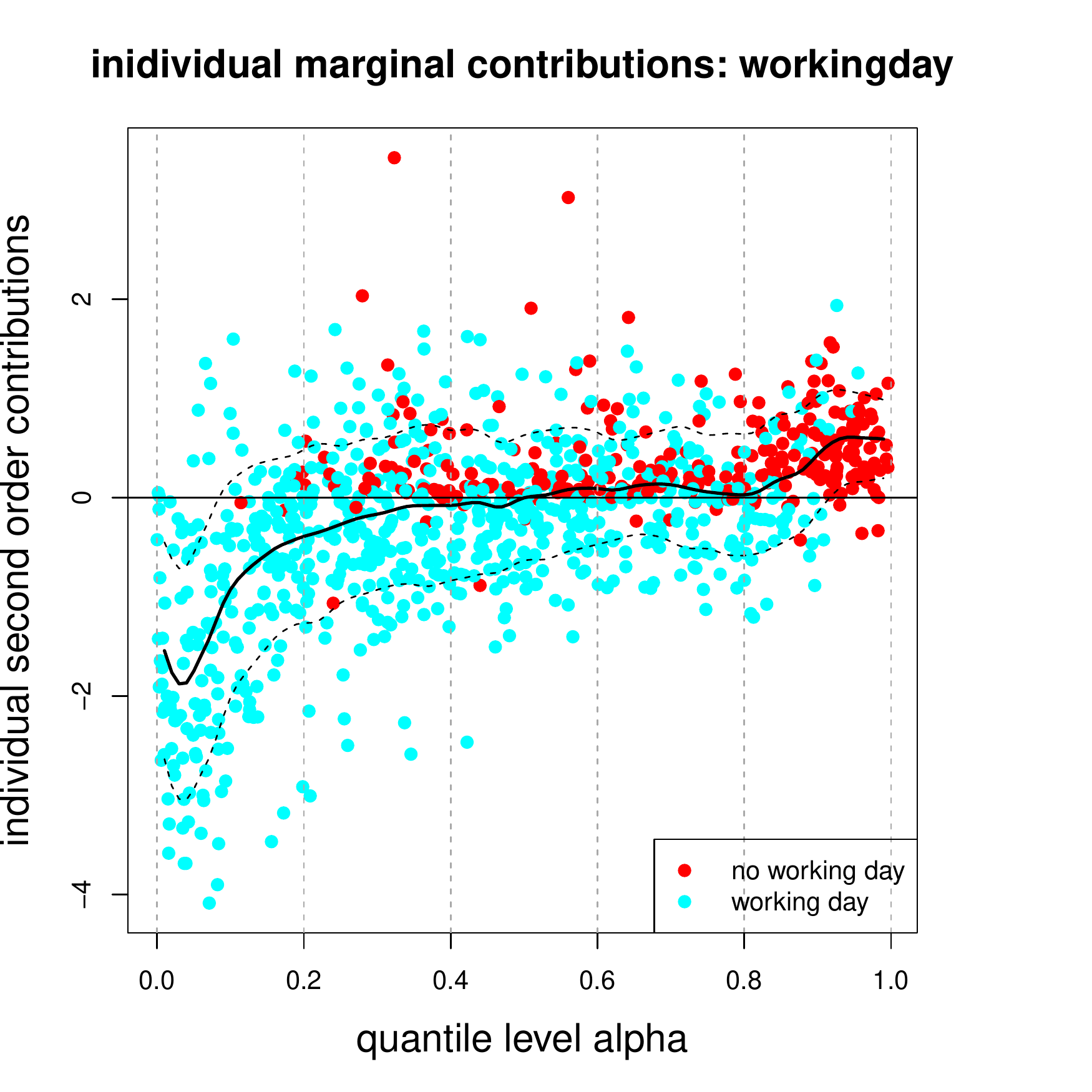}
\end{center}
\end{minipage}
\end{center}
\caption{Individual  marginal contributions $\omega_{i,j}$ of 1,000 randomly selected instances $\bx_i$
for (top-left) $j={\tt month}$, (top-right) $j={\tt hour}$, 
(bottom-left) $j={\tt temp}$ and (bottom-right) $j={\tt workingday}$; the black line shows
attribution $S_j(\theta; \alpha)-T_{j,j}(\theta; \alpha)/2$ and the black dotted line
gives one standard deviation; the $y$-scales differs in the plots and the colors illustrate the feature values $x_j$.
}
\label{individual sensitivities}
\end{figure}

For Figure \ref{individual sensitivities} we
select at random 1,000 different instances, and plot their individual marginal contributions
$\omega_{i,j}$ 
to the attributions $S_j(\theta; \alpha)-T_{j,j}(\theta; \alpha)/2$ (black solid line). The ordering on the
$x$-axis for the selected instances $\bx_i$ is obtained by considering the empirical quantiles of
the responses $\theta(\bx_k)$ over all instances $1\le k \le n$. 
We start with Figure \ref{individual sensitivities} (bottom-right) which
shows the binary variable {\tt workingday}. This variable clearly differentiates low from high
quantiles $F_{\theta(\bX)}^{-1}(\alpha)$, showing that the casual rental proportion $Y$ is
in average bigger for non-working days
(red dots). Moreover,
for low quantiles levels the working day variable clearly lowers (expected response) $\theta(\bx)$ compared
to the reference level $\theta(\ba)$, as the cyan dots are below the horizontal black line at 0 which
corresponds to the reference level.
In addition to the average attributions
$S_j(\theta; \alpha)-T_{j,j}(\theta; \alpha)/2$ (black solid line), the plot is complemented by
black dotted lines giving one (empirical) standard deviation
\begin{equation*}
{\rm Var}_P \left((X_j-a_j) \theta_j(\bX)-(X_j-a_j)^2 \theta_{j,j}(\bX)/2 \left| \theta(\bX)=F_{\theta(\bX)}^{-1}(\alpha)\right.\right)^{1/2}.
\end{equation*}
The sizes of these standard deviations quantify the heterogeneity in the individual marginal
contributions $\omega_{i,j}$. 
This can either be because
of heterogeneity of the portfolio $x_{i,j}$ on a certain quantile level, or because we have a rough 
regression surface implying heterogeneity in derivatives $\theta_j(\bx_i)$ and $\theta_{j,j}(\bx_i)$.

Next, we study the variable {\tt temp} of Figure \ref{individual sensitivities} (bottom-left).
In this plot we see a clear positive dependence between quantile levels and temperature, showing
that casual rentals are generally low for low temperatures, which can either be the calendar season
or bad weather conditions. We have clearly more heterogeneity in features (and resulting 
derivatives $\theta_j(\bx_i)$ and $\theta_{j,j}(\bx_i)$) contributing to low quantile levels than to higher ones. 
The variable {\tt temp} is highly correlated with calendar month, and the calendar month
plot in Figure \ref{individual sensitivities} (top-left) looks similar, saying that casual rental proportions $Y$
are negatively impacted by winter seasons. There are some low proportions, though, also for summer months, these need
to be explained by other variables, e.g., they may correspond to a rainy day or to a specific daytime.
The interpretation of the variable {\tt hour} in  Figure \ref{individual sensitivities} (top-right) is slightly more 
complicated since we do not have monotonicity of $\theta(\bx)$ in this variable, see 
also Figure \ref{descriptive counts 2}. Nevertheless we also see a separation between working and leisure
times (for the time-being ignoring interactions with holidays and weekends).

\begin{figure}[htb!]
\begin{center}
\begin{minipage}[t]{0.32\textwidth}
\begin{center}
\includegraphics[width=\textwidth]{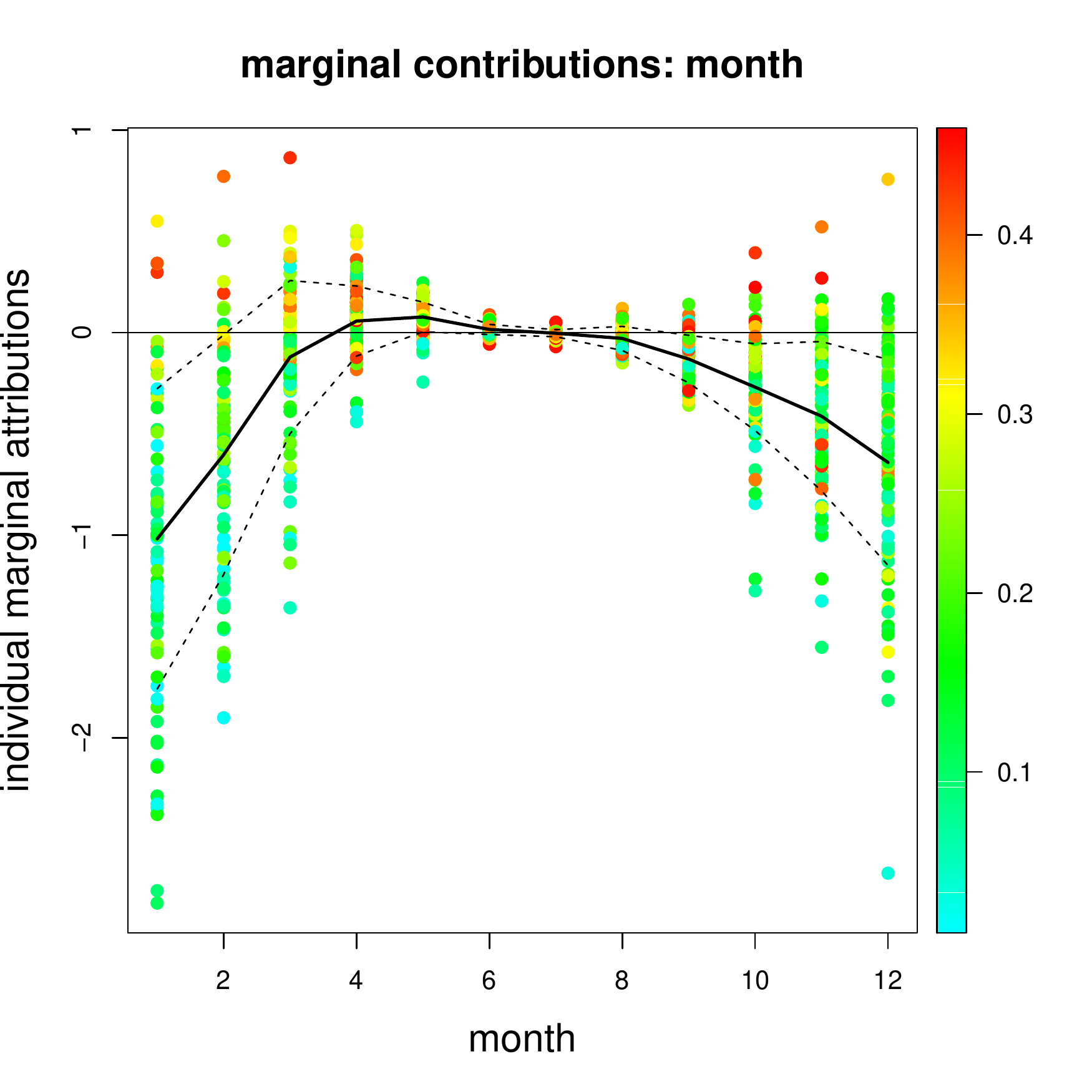}
\end{center}
\end{minipage}
\begin{minipage}[t]{0.32\textwidth}
\begin{center}
\includegraphics[width=\textwidth]{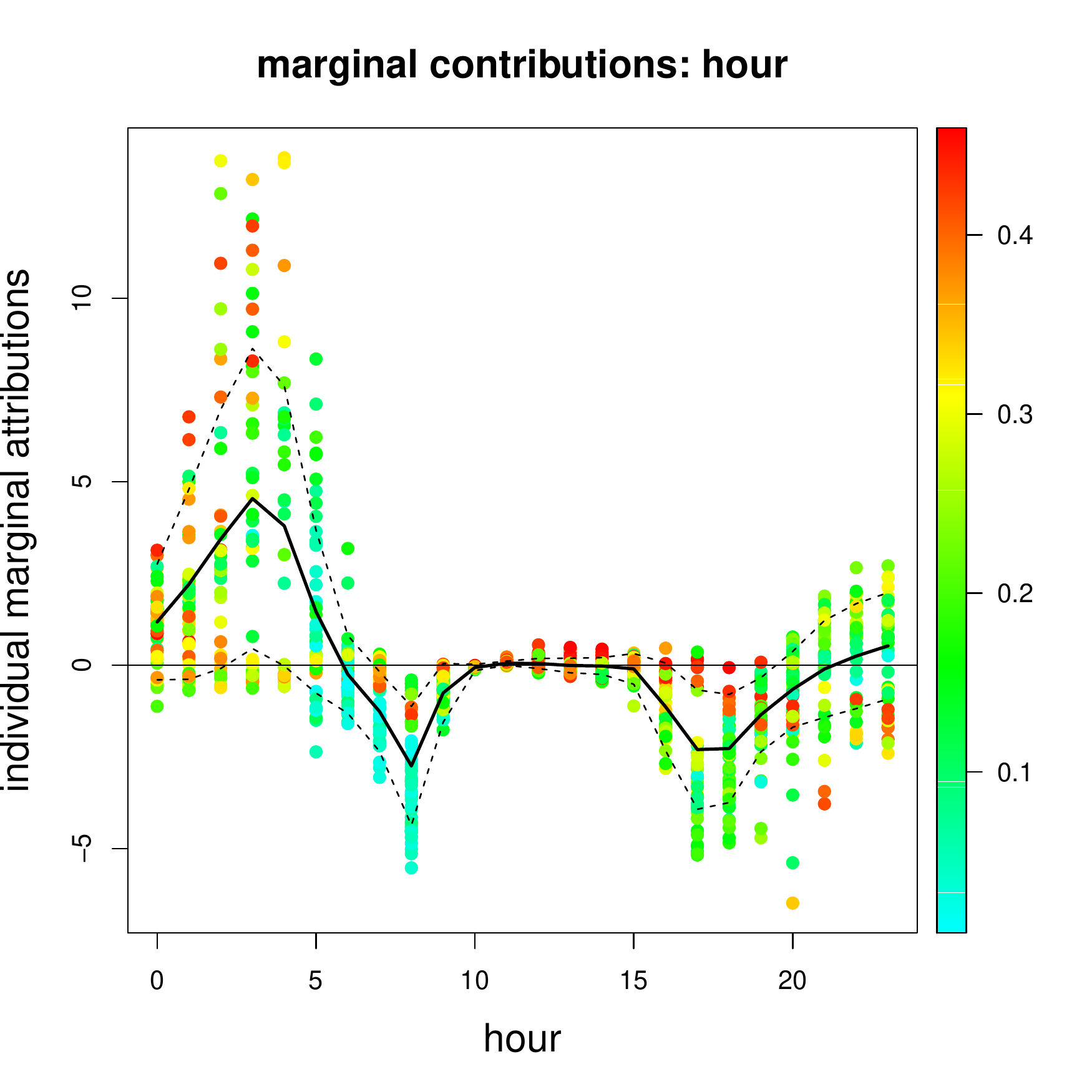}
\end{center}
\end{minipage}
\begin{minipage}[t]{0.32\textwidth}
\begin{center}
\includegraphics[width=\textwidth]{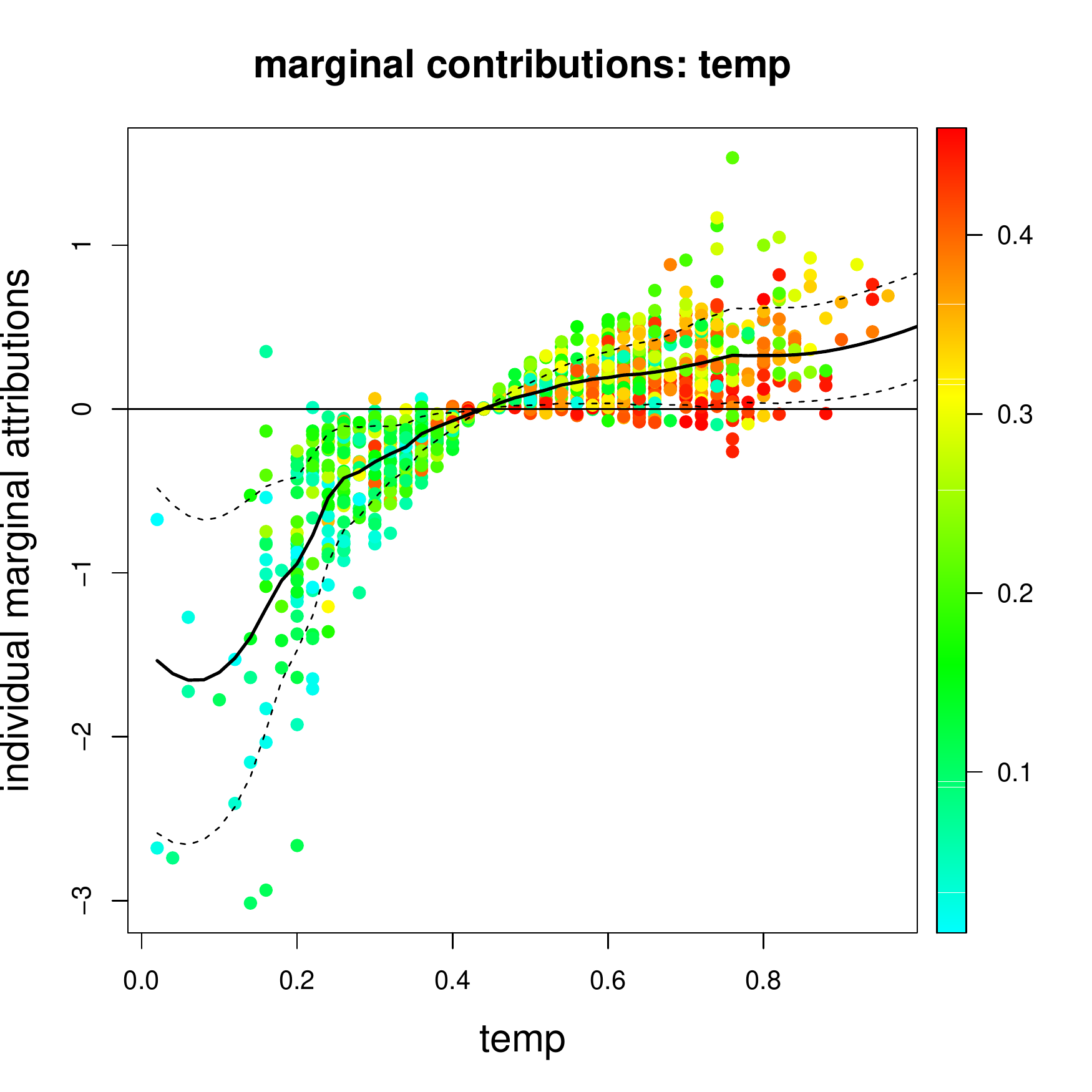}
\end{center}
\end{minipage}
\end{center}
\caption{Individual marginal contribuions $\omega_{i,j}$ of 1,000 randomly selected instances $\bx_i$
for (lhs) $j={\tt month}$, (middle) $j={\tt hour}$ and
(rhs) $j={\tt temp}$; the black line shows the empirical average; the colors show the expected responses 
$\mu(\bx_i)\in (0,1)$ (casual rental proportions).
}
\label{individual sensitivities 2}
\end{figure}

In Figure \ref{individual sensitivities} we have plotted the 
individual marginal contributions $\omega_{i,j}$ on the $y$-axis
against  the quantiles $\alpha \in (0,1)$ on the $x$-axis to  explain how the features $\bx_i$ enter 
the quantile levels $F_{\theta(\bX)}^{-1}(\alpha)$. This is the 3-ways analysis mentioned above, where
the third dimension is highlighted by using different colors in Figure \ref{individual sensitivities}.
Alternatively, we can also try to understand 
how this third dimension of different feature values $x_j$ contributes to the individual marginal contributions $\omega_{i,j}$.
Figure \ref{individual sensitivities 2} plots the individual marginal contributions $\omega_{i,j}$ on the $y$-axis against
the feature values $x_j$ on the $x$-axis.
The black line shows the averages of $\omega_{i,j}$ over all instances, and the colored dots show the 1,000
randomly selected instances $\bx_i$ with the colors illustrating the expected responses, i.e.~the 
expected casual rental proportions $\mu(\bx_i)=
\sigma(\theta(\bx_i)) \in (0,1)$. The general shape of the black lines in these graphs reflects well the marginal
empirical observations in Figure \ref{descriptive counts 2}. However, the detailed structure slightly differs in these
plots as they do not exactly show the same quantity, the latter shows a marginal empirical graph, whereas
Figure \ref{individual sensitivities 2} quantifies individual marginal contributions to expected responses
$\theta(\bx)$ in an additive way (on the canonical scale). Figure \ref{individual sensitivities 2} (rhs) shows a clear monotone plot which also 
results in a separation of the colors, whereas the colors in Figure \ref{individual sensitivities 2} (lhs, middle)
can only be fully understood by also studying contributions and interactions with other components $x_{i,k}$, $k\neq j$.

\subsection{Interaction terms}
\label{Interaction terms section}
\begin{figure}[htb!]
\begin{center}
\begin{minipage}[t]{0.45\textwidth}
\begin{center}
\includegraphics[width=\textwidth]{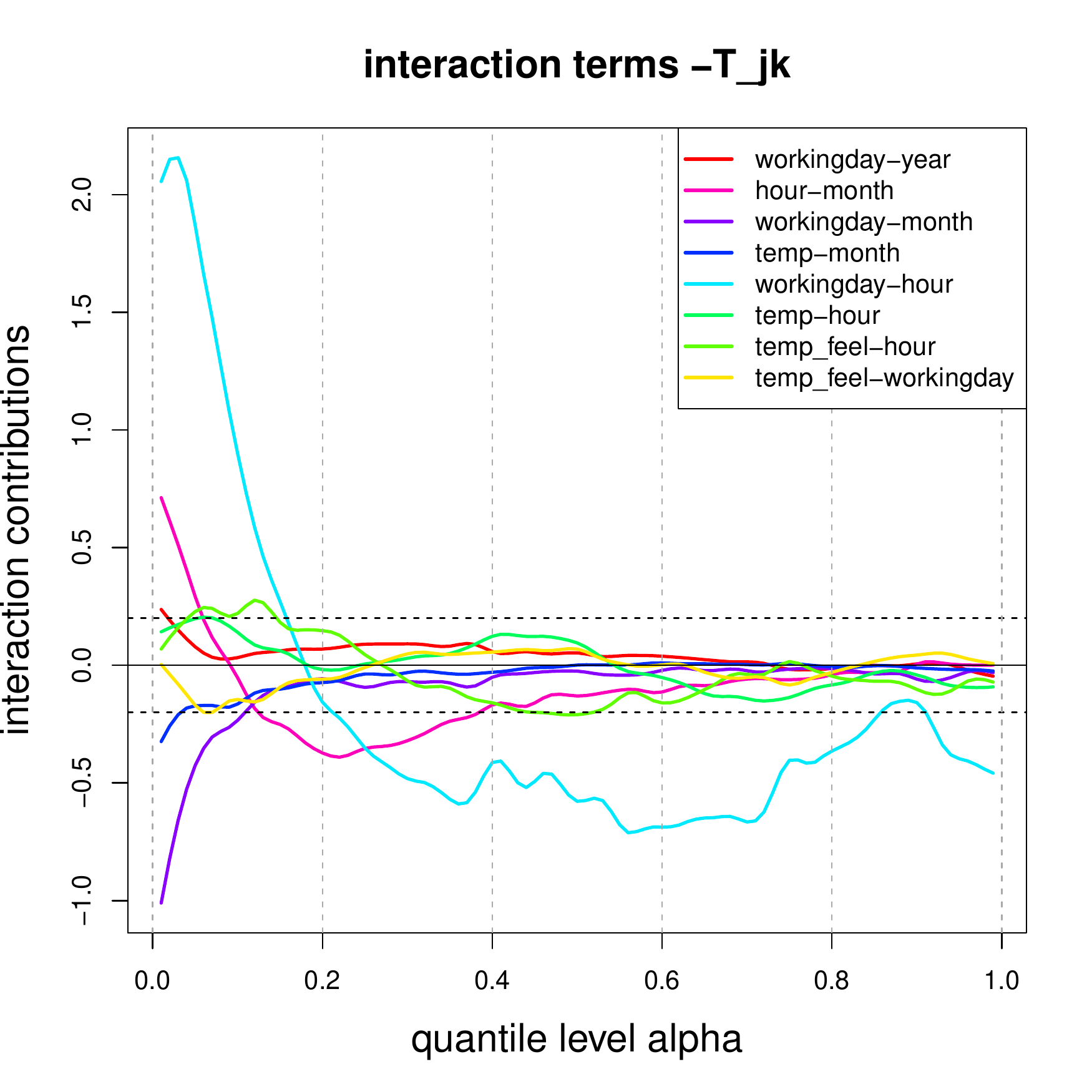}
\end{center}
\end{minipage}
\end{center}
\caption{Off-diagonal terms $-T_{j,k}(\theta;\alpha)$ giving the interactions.
}
\label{sensitivities part 2}
\end{figure}

There remains the analysis of the interaction terms $-T_{j,k}(\theta;\alpha)$, $j\neq k$,
that account for  the cyan shaded are in Figure \ref{gradient descent for reference} (rhs).
These interaction terms are shown in Figure \ref{sensitivities part 2}.

To not overload Figure \ref{sensitivities part 2} we only show the
interaction terms $T_{j,k}$ for which $\max_\alpha |T_{j,k}(\theta;\alpha)|>0.2$.
We identify three major interaction terms: {\tt workingday}-{\tt hour}, {\tt workingday}-{\tt month}
and {\tt hour}-{\tt month}. Of course, these interactions make perfect sense in describing
the casual rental proportion. For small quantiles also interactions {\tt temp}-{\tt month}
and {\tt temp}-{\tt hour} are important. Interestingly, we also find an interaction
{\tt workingday}-{\tt year}: in the data there is a positive trend 
of registered rental bike users (in absolute terms) which interacts differently on working and non-working
days because casual rentals are more frequent on non-working days. Identifying the importance
of these interactions highlights that it will not be sufficient to work within a generalized linear model (GLM)
or a generalized additive model (GAM) unless we add explicit interaction terms to them.

\begin{figure}[htb!]
\begin{center}
\begin{minipage}[t]{0.45\textwidth}
\begin{center}
\includegraphics[width=\textwidth]{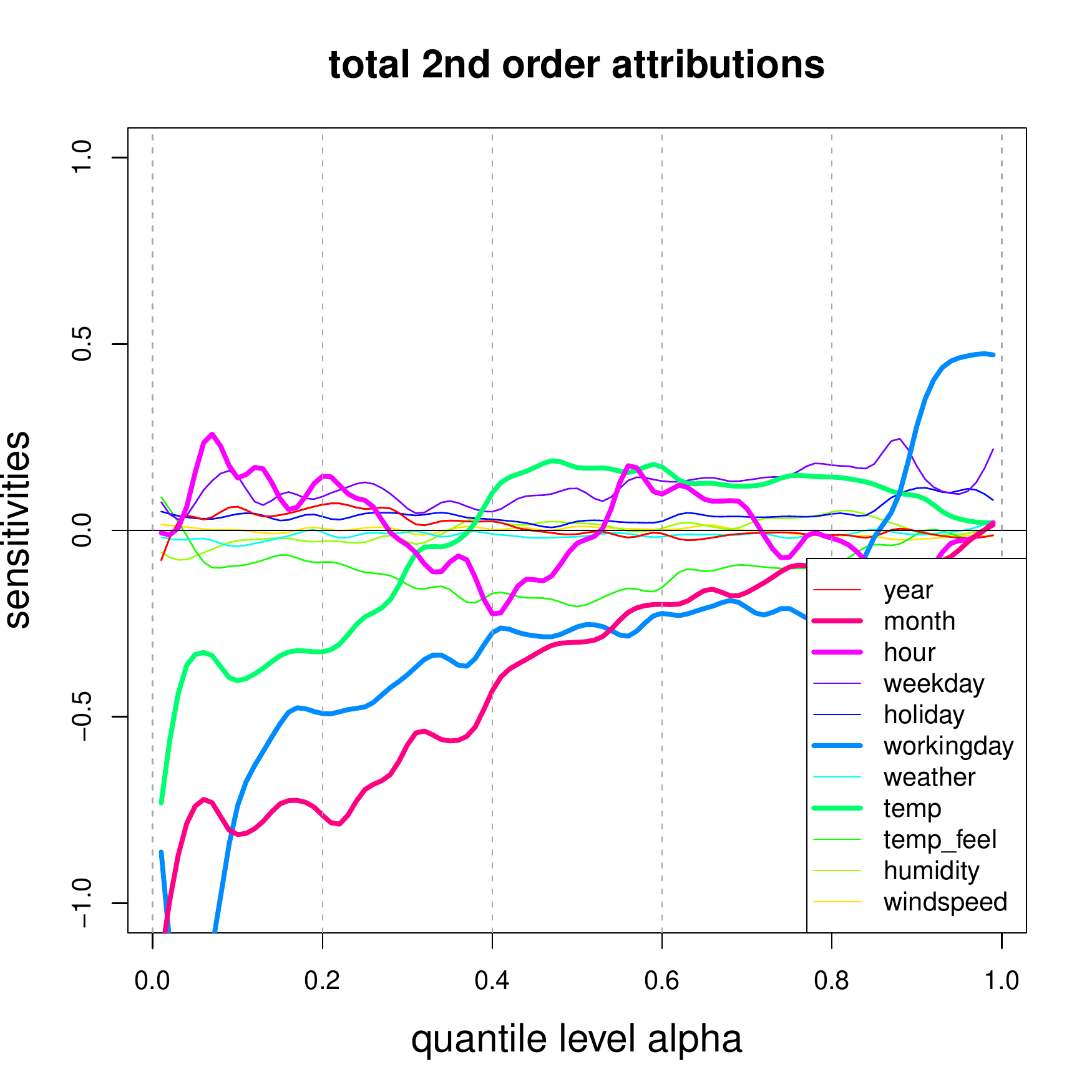}
\end{center}
\end{minipage}
\begin{minipage}[t]{0.45\textwidth}
\begin{center}
\includegraphics[width=\textwidth]{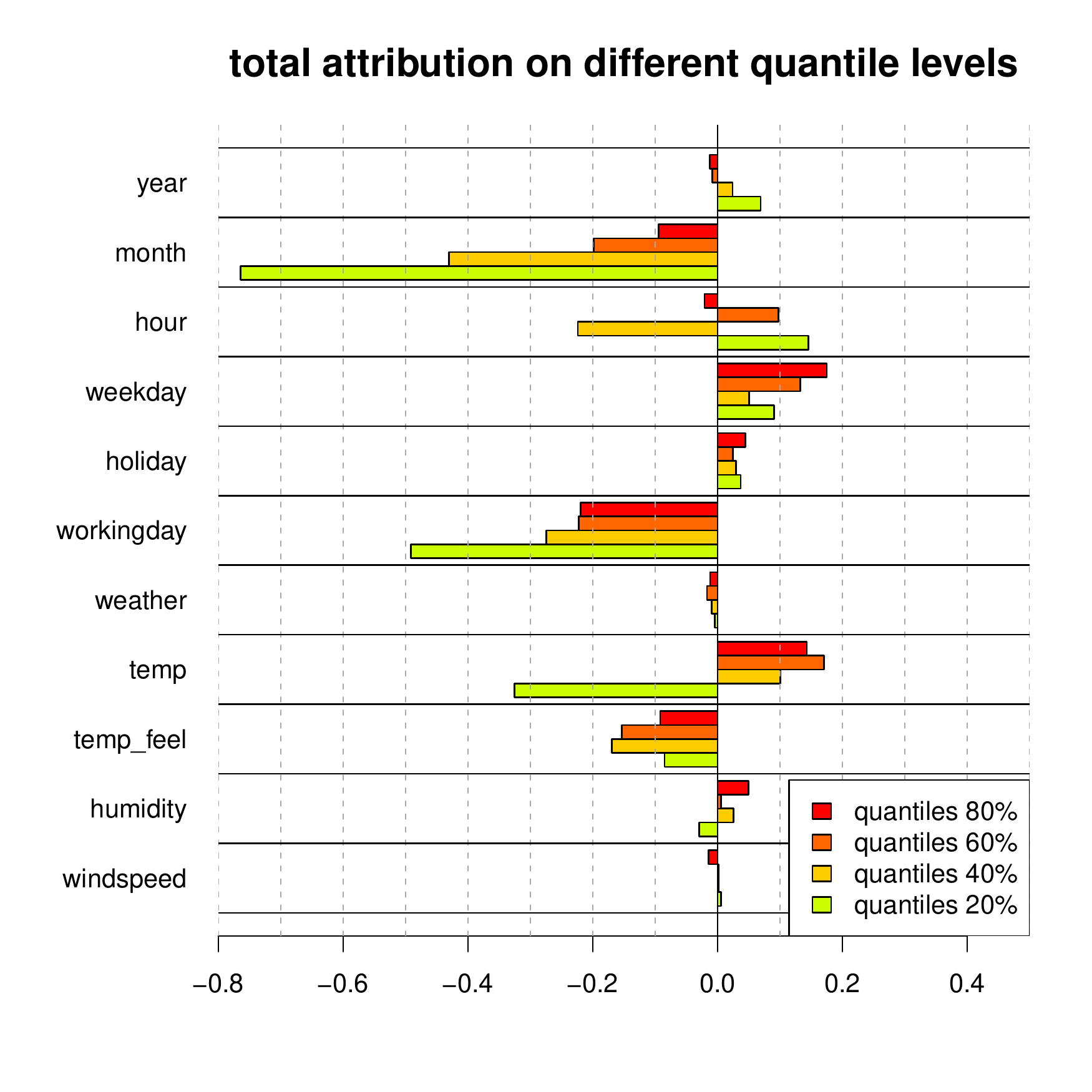}
\end{center}
\end{minipage}
\end{center}
\caption{(lhs) 2nd order attributions $V_j( \theta; \alpha)$
including interaction terms,  and (rhs) $V_j( \theta; \alpha)$ for
selected quantile levels
$\alpha \in \{20\%, 40\%,60\%, 80\%\}$.}
\label{2nd order contributions with interaction}
\end{figure}

In the final step we combine the attributions
$S_j(\theta; \alpha)-T_{j,j}(\theta; \alpha)/2$ with the interaction terms $T_{j,k}(\theta; \alpha)$, $k\le j$.
A natural way is to just allocate half of the interaction terms $T_{j,k}(\theta; \alpha)$ to each component
$j$ and $k$. This then provides allocated 2nd order attribution to components $1\le j \le q$
\begin{equation*}
V_j(\theta; \alpha)~=~
S_j(\theta; \alpha)-T_{j,j}(\theta; \alpha)/2 -\sum_{j\neq k} T_{j,k}(\theta; \alpha)/2
~=~S_j(\theta; \alpha) -\sum_{k=1}^q T_{j,k}(\theta; \alpha)/2.
\end{equation*}
Adding the reference level $\theta(\ba)$, we again receive the full 2nd order contributions
$C_{2,2}=\theta(\ba)+\sum_{j=1}^q V_j(\theta; \alpha)$ illustrated by the red line
in Figure \ref{gradient descent for reference} (rhs). In Figure 
\ref{2nd order contributions with interaction} we provide these 
attributions $V_j(\theta; \alpha)$ for quantiles $\alpha \in (0,1)$. These plots differ
from Figure \ref{2nd order contributions w/o interaction} only by the inclusion of the
2nd order off-diagonal (interaction) terms. Comparing the right-hand sides
of these two plots we observe that firstly the level is shifted, which is explained
by the shaded cyan area in Figure \ref{gradient descent for reference} (rhs).
Secondly, interactions impact mainly the small quantiles in our example,
this is clear from Figure \ref{sensitivities part 2} and, for instance, impacts the significance
of {\tt hour} on the 20\% quantile level.

\subsection{Scrolling through the network layers}
Up to this point our MACQ analysis has been fully general, in the sense that it can be applied
to any smooth deep learning model. In the last step of our analysis we specifically focus on the deep network
 introduced in Section \ref{Model choice and model fitting}, and 
we try to better understand how networks learn new representations through the network
layers. A deep feed-forward neural network $\theta: \R^q \to \R$ is a composition 
of $d$ hidden neural network layers $\bz^{(k)}:\R^{q_{k-1}} \to \R^{q_k}$, $1\le k \le d$; we initialize 
input dimension $q_0=q$. Define the composition $\bx \mapsto \bz^{(d:1)}(\bx)
=(\bz^{(d)}\circ \ldots \circ \bz^{(1)})(\bx)$ which maps input $\bx \in \R^q$
to the last hidden network layer having dimension $q_d$.
Network \eqref{logistic network definition} with logistic output
can then be written as
\begin{equation*}
\bx\in \R^q ~~\mapsto ~~ \mu(\bx) = \sigma (\theta(\bx))=
\sigma \left(\beta_0 + \bbeta^\top \bz^{(d:1)}(\bx)\right),
\end{equation*}
with bias/intercept $\beta_0 \in \R$ and regression parameter/weight $\bbeta \in \R^{q_d}$.
This should be compared to linear regression \eqref{linear regression definition}.

Each hidden layer learns a new representation of the inputs $\bx_i$, that is, 
the representations learned in layer $k$ are given by 
$\bx^{(k:1)}_i:=(\bz^{(k)}\circ \ldots \circ \bz^{(1)})(\bx_i)$, for $1\le i \le n$. We can view
these learned representations as new inputs to the
remaining network after hidden layer $k$
\begin{equation*}
\bx\in \R^{q_k} ~~\mapsto ~~  \sigma \left(\beta_0 + \bbeta^\top \bz^{(d:k+1)}(\bx)\right)=
\sigma \left(\beta_0 + \bbeta^\top (\bz^{(d)}\circ \ldots \circ \bz^{(k+1)})(\bx)\right).
\end{equation*}
In the following analysis we consider the instances $(Y_i,\bx^{(k:1)}_i)$ with these
learned features $\bx^{(k:1)}_i$ as inputs to the remaining network $\bz^{(d:k+1)}$
after layer $k$, and we perform the same MACQ analysis in this reduced setup.

\begin{figure}[htb!]
\begin{center}
\begin{minipage}[t]{0.32\textwidth}
\begin{center}
\includegraphics[width=\textwidth]{./PlotsBikes/linear_square_approximation2.pdf}
\end{center}
\end{minipage}
\begin{minipage}[t]{0.32\textwidth}
\begin{center}
\includegraphics[width=\textwidth]{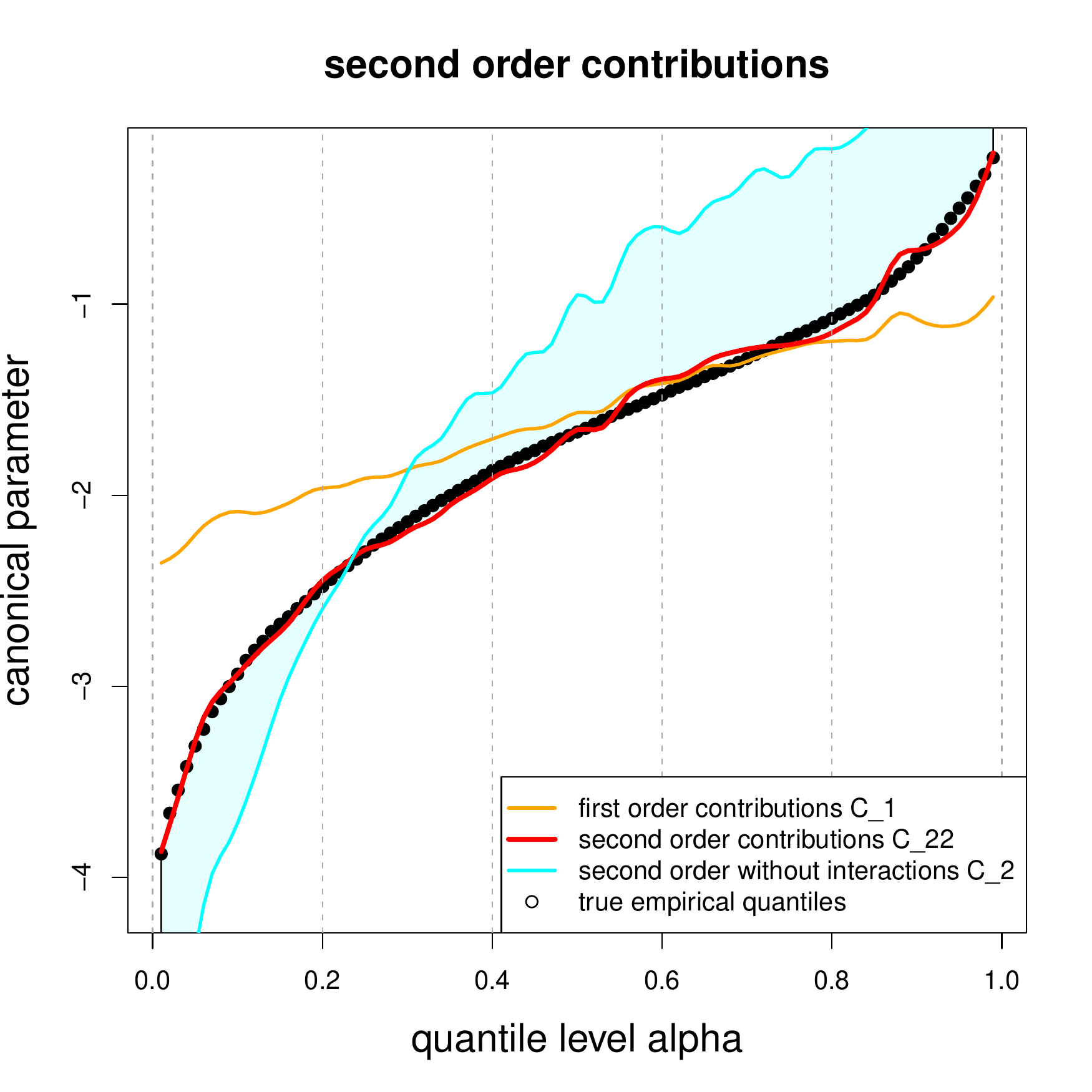}
\end{center}
\end{minipage}
\begin{minipage}[t]{0.32\textwidth}
\begin{center}
\includegraphics[width=\textwidth]{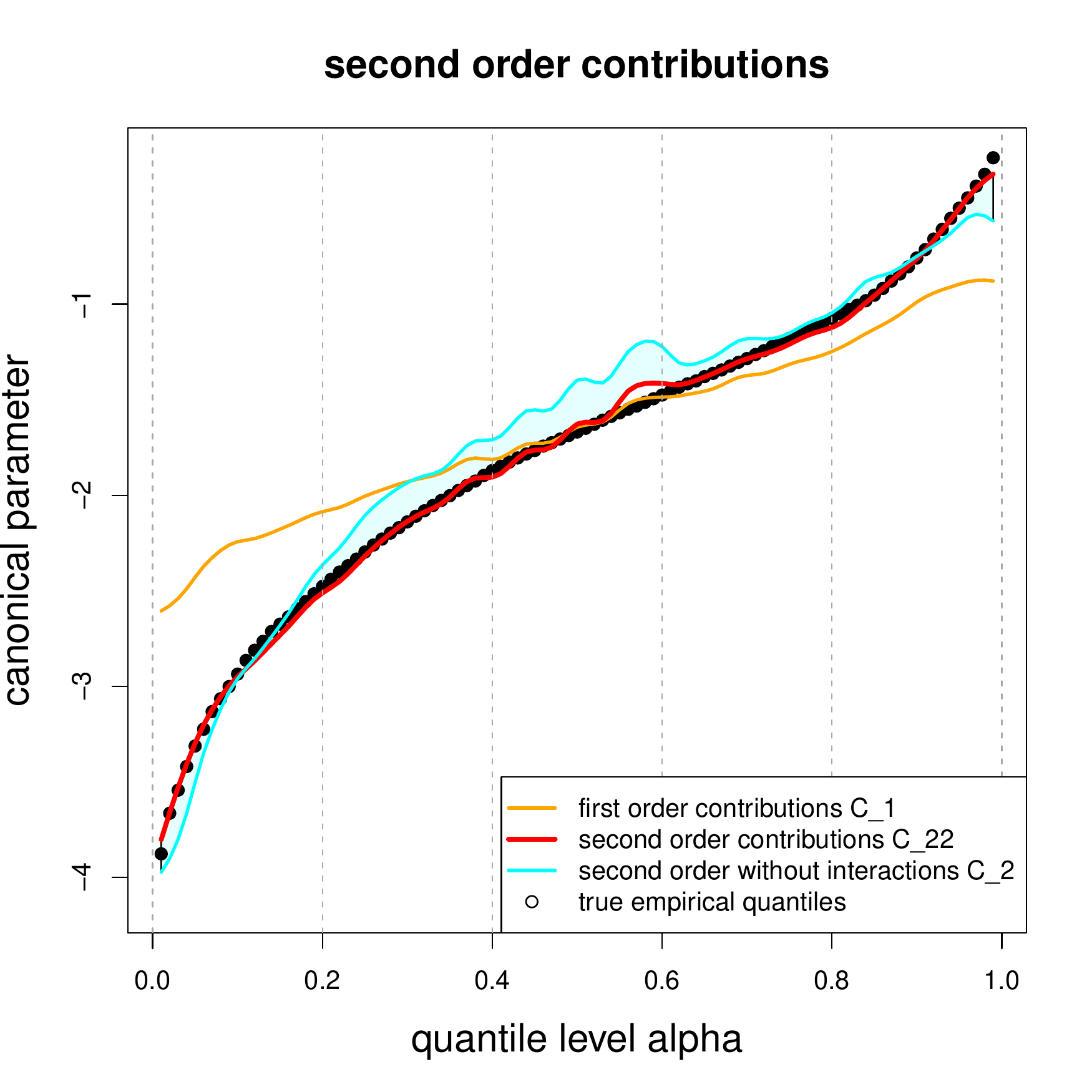}
\end{center}
\end{minipage}
\end{center}
\caption{2nd order contributions \eqref{second order contributions 3} of the (learned) representations:  (lhs) original inputs $\bx_i$, (middle) learned
representations $\bx^{(1:1)}_i$, and (rhs) learned
representations $\bx^{(2:1)}_i$.}
\label{learned representations}
\end{figure}

Figure \ref{learned representations} provides the 2nd order contributions \eqref{second order contributions 3} 
of the original inputs (lhs), the learned
representations $\bx^{(1:1)}_i$ in the first hidden layer (middle), and the learned
representations $\bx^{(2:1)}_i$ in the second hidden layer (rhs) on the corresponding
remaining networks $\bz^{(3:k+1)}$. We interpret these MACQ results as follows. The
first hidden layer (middle graph) has mainly a smoothing effect in recomposing the inputs $\bx_i$ suitably. The second
layer takes care of the interaction effects diminishing the cyan shaded area in Figure \ref{learned representations} (rhs).
Of course, this makes perfect sense as the output layer considers a linear function with weight $\bbeta \in \R^{q_d}$ which no longer allows for interactions. Therefore, interactions
need to be learned in the previous layers. The same applies to non-linear structures (on the canonical scale).
This completes our example.

\section{Conclusions}
\label{Conclusions}
This manuscript proposes a novel gradient-based global model-agnostic tool that can
be calculated efficiently for differentiable deep learning models and produces
informative visualizations. This tool studies
marginal attribution to feature components on a given response level. Marginal attributions
allow us to separate marginal effects of individual feature components from interaction effects,
and they allow us to study resulting variable importance plots on different parts of the decision
space characterized by different response levels. This variable importance is measured w.r.t.~a reference point 
that calibrates the entire space for our explanation. Finding a good reference
point has been efficiently performed by a simple gradient descent search.
A main result of our model-agnostic tool is a 3-way relationship between marginal attribution, output level and
feature value which can be illustrated in different ways. This extends response sensitivity analyses,
such as accumulated local effects, by an additional marginal attribution view.

\bibliographystyle{apalike}

{\small %\baselineskip.5em
\renewcommand{\baselinestretch}{.51}
}

\newpage

\appendix

\section{Sensitivities in distortion risk measures}
\label{Sensitivities in distortion risk measures}
The purpose of this appendix is to briefly explain distortion risk measures and how they
relate to marginal attribution. For this discussion we impose stronger assumptions
than we need above, i.e., these more restrictive assumptions are only made for the
explanation here.
Assume the expected response $\mu(\bX)$ has a continuous distribution function $F_{\mu(\bX)}$. It follows that
$U_{\mu(\bX)}=F_{\mu(\bX)}(\mu(\bX))$ is uniformly distributed on $[0,1]$. Choose a density $\zeta$ on
$[0,1]$. We can interpret $\zeta(U_{\mu(\bX)})$ as a probability distortion (probability re-weighting scheme
inducing a change of probability measure) because we have
\begin{equation*}
\E_P\left[\zeta(U_{\mu(\bX)})\right] = \int_0^1 \zeta(u) du =1.
\end{equation*}
The distorted expected response can then be defined by
\begin{equation*}
\varrho(\mu(\bX);\zeta) = \E_P \left[\mu(\bX) \zeta(U_{\mu(\bX)}) \right].
\end{equation*}
The functional $\varrho(\mu(\bX);\zeta)$ describes a {\it distortion risk measure}, see \cite{Wang}
and \cite{Acerbi}. It can
be interpreted as a Radon--Nikod\'ym derivative changed probability measure 
$dP_\zeta(\bX=\bx)=\zeta(U_{\mu(\bx)})dP(\bX=\bx)$.
We study the sensitivities of this distortion risk measure w.r.t.~the components of $\bX$. Assume that
the following directional derivatives exist in zero for all $1\le j \le q$
\begin{equation*}
S_j( \mu;\zeta) = \frac{\partial}{\partial \varepsilon} 
\left.\varrho\left(\mu\left((X_1,\ldots, X_{j-1}, X_j(1+\varepsilon),X_{j+1}, \ldots X_q)^\top\right);\zeta\right)\right|_{\varepsilon=0}.
\end{equation*}
Then, $S_j(\mu;\zeta) $ can be interpreted as the sensitivity of $\bX\mapsto\mu(\bX)$ in feature component $X_j$.
\cite{Hong1} and  \cite{Tsanakas1} prove under different sets of assumptions
 that these sensitivities satisfy
\begin{equation*}
S_j( \mu;\zeta) =\E_P \left[X_j \mu_j(\bX) \zeta(U_{\mu(\bX)}) \right].
\end{equation*}
Observe that this exactly uses the marginal attribution \eqref{marginal attribution}.
We still have the freedom of choosing the density $\zeta$ on $[0,1]$. If we choose the uniform
distribution $\zeta\equiv 1$ on $[0,1]$ we receive the average expected response and its 
average marginal attribution
\begin{equation*}
\varrho(\mu(\bX);\zeta\equiv 1)=\E_P[\mu(\bX)] \qquad \text{ and } \qquad
S_j(\mu;\zeta\equiv 1) = \E_P[X_j \mu_j(\bX)] .
\end{equation*}
If we choose for density $\zeta$ the Dirac measure $\delta_\alpha$ in $\alpha \in (0,1)$, which allocates
probability weight 1 to $\alpha$, this  gives us the  $\alpha$-quantile
\begin{equation*}
\varrho(\mu(\bX);\zeta=\delta_\alpha)=F_{\mu(\bX)}^{-1}(\alpha).
\end{equation*}
For its sensitivities we receive for $1\le j \le q$
\begin{equation*}
S_j( \mu; \zeta=\delta_\alpha) 
~=~\E_P \left[X_j \mu_j(\bX) \left| \mu(\bX)=F_{\mu(\bX)}^{-1}(\alpha)\right.\right],
\end{equation*}
which exactly corresponds to 1st order attribution \eqref{VaR sensitivity}.

\medskip

{\bf Remark.} We could choose any other density $\zeta$ on $[0,1]$ to obtain sensitivities of other
distortion risk measures. Such other choices may also have interesting counterparts in interpreting smooth deep learning models, by reflecting attention to different areas of the prediction space.

\section{Descriptive analysis of bike rental example}
\label{Descriptive analysis of bike rental example}
In this appendix, we give a brief descriptive analysis of the data used that helps us to interpret the
network regression models. The data comprises the number of casual and registered bike rentals
every hour from 2011/01/01 until 2012/12/31. This data has originally been studied in 
\cite{Fanaee} and \cite{Apley}, and it can be downloaded from
\url{https://archive.ics.uci.edu/ml/datasets/Bike+Sharing+Dataset}. Listing \ref{ExcerptOfData}
gives a short excerpt of the data.

\begin{lstlisting}[float=h,frame=tb,caption={Excerpt of bike rental data.},
label=ExcerptOfData]
'data.frame':   17379 obs. of  13 variables:
 $ date      : Date, format: "2011-01-01" "2011-01-01" "2011-01-01" ...
 $ year      : num  2011 2011 2011 2011 2011 ...
 $ month     : int  1 1 1 1 1 1 1 1 1 1 ...
 $ hour      : int  0 1 2 3 4 5 6 7 8 9 ...
 $ weekday   : int  6 6 6 6 6 6 6 6 6 6 ...
 $ holiday   : Factor w/ 2 levels "holiday","no-holiday": 2 2 2 2 2 2 2 2 2 2 ...
 $ workingday: Factor w/ 2 levels "no-working","workingday": 1 1 1 1 1 1 1 1 1 1 ...
 $ weather   : num  1 1 1 1 1 2 1 1 1 1 ...
 $ temp      : num  0.24 0.22 0.22 0.24 0.24 0.24 0.22 0.2 0.24 0.32 ...
 $ temp_feel : num  0.288 0.273 0.273 0.288 0.288 ...
 $ humidity  : num  0.81 0.8 0.8 0.75 0.75 0.75 0.8 0.86 0.75 0.76 ...
 $ windspeed : num  0 0 0 0 0 0.0896 0 0 0 0 ...
 $ casual    : int  3 8 5 3 0 0 2 1 1 8 ...
 $ registered: int  13 32 27 10 1 1 0 2 7 6 ...
 $ count     : int  16 40 32 13 1 1 2 3 8 14 ...
\end{lstlisting}

As response variable we consider the proportion of casual rentals relative
to all rentals, thus, we set response $Y={\tt casual}/{\tt count} \in [0,1]$ on
an hourly grid over the entire observation period. These are $n=17,379$ hours from
2011/01/01 until 2012/12/31, see line 1 of Listing \ref{ExcerptOfData}. We note that
${\tt count}\ge 1$ for all observations, which makes $Y$ well-defined throughout the whole
observation period. The goal is to predict this response variable $Y$ based on available 
feature information $\bx$ which is provided on lines 3-13 of Listing \ref{ExcerptOfData}.
These are the {\tt year}, {\tt month} and {\tt hour} of the observations $Y$. The
{\tt weekday} (with 0 for Sunday), {\tt holiday} (yes/no for public holiday), {\tt workingday} (yes/no, the former
neither being a public holiday nor a weekend), {\tt weather} (1,2 and 3 for clear, cloudy and rain/snow),
temperature {\tt temp}, the felt temperature {\tt temp\_feel}, {\tt humidity} and {\tt windspeed}.
Note that all these features are continuous or binary, thus, we can directly use this feature
encoding for regression modeling.

\begin{figure}[htb!]
\begin{center}
\begin{minipage}[t]{0.45\textwidth}
\begin{center}
\includegraphics[width=\textwidth]{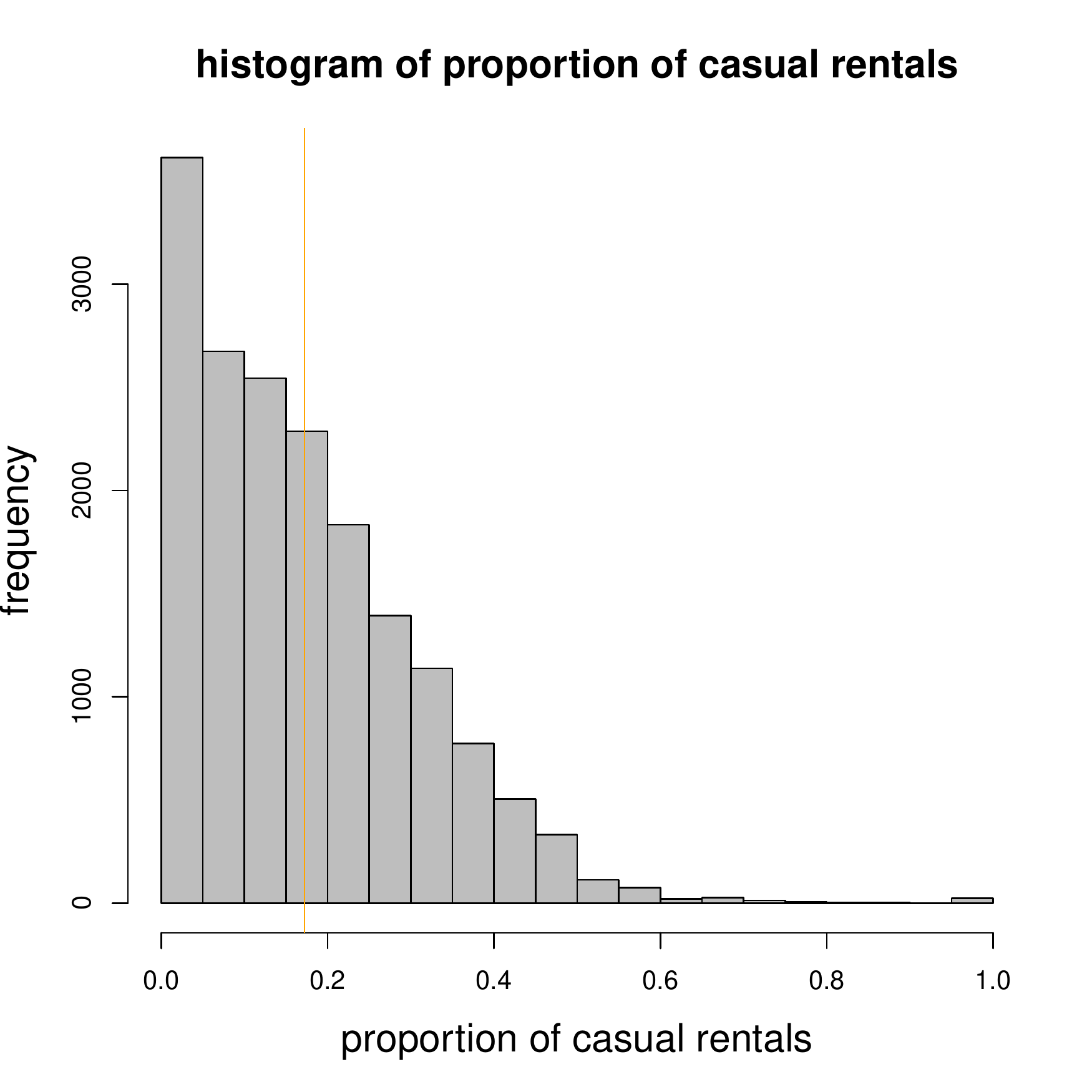}
\end{center}
\end{minipage}
\begin{minipage}[t]{0.45\textwidth}
\begin{center}
\includegraphics[width=\textwidth]{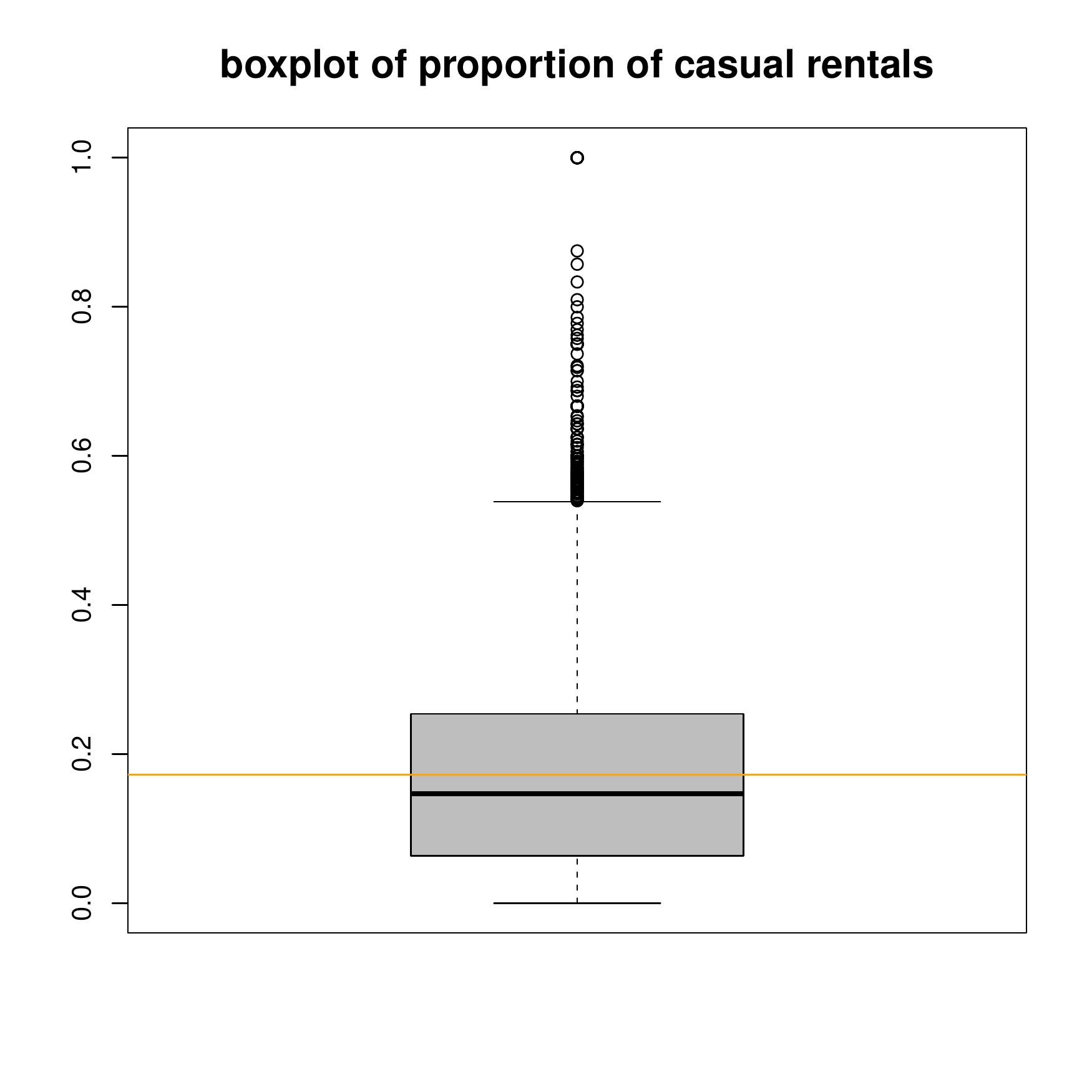}
\end{center}
\end{minipage}
\end{center}
\caption{(lhs) Histogram and (rhs) boxplot of (hourly) responses
$Y={\tt casual}/{\tt count} \in [0,1]$ over the entire observation period;
the orange line shows the empirical mean of  17\%.}
\label{descriptive counts}
\end{figure}

We illustrate this data. Figure \ref{descriptive counts} shows the
observed responses $Y={\tt casual}/{\tt count}$ over the entire observation period.
In average the casual rentals make 17\% of all rentals, and the empirical density
of $Y$ is strongly skewed.

\begin{figure}[htb!]
\begin{center}
\begin{minipage}[t]{0.32\textwidth}
\begin{center}
\includegraphics[width=\textwidth]{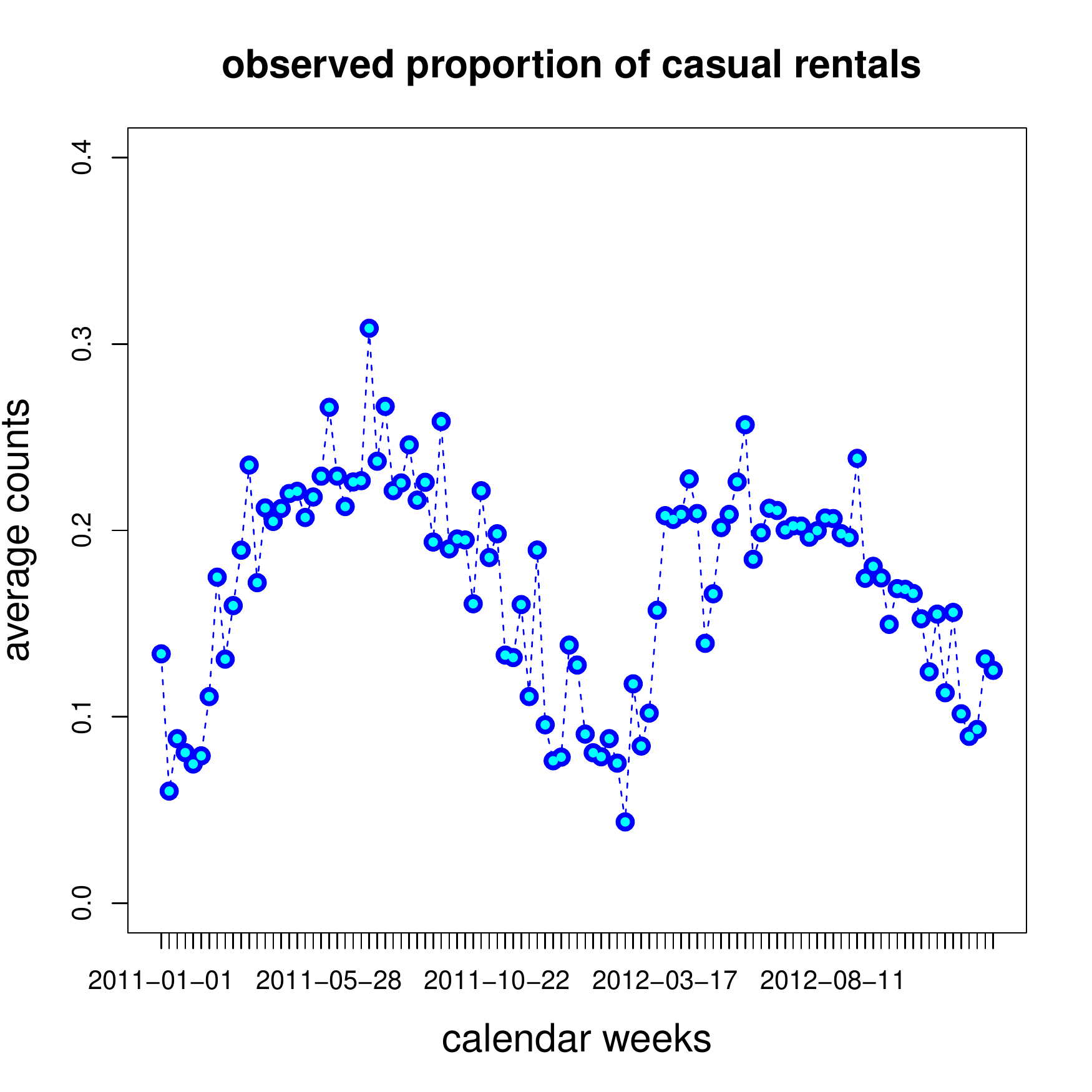}
\end{center}
\end{minipage}
\begin{minipage}[t]{0.32\textwidth}
\begin{center}
\includegraphics[width=\textwidth]{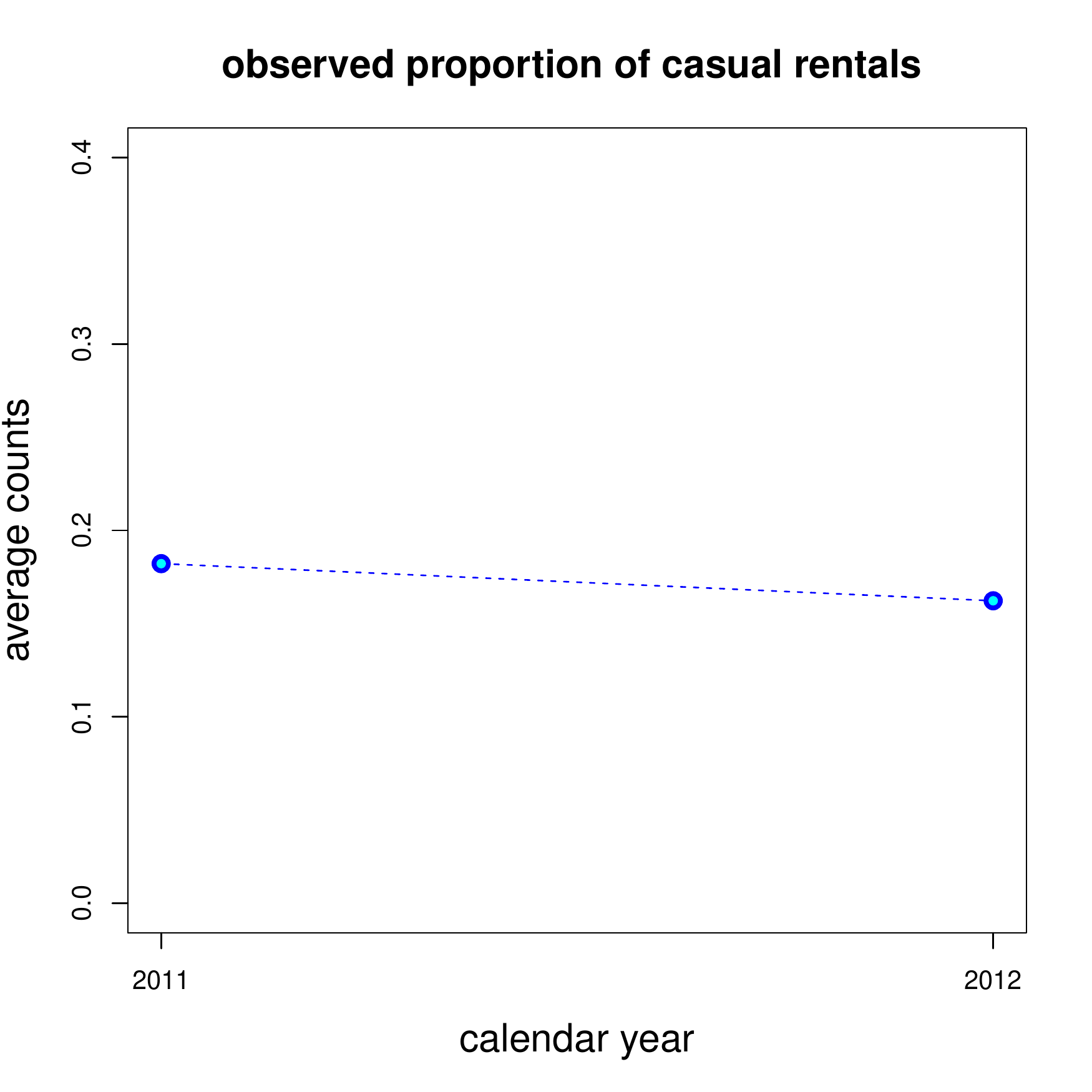}
\end{center}
\end{minipage}
\begin{minipage}[t]{0.32\textwidth}
\begin{center}
\includegraphics[width=\textwidth]{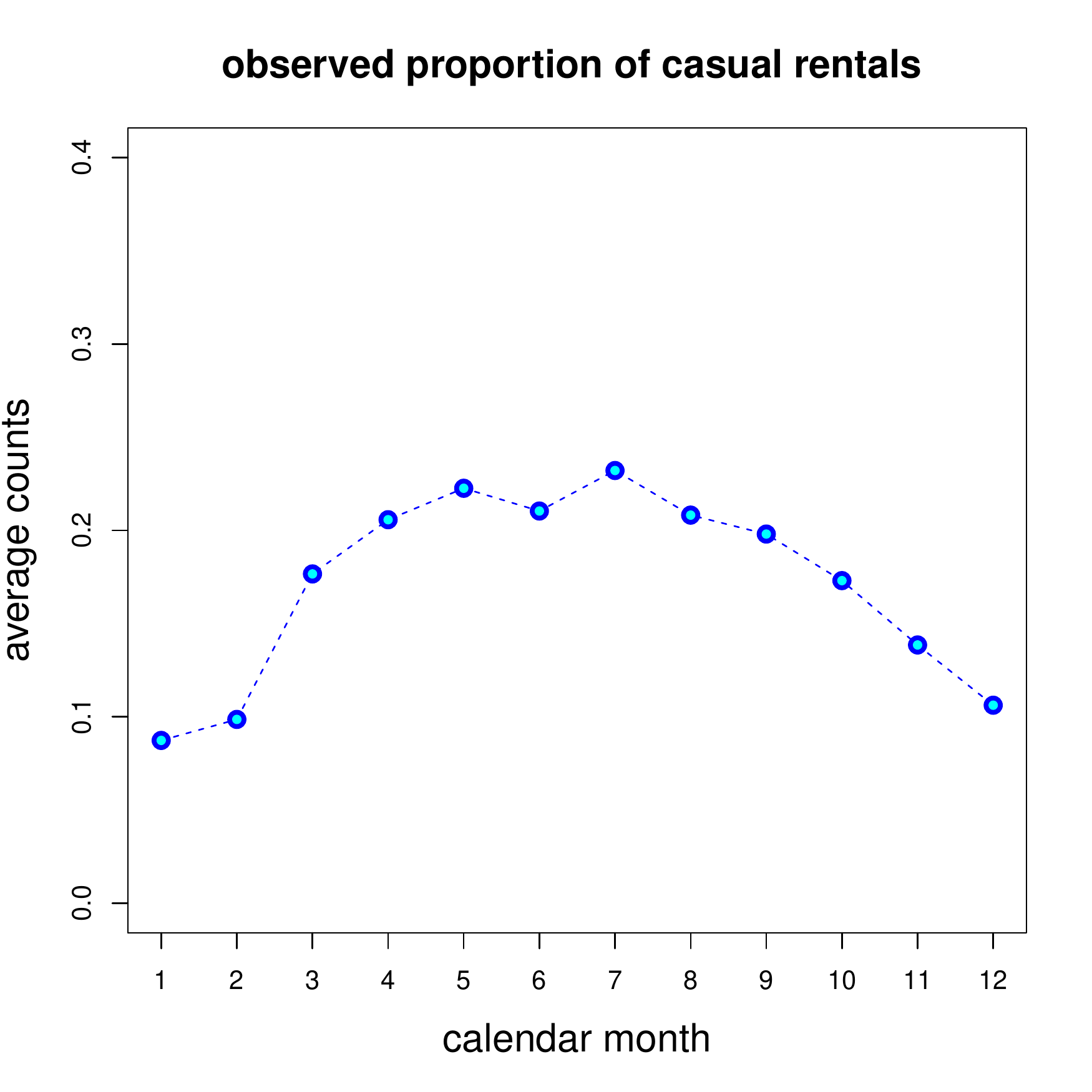}
\end{center}
\end{minipage}

\begin{minipage}[t]{0.32\textwidth}
\begin{center}
\includegraphics[width=\textwidth]{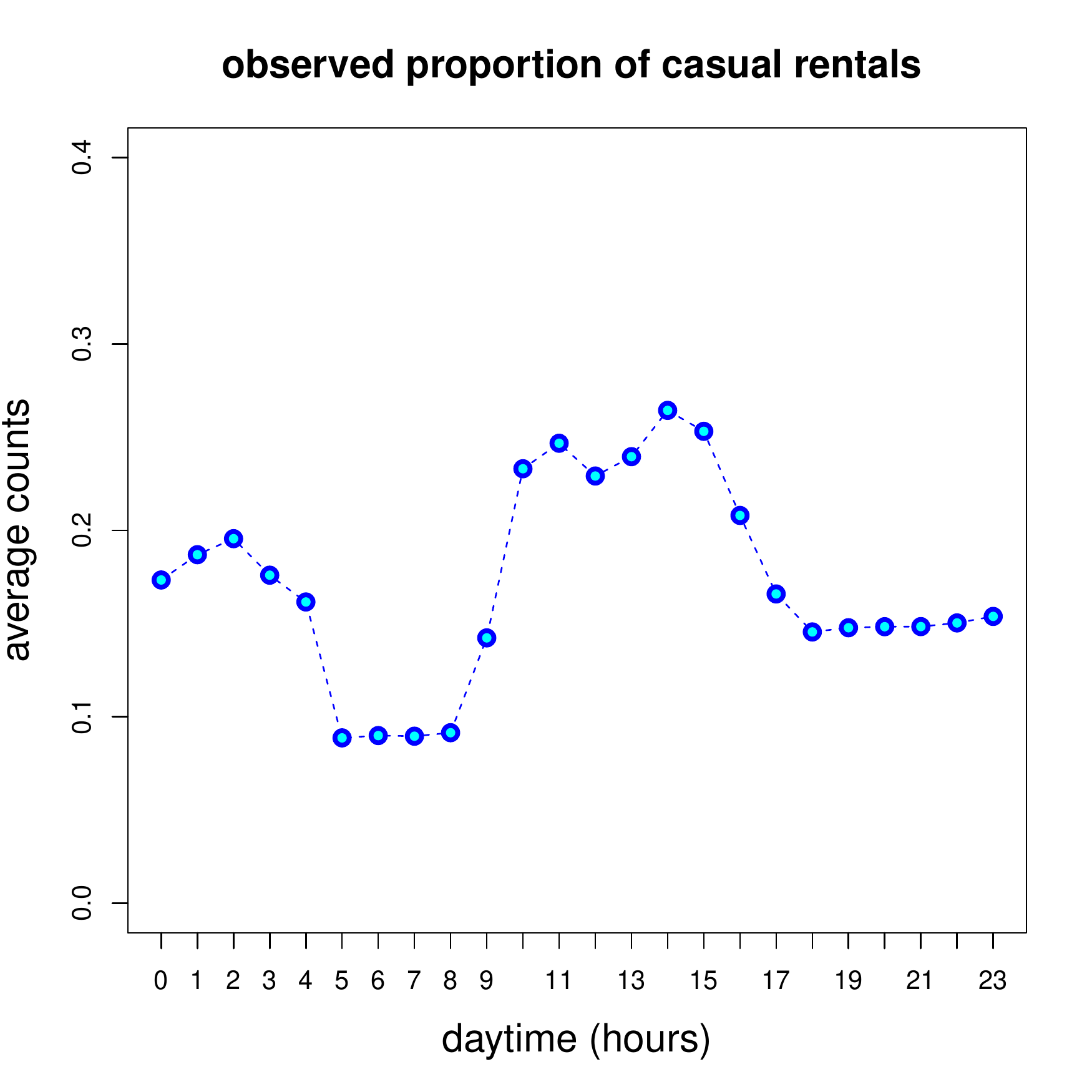}
\end{center}
\end{minipage}
\begin{minipage}[t]{0.32\textwidth}
\begin{center}
\includegraphics[width=\textwidth]{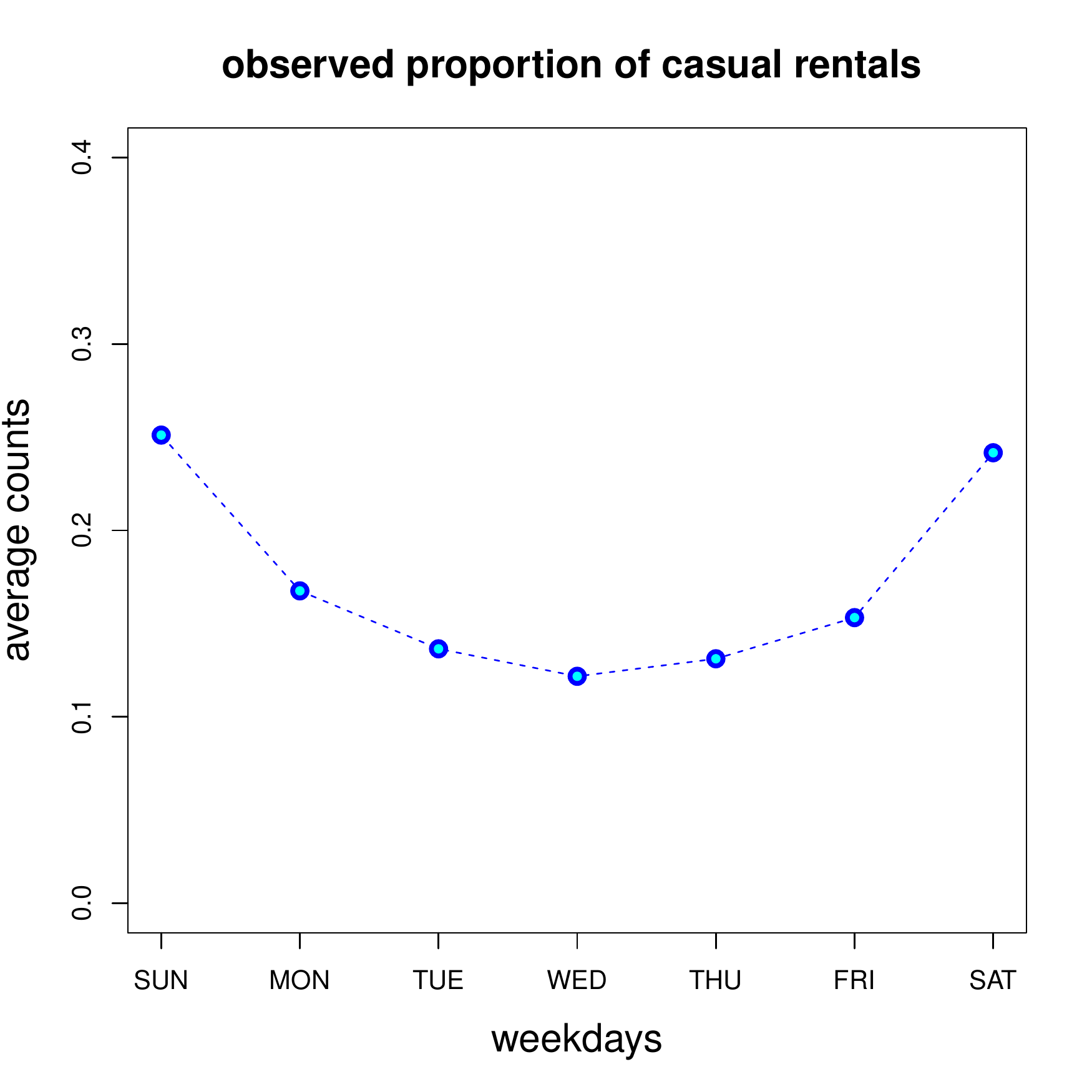}
\end{center}
\end{minipage}
\begin{minipage}[t]{0.32\textwidth}
\begin{center}
\includegraphics[width=\textwidth]{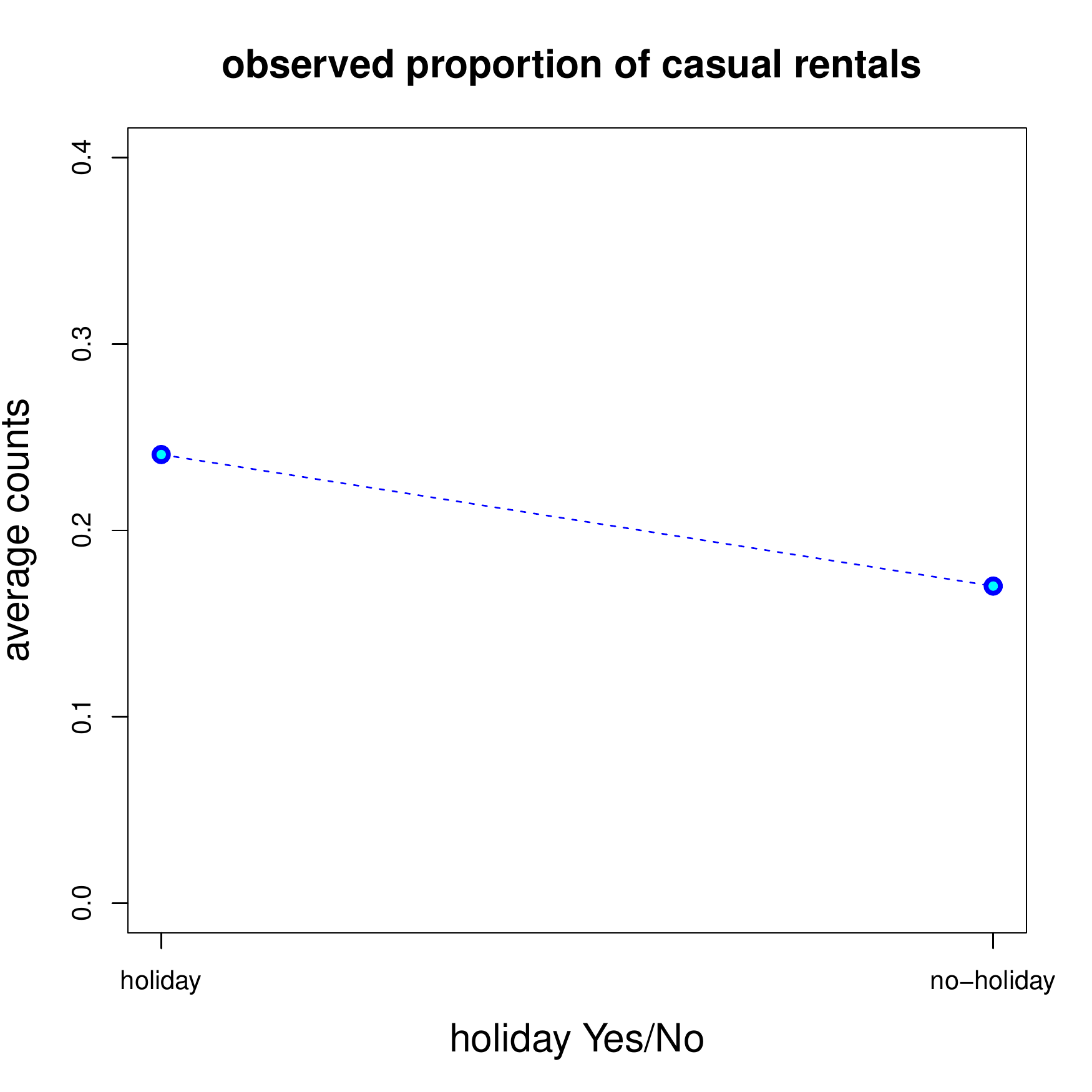}
\end{center}
\end{minipage}

\begin{minipage}[t]{0.32\textwidth}
\begin{center}
\includegraphics[width=\textwidth]{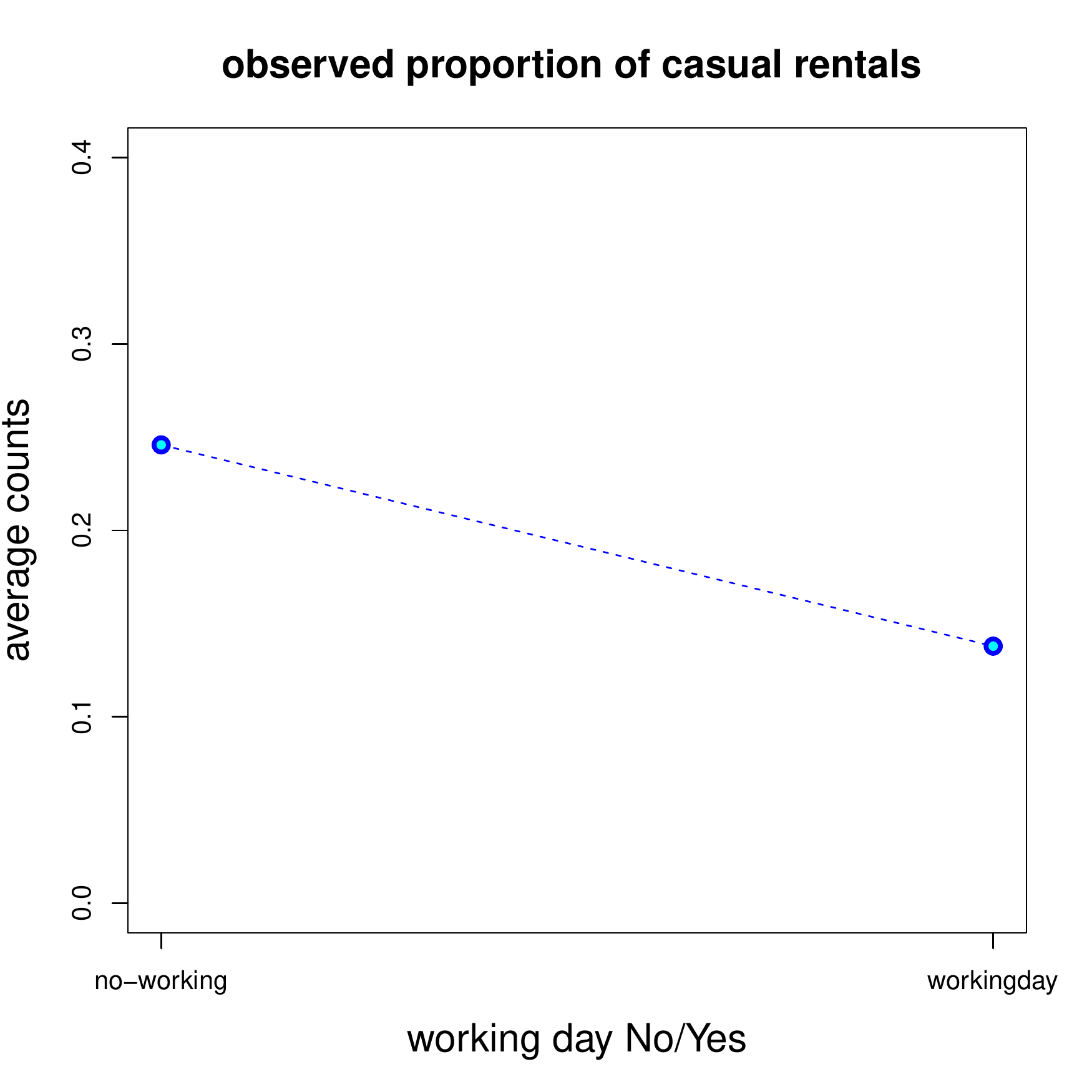}
\end{center}
\end{minipage}
\begin{minipage}[t]{0.32\textwidth}
\begin{center}
\includegraphics[width=\textwidth]{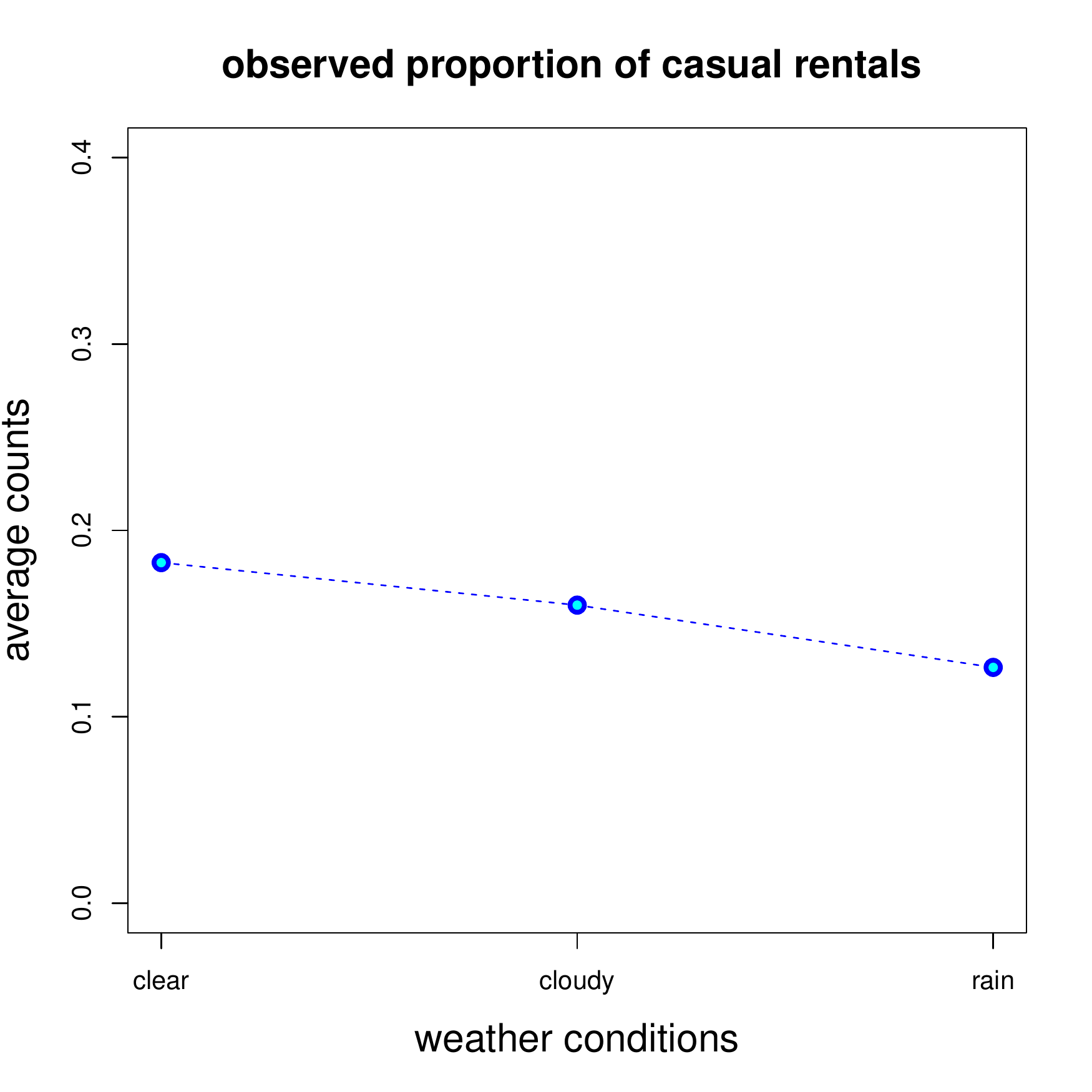}
\end{center}
\end{minipage}
\begin{minipage}[t]{0.32\textwidth}
\begin{center}
\includegraphics[width=\textwidth]{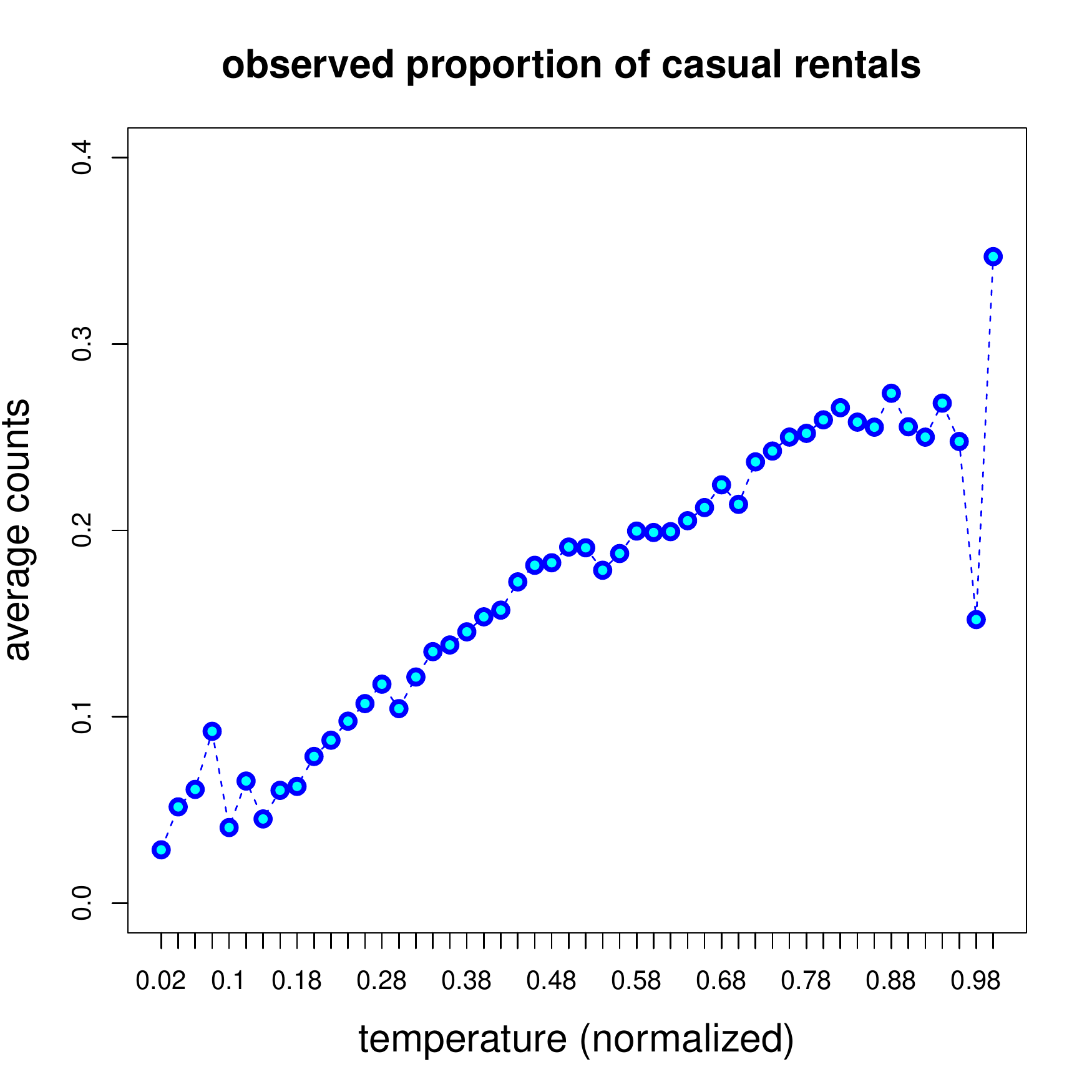}
\end{center}
\end{minipage}

\begin{minipage}[t]{0.32\textwidth}
\begin{center}
\includegraphics[width=\textwidth]{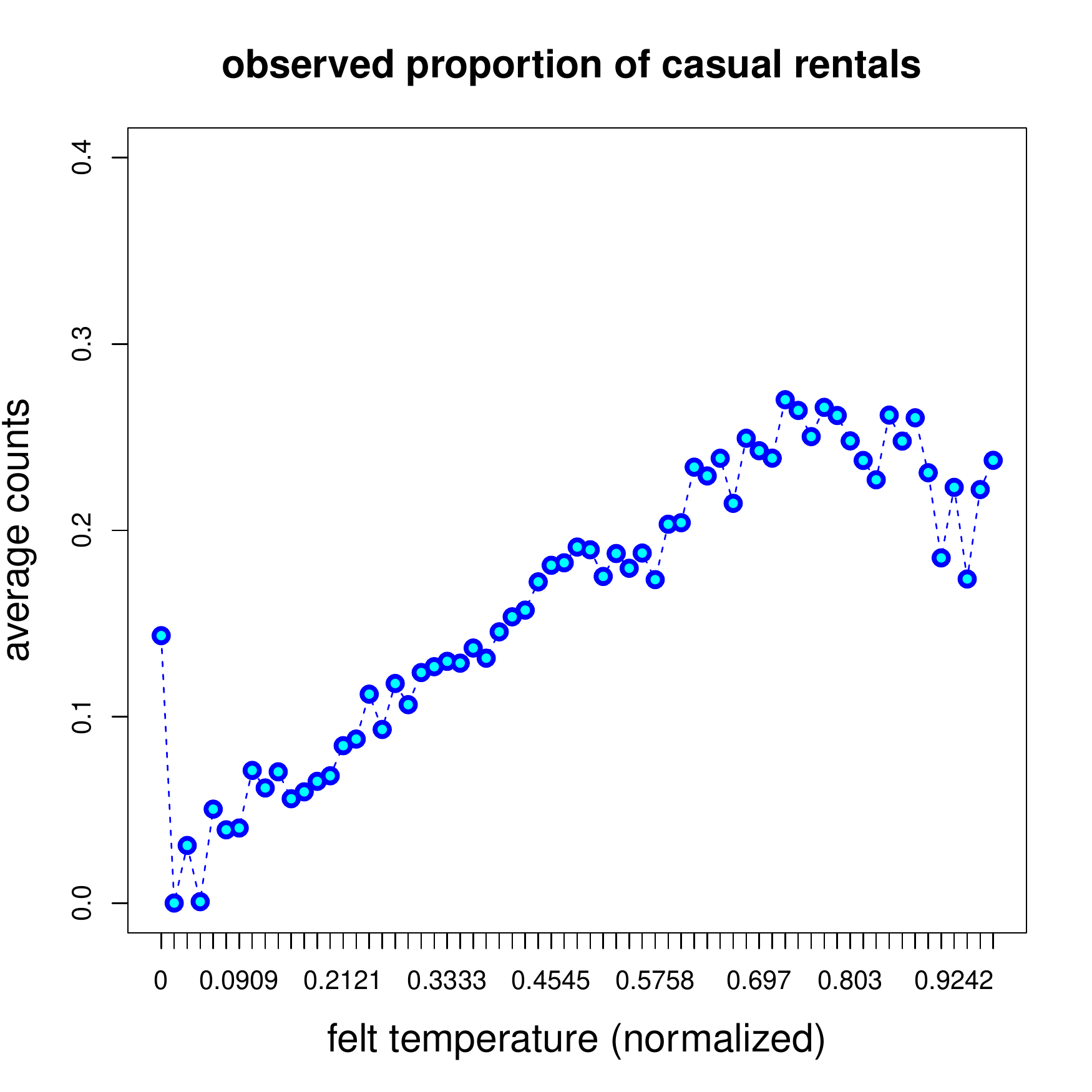}
\end{center}
\end{minipage}
\begin{minipage}[t]{0.32\textwidth}
\begin{center}
\includegraphics[width=\textwidth]{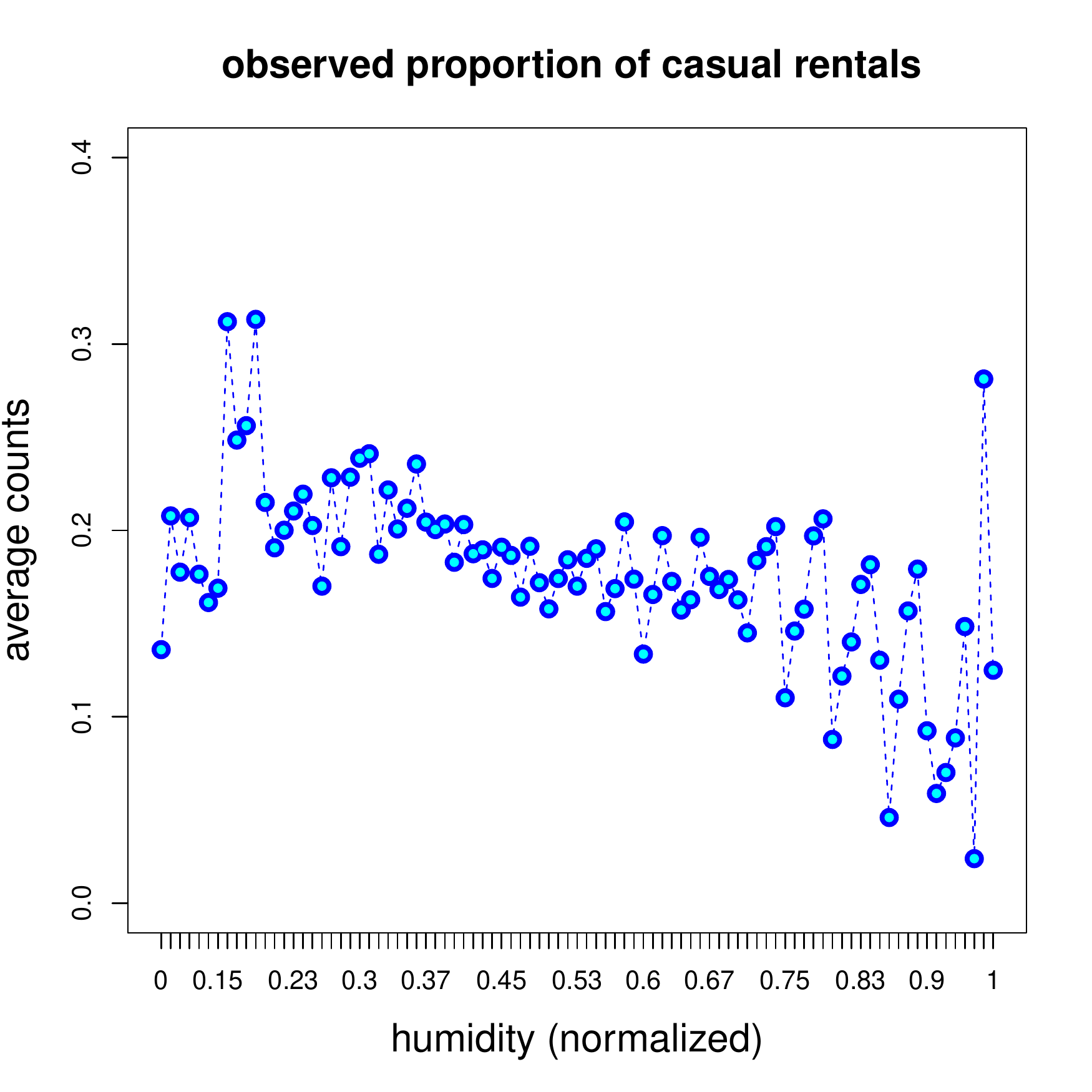}
\end{center}
\end{minipage}
\begin{minipage}[t]{0.32\textwidth}
\begin{center}
\includegraphics[width=\textwidth]{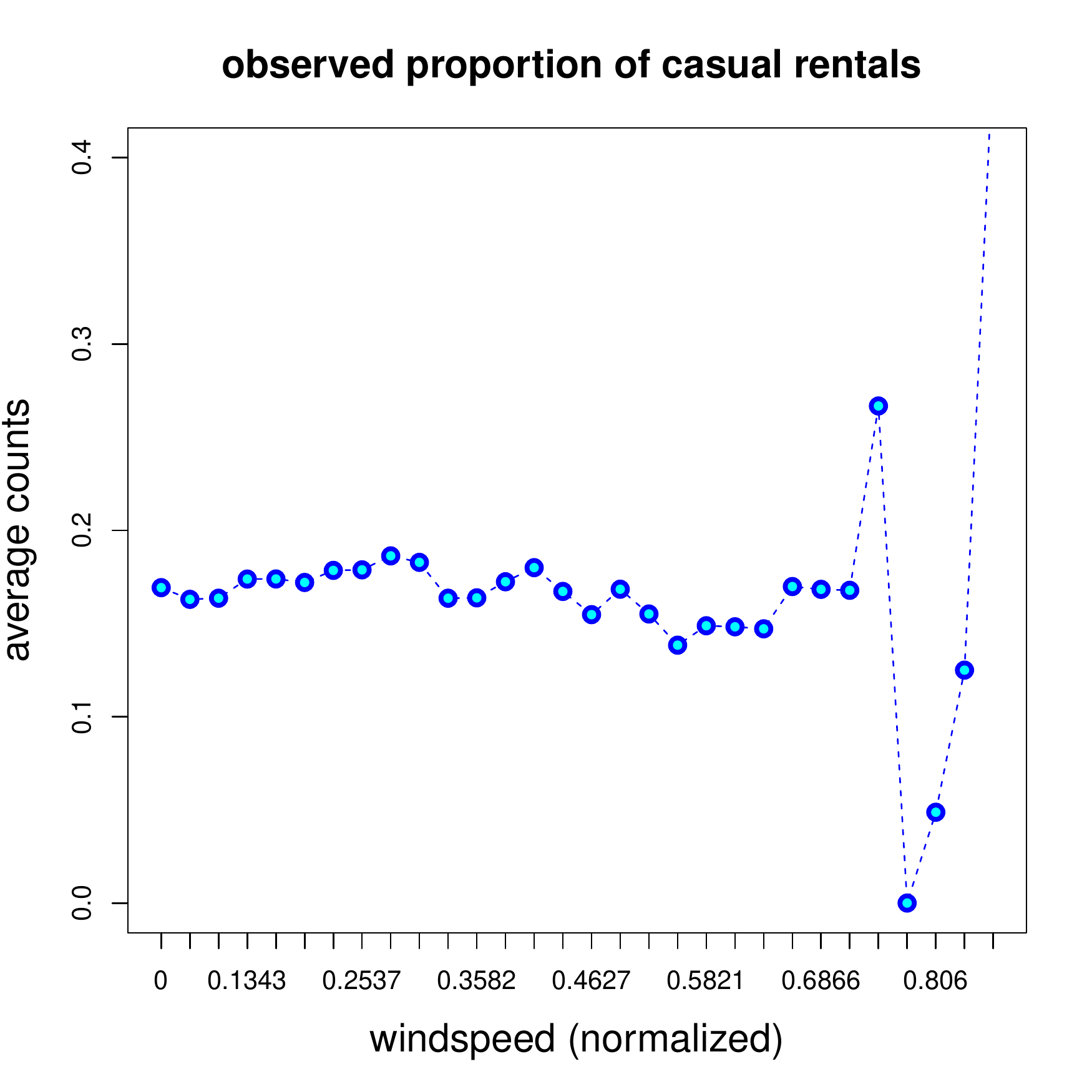}
\end{center}
\end{minipage}
\end{center}
\caption{Average response $Y$ for each label of all features {\tt date} (in weekly units), {\tt year}, {\tt month},
{\tt hour}, {\tt weekday}, {\tt holiday}, {\tt workingday}, {\tt weather}, {\tt temp}, {\tt temp\_feel},
{\tt humidity} and {\tt windspeed}.}
\label{descriptive counts 2}
\end{figure}

In Figure \ref{descriptive counts 2} we provide the marginal observed responses for 
each label of all features. The top-left shows the average response for each calendar
week from 2011/01/01 until 2012/12/31. This depicts a strong seasonal pattern of
the casual rentals proportion. Moreover, daytime, weekdays, working days/holidays
and weather conditions such as temperature is important information for predicting
the proportion of casual rentals. Only wind speed does not seem to be
very relevant. From the top-middle we also observe that the proportion of casual
rentals slightly decreases over time which can be explained by increasing regular
rental subscriptions from 2011 to 2012.

\begin{figure}[htb!]
\begin{center}
\begin{minipage}[t]{0.32\textwidth}
\begin{center}
\includegraphics[width=\textwidth]{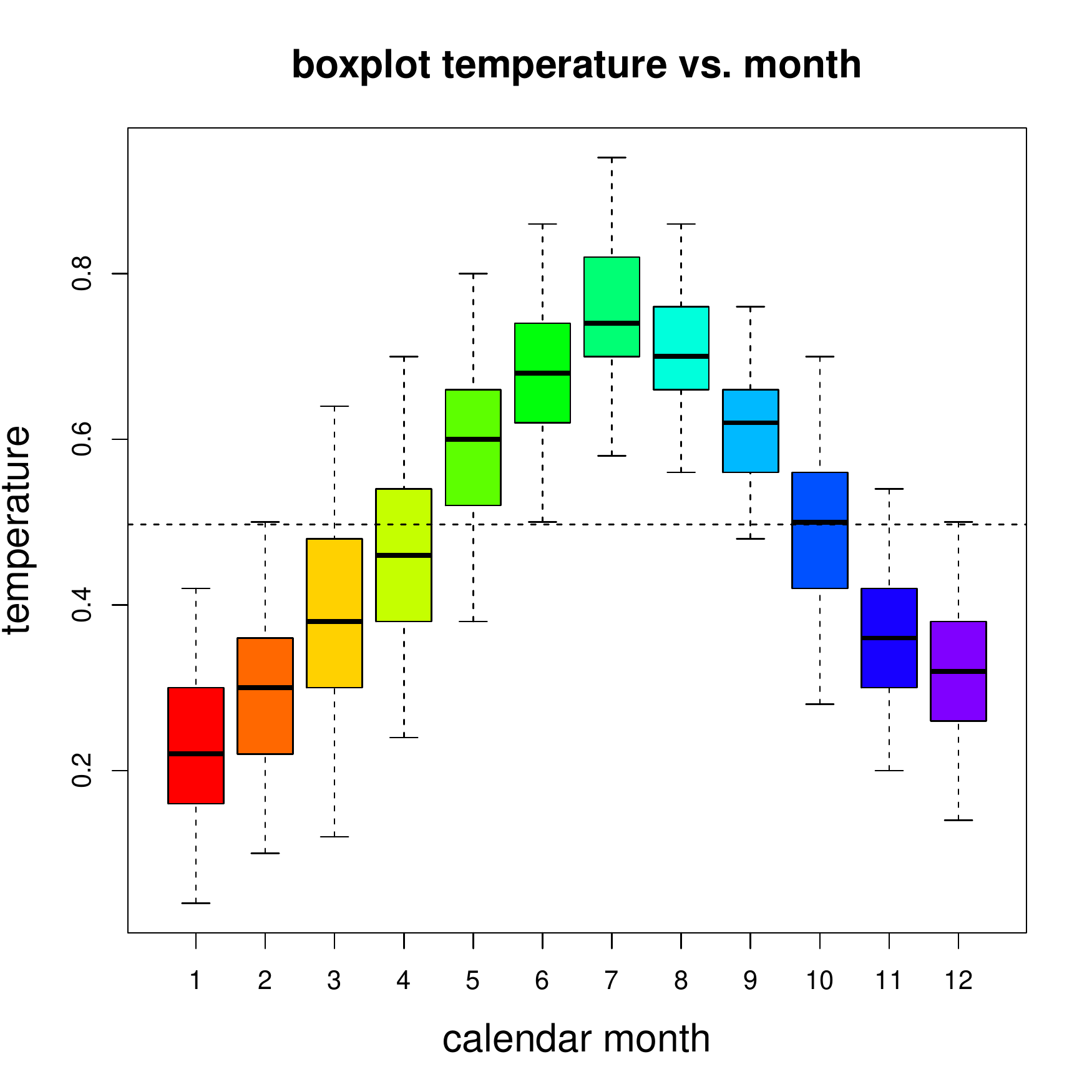}
\end{center}
\end{minipage}
\begin{minipage}[t]{0.32\textwidth}
\begin{center}
\includegraphics[width=\textwidth]{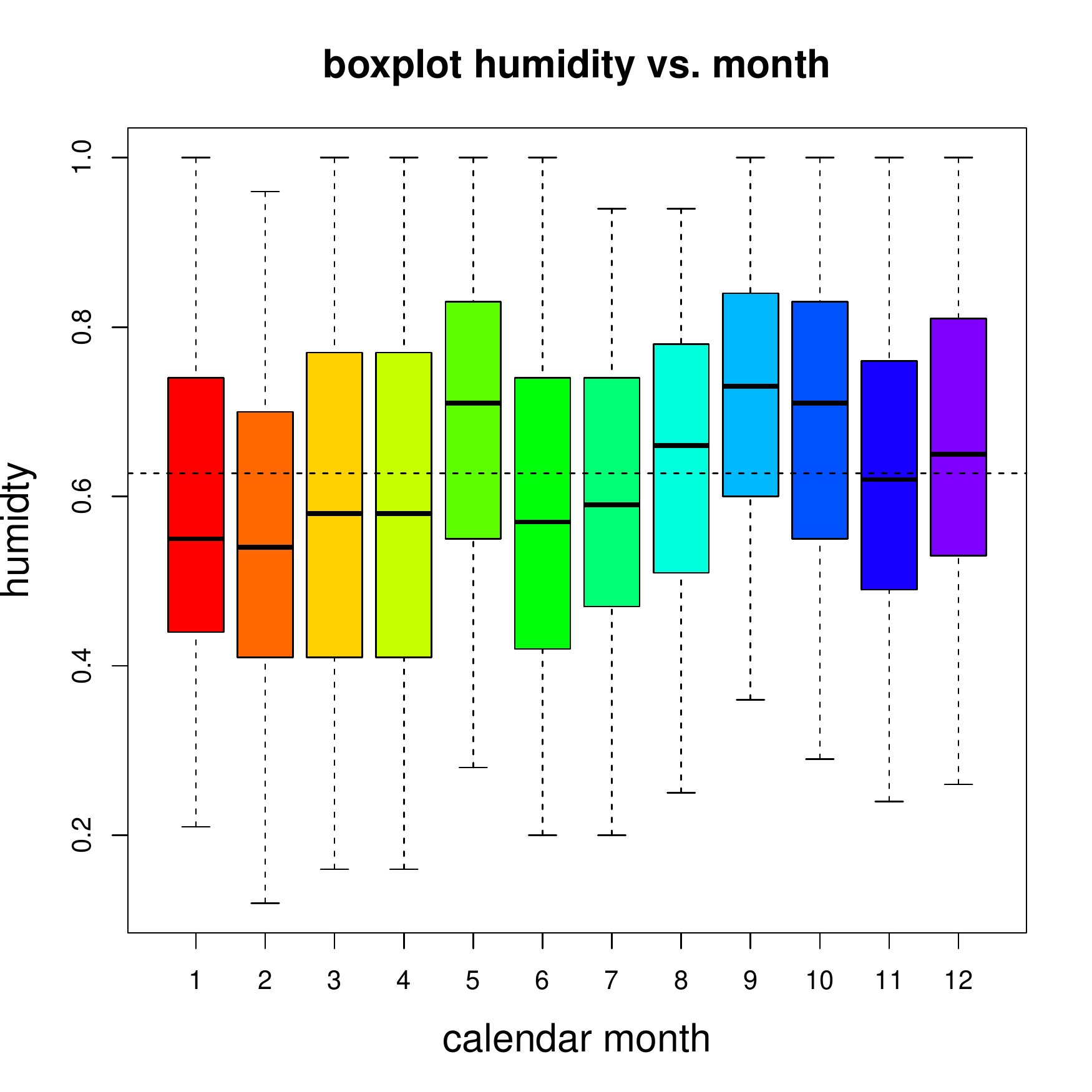}
\end{center}
\end{minipage}
\begin{minipage}[t]{0.32\textwidth}
\begin{center}
\includegraphics[width=\textwidth]{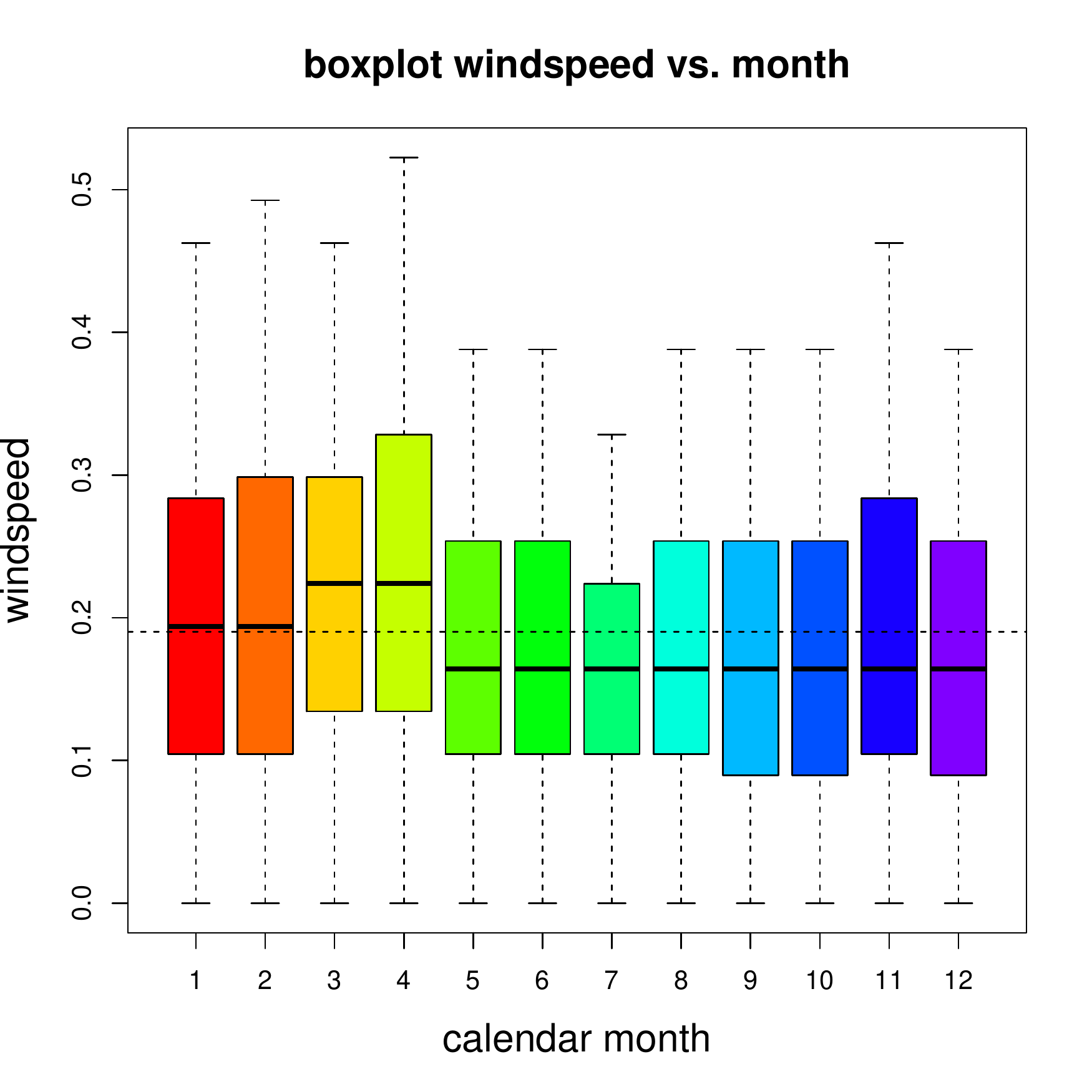}
\end{center}
\end{minipage}

\begin{minipage}[t]{0.32\textwidth}
\begin{center}
\includegraphics[width=\textwidth]{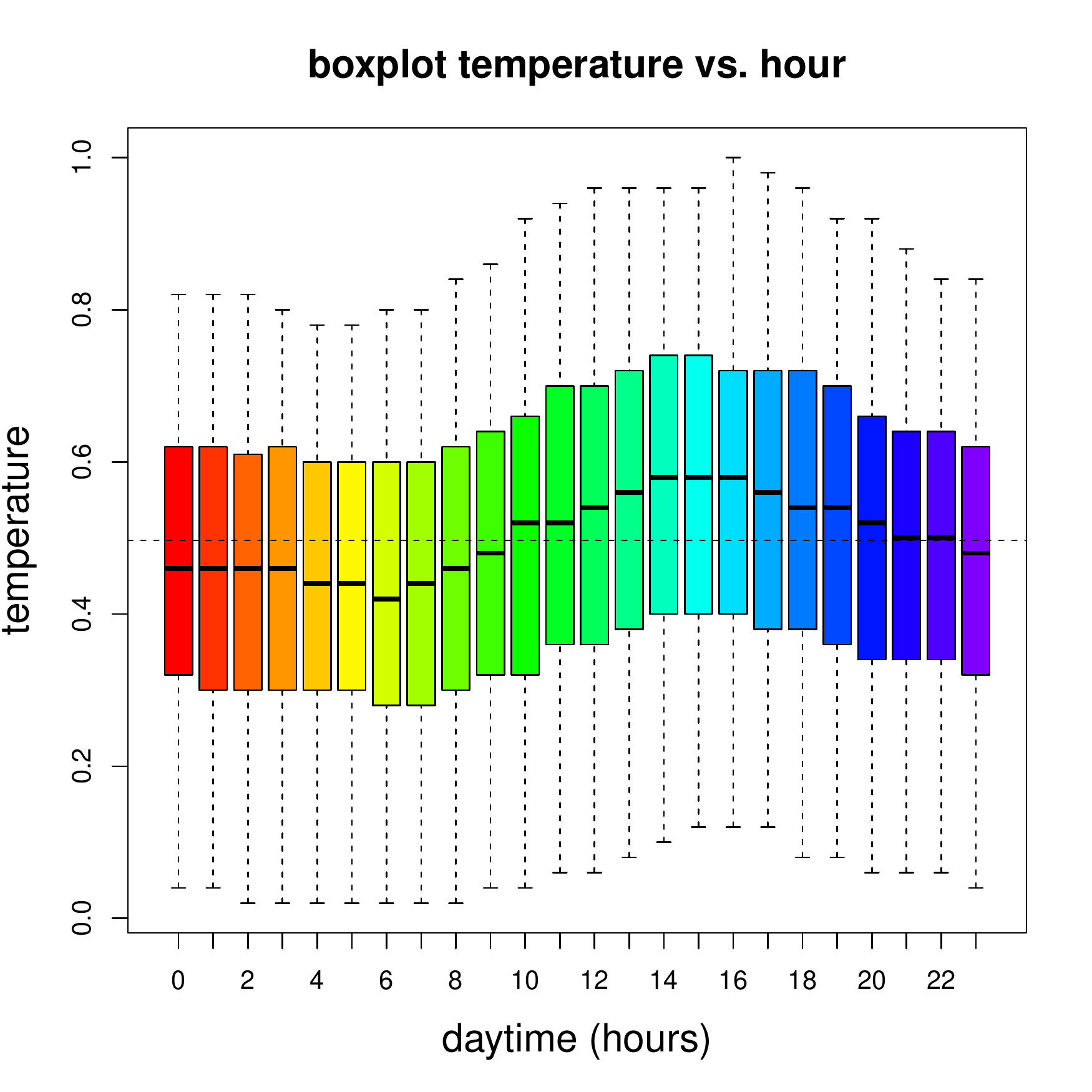}
\end{center}
\end{minipage}
\begin{minipage}[t]{0.32\textwidth}
\begin{center}
\includegraphics[width=\textwidth]{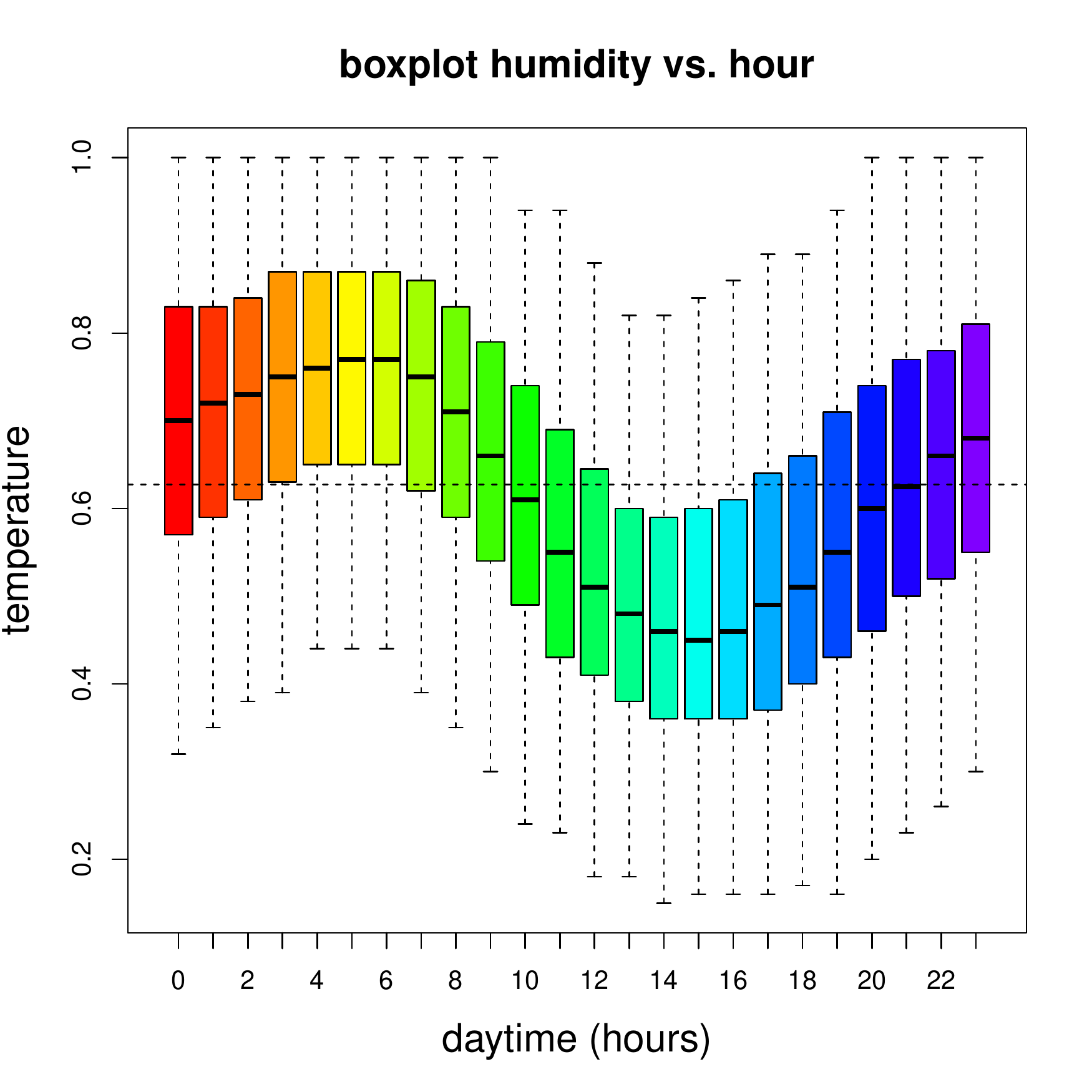}
\end{center}
\end{minipage}
\begin{minipage}[t]{0.32\textwidth}
\begin{center}
\includegraphics[width=\textwidth]{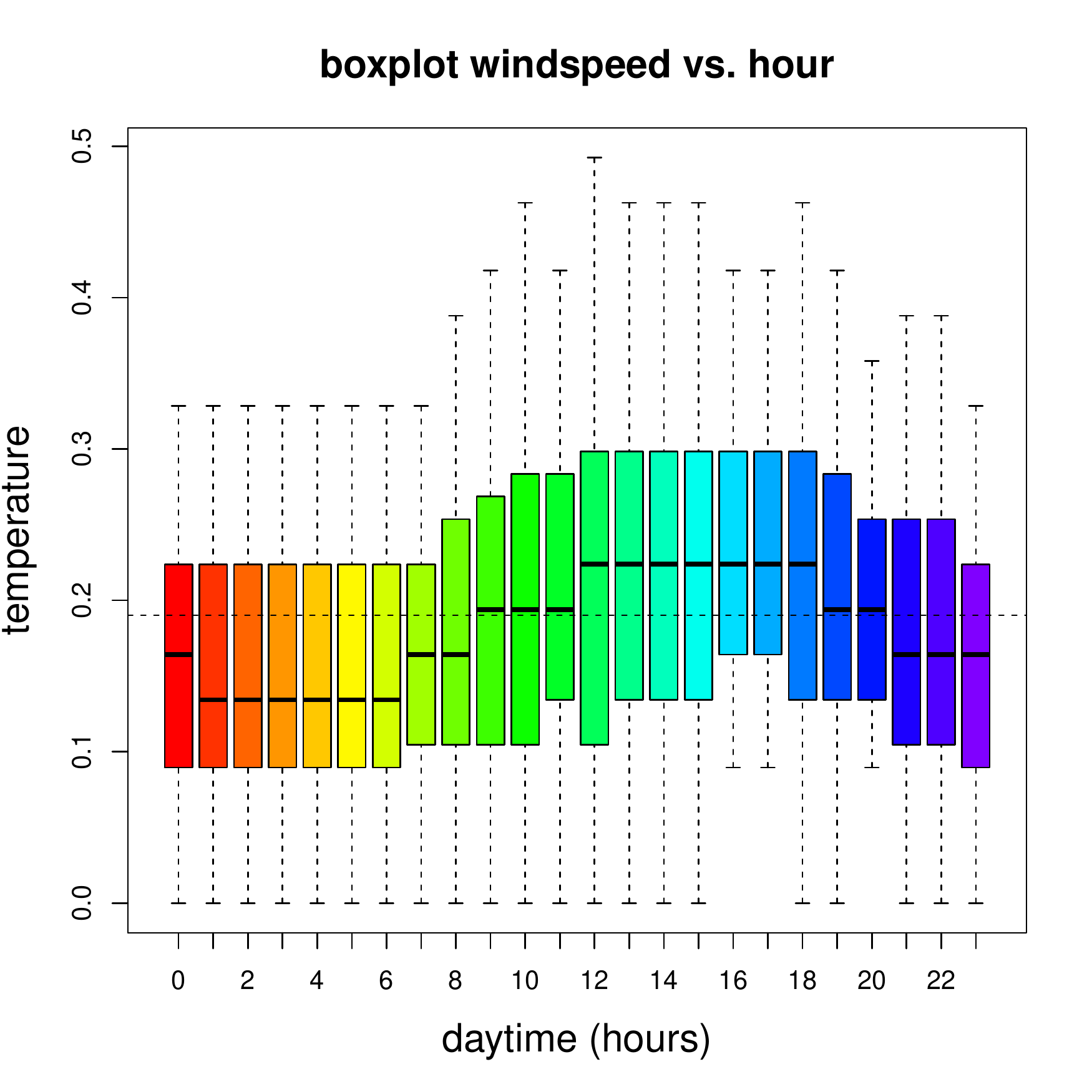}
\end{center}
\end{minipage}

\begin{minipage}[t]{0.32\textwidth}
\begin{center}
\includegraphics[width=\textwidth]{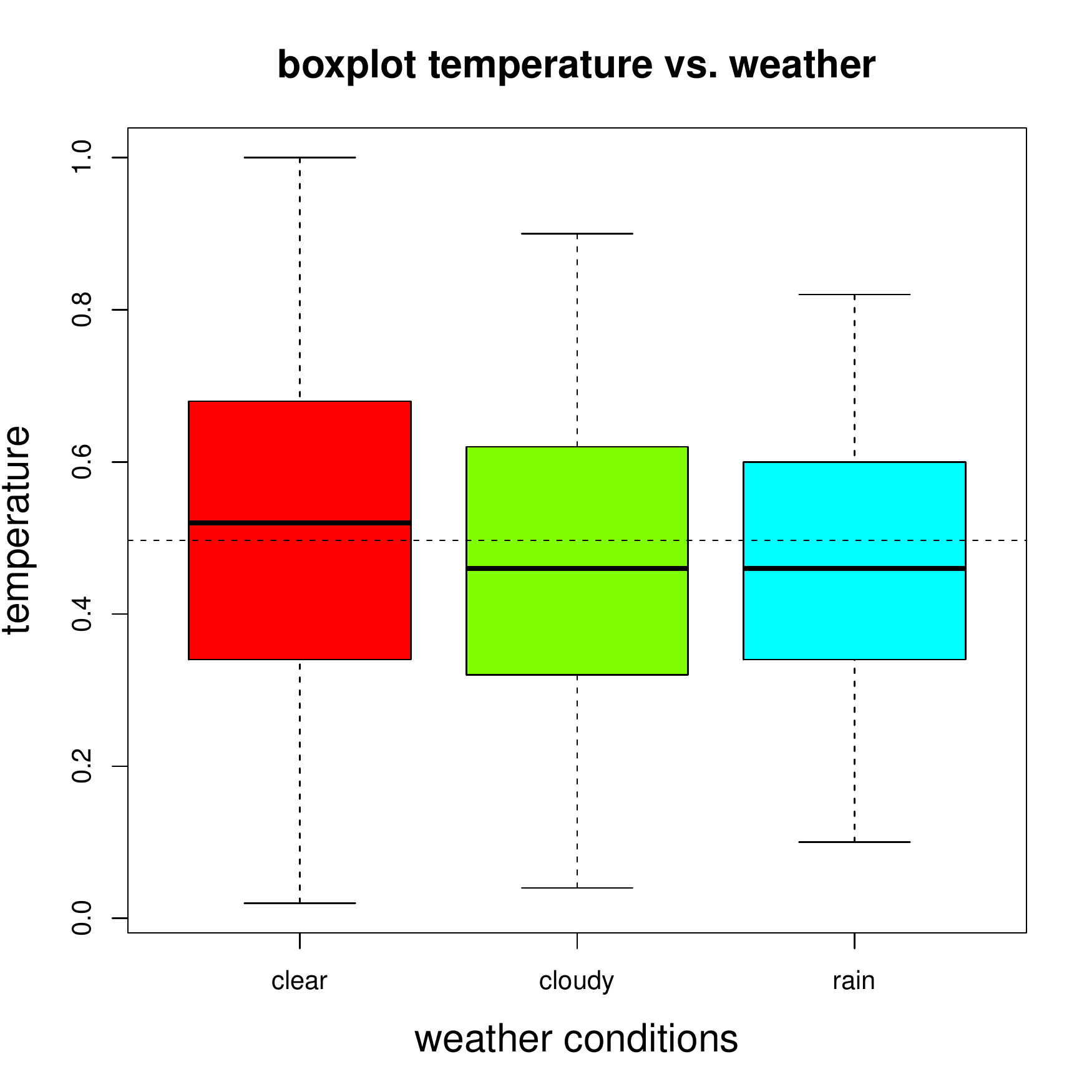}
\end{center}
\end{minipage}
\begin{minipage}[t]{0.32\textwidth}
\begin{center}
\includegraphics[width=\textwidth]{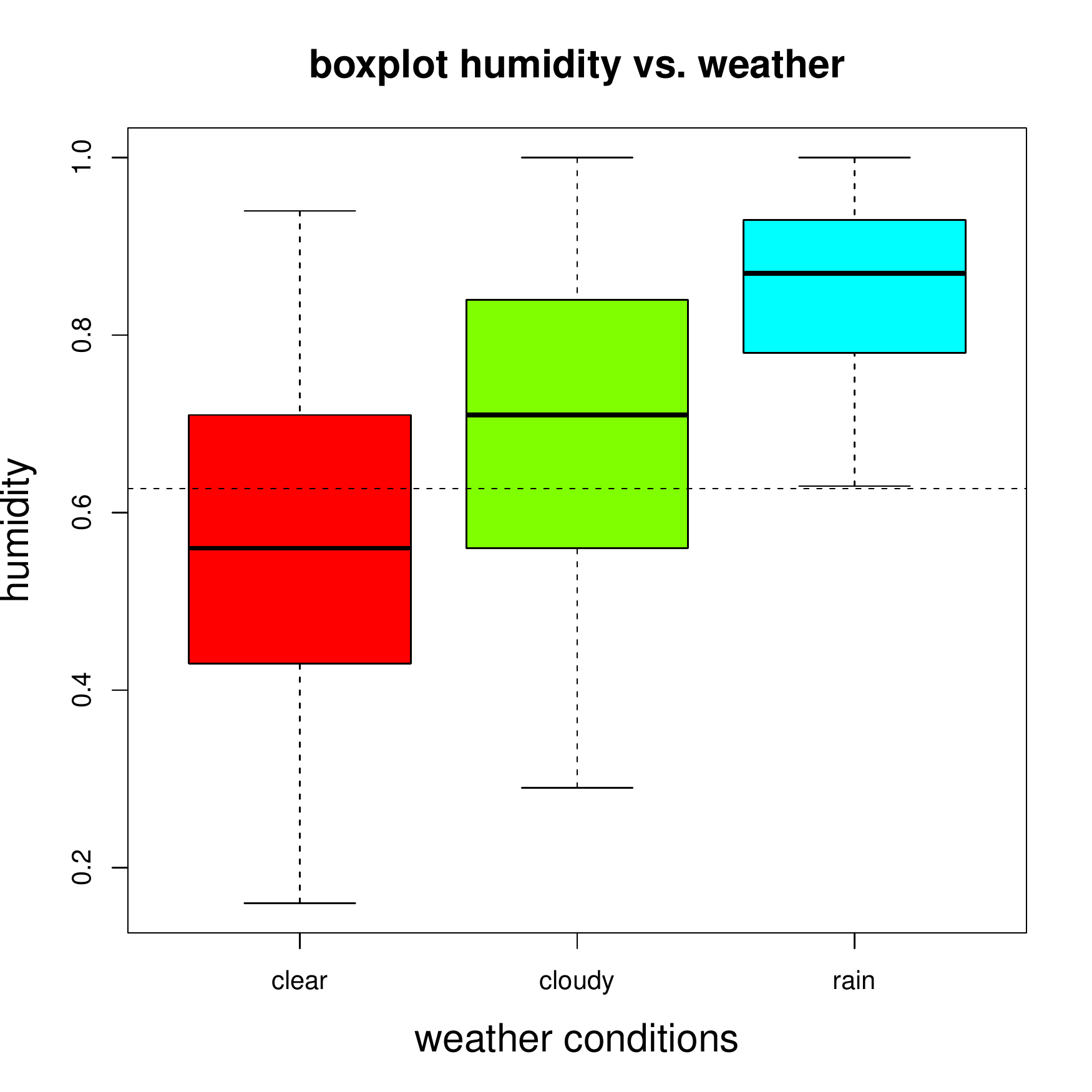}
\end{center}
\end{minipage}
\begin{minipage}[t]{0.32\textwidth}
\begin{center}
\includegraphics[width=\textwidth]{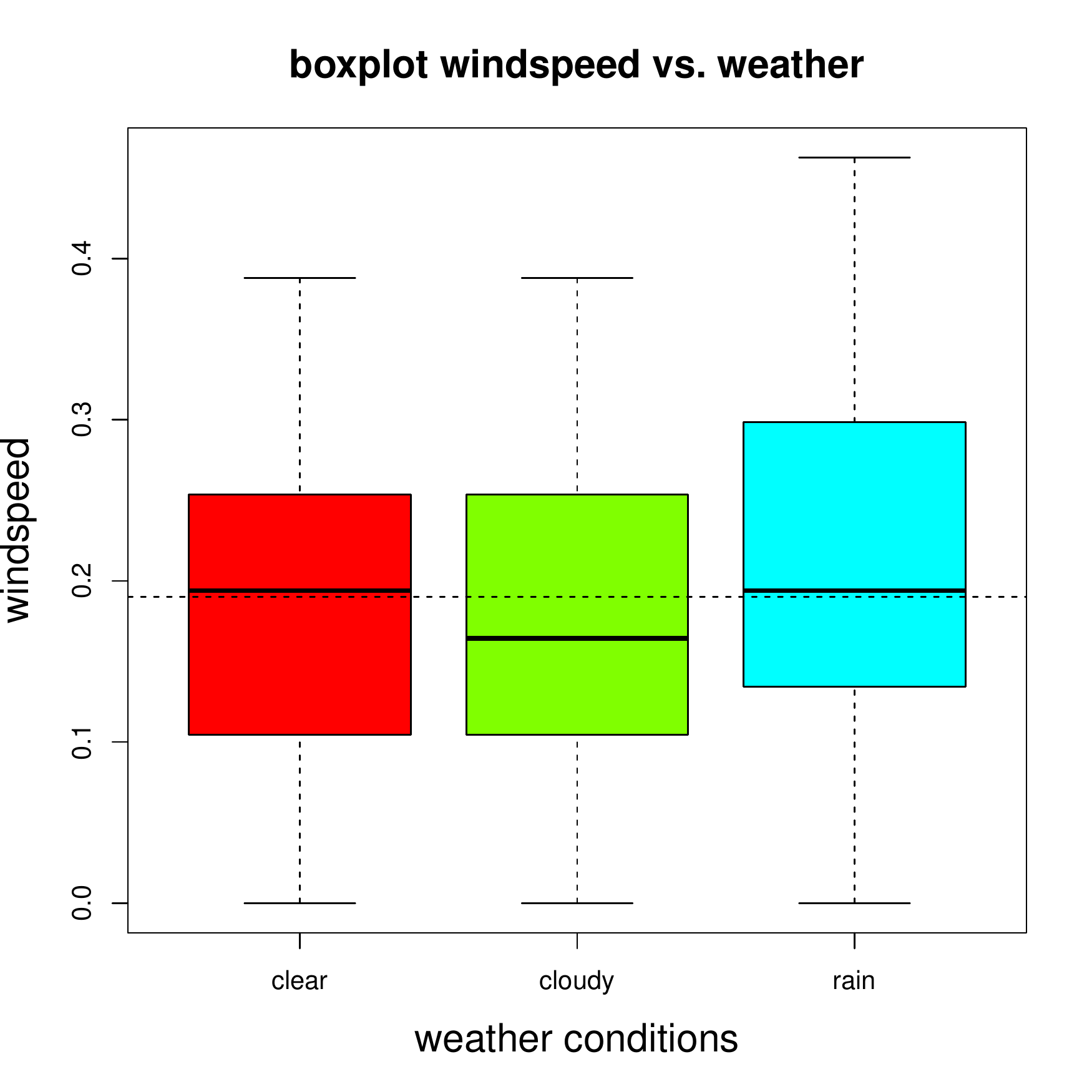}
\end{center}
\end{minipage}

\end{center}
\caption{Dependence between feature components: (top) temperature, humidity and wind speed
against calendar month, (middle) temperature, humidity and wind speed
against daytime, (bottom) temperature, humidity and wind speed
against weather conditions.}
\label{correlation plots}
\end{figure}

For many of the feature components it is clear that they are highly correlated.
In Figure \ref{correlation plots} we plot temperature, humidity and wind speed
against calendar month (top row), daytime (middle row) and weather conditions (bottom row).
These plots clearly show this dependence. Moreover, humidity is negatively correlated with
wind speed and positively correlated with temperature (at least up to moderate temperatures).

\end{document}